%% file: main.tex
\documentclass[preprint,12pt,authoryear]{elsarticle}

\usepackage{amssymb}
\usepackage{mathptmx}
\usepackage{amsmath,amsthm,amssymb}
\usepackage{xspace}
\usepackage{url}
\usepackage{complexity}

\usepackage{tabularx}
\newcolumntype{L}{>{\raggedright\arraybackslash}X}

\usepackage{pdflscape}
\usepackage{afterpage}
\usepackage{caption}

\usepackage[textsize=small,textwidth=2.5cm]{todonotes}

\usepackage{algorithm}
\usepackage{algorithmicx}
\usepackage{algpseudocode}

\usepackage{tablefootnote}

\theoremstyle{definition}

\newtheorem{example}{Example}

\newcommand{\SigmaP}[1]{\ComplexityFont{\Sigma}_{#1}^{\P}}
\newcommand{\PiP}[1]{\ComplexityFont{\Pi}_{#1}^{\P}}
\newcommand{\DeltaP}[1]{\ComplexityFont{\Delta}_{#1}^{\P}}
\newcommand{\ThetaP}[1]{\ComplexityFont{\Theta}_{#1}^{\P}}
\newcommand{\nOP}{\ComplexityFont{nOP}}
\newcommand{\OutputP}{\ComplexityFont{OutputP}}
\newcommand{\TotalP}{\ComplexityFont{TotalP}}
\newcommand{\DelayP}{\ComplexityFont{DelayP}}

\newcommand{\Cred}[1]{\textit{Cred}_{#1}}
\newcommand{\Skept}[1]{\textit{Skept}_{#1}}
\newcommand{\Exists}[1]{\textit{Exists}_{#1}}
\newcommand{\Ver}[1]{\textit{Ver}_{#1}}
\newcommand{\Enum}[1]{\textit{Enum}_{#1}}
\newcommand{\powerset}[1]{\ensuremath{2^{#1}}}							% power set
\newcommand{\F}{F}						% Abstract argumentation framework
\newcommand{\args}{\ensuremath{\mathsf{A}}\xspace}				% Set of arguments
\newcommand{\atts}{\ensuremath{\rightarrow}\xspace}					% Attack relation
\newcommand{\AFC}{\ensuremath{\F=(\args,\atts)}\xspace}		% Definition of an abstract argumentation framework

\newcommand{\Given}{\textit{Given}\xspace}
\newcommand{\decide}{\textit{decide}\xspace}
\newcommand{\Enumerate}{\textit{enumerate}\xspace}
\newcommand{\return}{\textit{return}\xspace}

\newcommand{\af}{AF}

\newcommand{\com}{\mathbf{CO}}
\newcommand{\pr}{\mathbf{PR}}
\newcommand{\prf}{\pr}
\newcommand{\st}{\mathbf{ST}}
\newcommand{\stb}{\st}
\newcommand{\sst}{\mathbf{SST}}
\newcommand{\semi}{\sst}
\newcommand{\sem}{\sst}
\newcommand{\stg}{\mathbf{STG}}
\newcommand{\stage}{\stg}
\newcommand{\gr}{\mathbf{GR}}
\newcommand{\grd}{\gr}
\newcommand{\id}{\mathbf{ID}}
\newcommand{\ideal}{\id}

\newcommand{\se}{\mathbf{SE}}
\newcommand{\ee}{\mathbf{EE}}
\newcommand{\dc}{\mathbf{DC}}
\newcommand{\ds}{\mathbf{DS}}
\newcommand{\dt}{\mathbf{D3}}

\newcommand{\yes}{\texttt{YES}}
\newcommand{\no}{\texttt{NO}}

\newcommand{\uni}{University\@}

\newcommand{\probCycles}{\texttt{probCycles}}
\newcommand{\probAttacks}{\texttt{probAttacks}}

\newcommand{\random}{\texttt{random}}
\newcommand{\innerAttackProb}{\texttt{innerAttackProb}}
\newcommand{\outerAttackProb}{\texttt{outerAttackProb}}
\newcommand{\nSCCs}{\texttt{nSCCs}}

\journal{Artificial Intelligence}

\begin{document}

\begin{frontmatter}


%%\title{Title\tnoteref{label1}}
\title{Design and Results of the Second International\\ Competition on Computational Models of Argumentation}
\author[ins1]{Sarah A. Gaggl}
\ead{sarah.gaggl@tu-dresden.de}
\author[ins2]{Thomas Linsbichler}
\ead{linsbich@dbai.tuwien.ac.at}
\author[ins3]{Marco Maratea\corref{cor1}}
\ead{marco@dibris.unige.it}
\author[ins2]{Stefan Woltran}
\ead{woltran@dbai.tuwien.ac.at}

\tnotetext[tref]{This paper is an extended and revised version of a paper presented at the First International Workshop on Systems and Algorithms for Formal Argumentation~\citep{GagglLMW16}, which included the design of the event before the competition was run. A brief survey of the competition is to be published in AI Magazine~\citep{GagglLMW18}.}
\cortext[cor1]{Corresponding author}
%\fntext[ref1]{Also affiliated with the University of Potsdam, Germany.}
%% \author{Francesco Calimeri\fnref{ins1}, Martin Gebser\fnref{ins2}, Marco Maratea\fnref{ins3}, Francesco Ricca\fnref{ins1}}
% \address{Address\fnref{ins3}}

%\title{}

\address[ins1]{Faculty of Computer Science, TU Dresden, Germany}
\address[ins2]{Faculty of Informatics, TU Wien, Austria}
\address[ins3]{Dipartimento di Informatica, Bioingegneria, Robotica e Ingegneria dei Sistemi,\\ Universit{\`a} di Genova, Italy}

%\author{}

%\address{}

\begin{abstract}

Argumentation is a major topic in the study of Artificial Intelligence. Since the first edition in 2015, advancements in solving (abstract) argumentation frameworks are assessed in competition events, similar to other closely related problem solving technologies. 
In this paper, we report about the design and results of the Second International Competition on Computational Models of Argumentation, which has been jointly organized by TU Dresden (Germany), TU Wien (Austria),
and the University of Genova (Italy), in affiliation with the
2017 International Workshop on Theory and Applications
of Formal Argumentation. %% Following the first edition in 2015, the competition evaluates the performance of submitted solvers on computational problems within abstract argumentation. 
This second edition maintains some of the design choices made in the first event, e.g. the I/O formats, the basic reasoning problems, and the organization into tasks and tracks. At the same time, it introduces significant novelties, e.g. three additional prominent semantics, and an instance selection stage for classifying instances according to their empirical hardness.

%% In addition to the four original semantics, ICCMA'2017 includes . Moreover, a dedicated call for benchmarks allowed for introducing a sophisticated 

\end{abstract}


\end{frontmatter}


%\input{introduction}
\section{Introduction}
\label{sec:intro}
%[{\bf TODO: Stressing why this is important for the (wider) AI audience...}]

Computational Argumentation is a multidisciplinary area at the intersection of Philosophy, Artificial Intelligence (AI), Linguistics, Psychology, and several application domains \citep{Bench-CaponD07}.
%the most prominent of which certainly is legal reasoning. 
Within AI, several subfields are particularly relevant to -- and benefit from -- studies of argumentation. 
These include decision support, knowledge representation, nonmonotonic reasoning, and multiagent systems. 
%The field has seen a steady rise of interest over the last two decades%[BD2007,BH2008,RS2009]. 
%The success of the field is also due to new AI techniques which allow for a 
Moreover, computational argumentation provides a
formal investigation of problems that have been studied informally only by philosophers, and which consequently allow for the development of computational tools for argumentation, see \citep{aimag07}.

Since its invention by 
\cite{Dung:1995}, abstract argumentation based on argumentation frameworks (AFs) has become a key concept 
for the field. In AFs, argumentation scenarios are modeled as simple directed graphs, where the vertices represent
arguments and each edge corresponds to an attack between two arguments.
%direction. %within AI. %\cite{argu-book,Bench-CaponD07}.
Besides its simplicity, 
there are several reasons for the success story of this concept:
First, a multitude of
semantics  \citep{Baroni:2011,BaroniCG18} %,CaminadaG2009} 
allows for tight coupling
of argumentation with existing formalisms from the areas of knowledge representation and logic programming;
indeed, one of the main motivations of Dung's work \citep{Dung:1995} was to give a uniform
representation of several nonmonotonic formalisms including Reiter's Default Logic, 
Pollock's Defeasible Logic, and Logic Programming (LP) with default negation; the latter
lead to a series of works that investigated the relationship between different LP semantics 
and different AF semantics, see e.g.\ \citep{WuCG09,CaminadaSAD15a}.
Second, abstract argumentation is employed as a core method in advanced
argumentation formalisms like ASPIC+ \citep{ModgilP14} or the ABA framework \citep{Cyras18};
in particular, semantics for such formalisms are often defined via a representation that makes use of AFs, and moreover, some of the systems implementing 
 ASPIC+  or ABA rely on efficient solvers for abstract argumentation.
%Finally, %and more recently, 
Consequently,
an increasing amount of work has been focused on the development of efficient algorithms 
and systems for AFs, see 
\citep{DBLP:journals/ai/CharwatDGWW15} for a survey. 
%an increasing number of systems
%for evaluating AFs has been developed, e.g.\
%\cite{BistarelliS11,EglyGW10,DvorakJWW2014,CeruttiGV2014,NofalAD2014}.

Given this development, it was soon recognized that there 
is a need for systematic benchmarking in order 
%a structured comparison of 
to have a solid comparison 
of the different methods and systems that have been proposed. 
This is witnessed by a number of papers on the topic, e.g.\
%While different 
%further evaluations, in particular
\citep{DBLP:journals/fuin/BistarelliRS15,CeruttiVG16a,Bistarelli17,VallatiCG18}
%% [Also check (if gets available)
%% ``Predictive Models and Abstract Argumentation: the case of
%% High-Complexity Semantics'' submitted to KER; 
%% and maybe.]
and cumulated in the creation and organization of 
the International Competition on Computational Models of Argumentation (ICCMA).
%Inspired by the success of 
%competitions in other fields, a dedicated competition 
%has been organized, 
The first edition took place in 2015 and focused on four prominent semantics; 
18 solvers were competing in this event, see \citep{Thimm:2016,ThimmV17} for details.

In this report, we present the design and results of the Second International Competition on Computational Models of Argumentation (ICCMA'17)\footnote{%
%\url{https://www.dbai.tuwien.ac.at/iccma17/}},
\url{http://argumentationcompetition.org/2017/}},
which has been jointly organized by TU Dresden (Germany), TU Wien (Austria), and the University of Genova (Italy),
in affiliation with the 2017 International Workshop on Theory and Applications of Formal Argumentation (TAFA'17). ICCMA'17 has been conducted in the first half of 2017, and comes two years after the first edition.
%, ICCMA'15~\cite{Thimm:2016}.\footnote{\url{http://argumentationcompetition.org/2015}}

The general goal of this competition is to consolidate and strengthen the ICCMA series, which in its first edition had very good outcomes in some respects, e.g.\ in terms of the  number of submitted solvers (18, as already mentioned above).
The second edition maintains some of the design choices previously made, e.g.\
the I/O formats and the basic reasoning problems.
With a slight modification to the first edition,
%Also similar to the first edition, 
the competition is organized into \emph{tasks} and \emph{tracks},
where a \emph{task} is a reasoning problem under a particular semantics,
and a \emph{track} collects different tasks over a semantics.
ICCMA'17
also
introduces several novelties:
%\begin{itemize}
$(i)$
%\item 
a new scoring scheme is implemented for better reflecting the solvers' behavior,
%\item 
$(ii)$
three new semantics are included, namely semi-stable, stage and ideal semantics,
%\item 
$(iii)$
a special ``Dung's Triathlon'' track is added, where solvers are required to deal with different problems simultaneously, with the goal of testing the solvers' capability of
exploiting interrelationships among semantics, and
%\item 
$(iv)$
a ``call for benchmarks'' has been performed, to enrich the suite of instances for the competition, followed by a novel instance selection stage.

In addition to the report of the competition, we also compare in this article the performance
of the ICCMA'15 winning systems to the current leaders.
%in order to document the progress made within the community on the solving side.  

Besides its importance for the argumentation community, the ICCMA series is also of interest
for researchers beyond this field. This is due to the following two reasons:
\begin{itemize}
\item Solvers need to handle a variety of different semantics which range over different levels of complexity; in ICCMA'17 we put even more emphasis on this rather unique feature by the introduction of the Dung's triathlon, where the systems are required to solve problems situated at three different complexity layers, preferably exploiting interrelationships between these problems. (We note that problems of different complexity are also 
present in other competitions, e.g.\ in Quantified Satisfiability (QBF) or in Answer Set Programming (ASP) competitions, see \citep{Pulina16,CalimeriGMR16,GebserMR17}); however, the situation is more challenging 
in argumentation since the diverse complexity actually stems from the different semantics which require
different computational tasks including subset-maximization, fixed-point computations, etc.)
%Compared to other competitions where problems from different complexity levels are 
%common (QBF solving, ASP), in argumentation the situation is different since also the underlying problem differ (make clear!)
\item Given the range of submitted solvers, we see a great variety of approaches. In particular, various methods including (different forms of) reductions to SAT, ASP, constraint satisfaction, and circumscription are employed in the submitted systems. Thus, ICCMA also provides (to a certain extent) an interdisciplinary comparison between different reasoning paradigms in AI.\footnote{%
It has to be mentioned that this not a completely new phenomenon. For instances, SAT-based approaches competed in ASP competitions, see, e.g. \citep{GiunchigliaLM06}, and likewise, an ASP-based approach for $2$-QBF solving participated \citep{AmendolaDR16} to the 2016 QBF evaluation.}. %However, we are not aware of such a xxx
%\item what else?
\end{itemize}
 
\noindent
The report is structured as follows. Section~\ref{sec:back} introduces preliminaries about abstract argumentation, with focus on the semantics evaluated in the competition. Then, Section~\ref{sec:format} presents the design of the competition. Section~\ref{sec:bench} and~\ref{sec:bench-sel} are devoted to the description of the benchmark suite employed in the competition, and the instance selection process, respectively. Section~\ref{sec:part} then presents the participating solvers. The results of the competition, with respective award winners, are then presented in Section~\ref{sec:res}. The report ends in Section~\ref{sec:rel} with a discussion on how the novelties introduced are treated in related competitions, and in Section~\ref{sec:conc} with conclusions and final remarks.

\section{Background}
\label{sec:back}

An \emph{abstract argumentation framework} (\af, for short) %as introducted by
is a tuple $\F=(\args,\atts)$ where
\args is a set of arguments and
\atts is a relation $\atts\,\,\subseteq \args\times\args$~\citep{Dung:1995}.
For two arguments $a,b\in\args$ the relation $a \atts b$ means that argument $a$ \emph{attacks} argument $b$.
%For $\cA\in\args$ define $\attackers{\F}{\cA}=\{\cB\mid \cB\atts \cA\}$.
An argument $a \in \args$ is \emph{defended} by $S \subseteq \args$ (in $\F$)
if for each $b \in \args$ such that $b \atts a$
there is some $c \in S$ such that $c \atts b$.
%Define $F: 2^{\args} \rightarrow 2^{\args}$ via $F(S)  = \{ \cA\in\args \mid S \text{~defends~} \cA\}$ where a set $S\subseteq\args$ defends an argument $\cA$ if for all arguments $\cB\in \args$, if $\cB\atts \cA$ then there is $\cC\in E$ with $\cC\atts \cB$.
A set $E \subseteq \args$ is \emph{conflict-free} (in \F) if and only if there are no $a,b\in E$ with $a \atts b$.
$E$ is \emph{admissible} (in \F) if and only if it is conflict-free and each $a \in E$ is defended by $E$.
Finally, the range of $E$ (in \F) is given by $E^{+}_\F=E \cup \{a\in\args \mid \exists b\in E: b\atts a\}$.

\emph{Semantics} are used to determine
sets of jointly acceptable arguments by
mapping each \af{} \AFC to a set of \emph{extensions} $\sigma(\F) \subseteq \powerset{\args}$.
The extensions under complete ($\com$), preferred ($\prf$), stable ($\st$), semi-stable ($\sst$) \citep{CaminadaCD12}, stage ($\stg$) \citep{Verheij:1996}, grounded ($\gr$) and ideal ($\id$) \citep{DungMT:2007}
semantics are defined as follows.
Given an \af{} \AFC and a set $E \subseteq \args$,
%given to abstract argumentation frameworks by means of extensions \cite{Dung:1995} or labelings \cite{Caminada2009,Wu:2010}. For what follows, we use the former.

\begin{itemize}
	\item $E \in \com(\F)$ iff $E$ is admissible in $\F$ and if $a \in \args$ is defended by $E$ then $a\in E$,	
	\item $E \in \prf(\F)$ iff $E \in \com(\F)$ and there is no $E' \in \com(\F)$ s.t.\ $E' \supset E$,
        \item $E \in \stb(\F)$ iff $E \in \com(\F)$ and $E^+_\F = \args$,
	\item $E \in \sst(\F)$ iff $E \in \com(\F)$ and there is no $E' \in \com(\F)$ s.t.\ $E'^+_\F \supset E^+_\F$,
	\item $E \in \stg(\F)$ iff $E$ is conflict-free in $\F$ and there is no 
$E'$ such that 
$E'$ is 
conflict-free  in $\F$ and 
$E'^+_\F \supset E^+_\F$,
        \item $E \in \gr(\F)$ iff $E \in \com(\F)$ and there is no $E' \in \com(\F)$ s.t.\ $E' \subset E$,
	\item $E \in \id(\F)$ iff $E$ is admissible in $\F$, $E \subseteq \bigcap{\pr(\F)}$ and
          there is no $E' \subseteq \bigcap{\pr(\F)}$ s.t.\ $E'$ is admissible in $F$ and $E' \supset E$,
\end{itemize}

%% where $\com$, $\prf$, $\st$, $\sst$, $\stg$, $\gr$ and $\id$ stand for complete, preferred, stable, semi-stables
For more discussion on these semantics we refer to \cite{Baroni:2011}.

Note that both grounded and ideal extensions are uniquely determined and always exist \citep{Dung:1995,DungMT:2007}.
Thus, they are also called
\emph{single-status} semantics.
The other semantics introduced are \emph{multi-status} semantics.
That is, there is not always a unique extension induced by the semantics.
For all semantics except stable semantics, there always exists
at least one extension, whereas the set of stable extensions can be empty.
If the set of stable extensions is non-empty,
it coincides with the set of semi-stable extensions and
with the set of stage extensions,
i.e.\ $\stb(\F)=\semi(\F)=\stage(\F)$ whenever $\stb(\F)\neq \emptyset$.

\tikzstyle{arg}=[draw, thick, circle, fill=gray!15,inner sep=2pt, minimum size=.6cm]
\tikzstyle{slt}=[loop left,thick, distance=0.5cm, out=50, in=130,->]
\tikzstyle{slb}=[loop left,thick, distance=0.5cm, out=-130, in=-50,->]
\tikzstyle{slr}=[loop left,thick, distance=0.5cm, out=-40, in=40,->]
\tikzstyle{sll}=[loop left,thick, distance=0.5cm, out=140, in=-140,->]

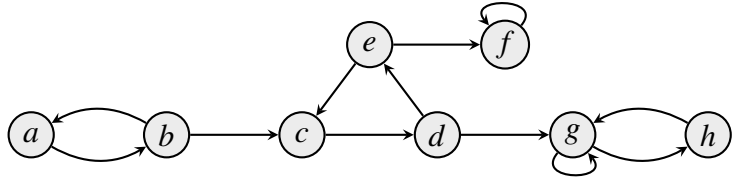
\begin{figure}[t]
\centering
\begin{tikzpicture}[>=stealth]
\path 	node[arg](a){$a$}
                ++(1.8,0)node[arg](b){$b$}
		++(1.8,0) node[arg](c){$c$}
		++(1.8,0) node[arg](d){$d$}
		++(-0.9,1.2) node[arg](e){$e$}
                ++(1.8,0) node[arg](f){$f$}
                ++(0.9,-1.2)node[arg](g){$g$}
                ++(1.8,0) node[arg](h){$h$};
\path [left,->, thick]
  (b) edge (c)
  (c) edge (d)
  (d) edge (e)
  (e) edge (c)
  (e) edge (f)
  (d) edge (g)
  (a) edge[bend right] (b)
  (b) edge[bend right](a)
  (g) edge[bend right] (h)
  (h) edge[bend right](g);
  
\draw[slt] (f) edge (f);
\draw[slb] (g) edge (g);

\end{tikzpicture}
\caption{An argumentation framework.}
\label{fig:semantics}
\end{figure}

\begin{example}
\label{ex:semantics}
To illustrate the semantics,
consider the following \af{}:
\begin{align*}
F=(&\{a,b,c,d,e,f,g,h\},\\
   &\{(a,b),(b,a),(b,c),(c,d),(d,e),(d,g),(e,c),(e,f),(f,f),(g,g),(g,h),(h,g)\}).
\end{align*}
$F$ is depicted in Figure~\ref{fig:semantics},
%Recall that
%Note that
where
nodes represent arguments and directed edges represent attacks.
First,
the conflict-free sets of $F$ are as follows:
\begin{align*}
\{&\emptyset,\{a\},\{b\},\{c\},\{d\},\{e\},\{h\},\{a,c\},\{a,d\},\{a,e\},\{a,h\},\{b,d\},\{b,e\},\\
&\{b,h\},\{c,h\},\{d,h\},\{e,h\},\{a,c,h\},\{a,d,h\},\{a,e,h\},\{b,d,h\},\{b,e,h\}\}.
\end{align*}
Note that no set containing $f$ or $g$ can be conflict-free,
since both $f$ and $g$ are self-attacking.
Among the conflict-free sets, the following sets are admissible:
\[
\{\emptyset,\{a\},\{b\},\{h\},\{a,h\},\{b,d\},\{b,h\},\{b,d,h\}\}.
\]
%The empty set is always conflict-free and admissible.
The conflict-free set $\{a,d\}$, for instance,
is not admissible since $d$ is attacked by $c$ in $F$,
but $\{a,d\}$ does not attack $c$,
i.e.\ it does not defend $d$.\\

%The naive extensions are just the $\subset$-maximal
%conflict-free sets:
%\[
%\naive(F)=\{\{a,c,h\},\{a,d,h\},\{a,e,h\},\{b,d,h\},\{b,e,h\}\}.
%\]
\noindent
For stable semantics, it can be checked
that there is no conflict-free set of arguments
in $F$ attacking all other arguments, hence:
\[
\stb(F)=\emptyset.
\]
The complete extensions of $F$ are those admissible sets
which do not defend any argument not contained in the set:
\[
\com(F)=\{\emptyset,\{a\},\{h\},\{a,h\},\{b,d,h\}\}.
\]
For instance, the admissible set $\{b,d\}$ is not complete
since it defends $h$.
As no argument of $F$ is unattacked,
the grounded extension is empty:
\[
\grd(F)=\{\emptyset\}.
\]
The preferred extensions are just the $\subseteq$-maximal
admissible sets, which always coincide
with the $\subseteq$-maximal complete extensions:
\[
\prf(F)=\{\{a,h\},\{b,d,h\}\}.
\]
The semi-stable and stage extensions of $F$ are given as follows:
\begin{align*}
\semi(F)=&\{\{b,d,h\}\}.\\
\stage(F) =&\{\{a,e,h\},\{b,e,h\},\{b,d,h\}\}.
\end{align*}
%Note that the set of semi-stable extensions
%are always a subset of the set of preferred extensions,
%and likewise for stage and naive.
%Note that $\{b,d\}$ is the only semi-stable extension of $F$
%by
%$\{a,h\}^+_F = \{a,b,g,h\} \subset \{a,b,c,d,e,g,h\} = \{b,d,h\}^+_F$.
%Likewise,
%for instance,
%$\{a,c,h\}\notin\stage(F)$ since
%$\{a,c,h\}^+_F = \{a,b,c,d,g,h\} \subset \{a,b,c,d,e,g,h\} = \{b,d,h\}^+_F$.
Finally,
$\{h\} = \bigcap{\prf(F)}$ and $\{h\}$ is admissible, hence
\[
\ideal(F) = \{\{h\}\}.
\]
\end{example}

In order to reason with multi-status semantics, usually, one takes either a credulous or skeptical perspective.

Given a semantics\footnote{For the sake of uniformity, we include here also the single-status semantics $\gr$, $\id$; clearly, in this case credulous and skeptical acceptance coincides.} $\sigma\in\{\com,\pr,\st,\sst,\stg,\gr,\id\}$, we thus define the following decision problems:
\begin{itemize}
\item
$\Cred{\sigma}$:
Given an \af{} \AFC and 
argument $a \in \args$, $a$ is
\emph{credulously accepted} in $\F$ under semantics $\sigma$ if there is a $\sigma$-extension $E \in \sigma(\F)$ with $a\in E$;
\item
$\Skept{\sigma}$:
Given an \af{} \AFC and 
argument $a \in \args$, $a$ is
\emph{skeptically accepted} in $\F$ with semantics $\sigma$ if for all $\sigma$-extensions $E\in \sigma(\F)$ it holds that $a\in E$.
\end{itemize}

Recall that stable semantics is the only case where an \af{} might possess no extension. In such a situation,
each argument is defined to be skeptically accepted.

Further reasoning problems for any semantics $\sigma$ are defined as follows:
\begin{itemize}
\item $\Ver{\sigma}$:
Given an AF \AFC and a set of arguments $S \subseteq \args$,
decide whether $S \in \sigma(\F)$.
\item $\Exists{\sigma}$:
Given an AF \AFC,
decide whether there exists an $S \in \sigma(\F)$.
\item $\Exists{\sigma}^{\neg\emptyset}$:
Given an AF \AFC,
decide whether there exists an $S \in \sigma(\F)$ with $S \neq \emptyset$.
\item $\Enum{\sigma}$:
Given an AF \AFC,
enumerate the set $\sigma(\F)$.
\end{itemize}

Complexity of reasoning problems under the various semantics
has been studied in~\citep{DimopoulosT96,DunneB02,CaminadaCD12,DvorakW10,DunneDW13,KroellPW17}.
The most recent survey can be found in~\citep{DvorakD18}.
Table~\ref{tab:complexity_af} provides an overview.
We thereby assume familiarity with basic concepts such as
completeness and the polynomial hierarchy (see~\citep{AroraB09} for more details).
The class $\ThetaP{k}$ is a refinement of the class $\DeltaP{k}$:
it contains the problems that can be decided in polynomial time by a deterministic
Turing machine with at most $\mathcal{O}(\log m)$ calls to a $\SigmaP{k-1}$ oracle,
%in which the number of oracle
%calls in bounded by ,
where $m$ is the input size.
By $\nOP$ we denote that the enumeration problem
is not contained in the class $\OutputP$ (also called $\TotalP$),
i.e.\ it is not solvable in polynomial time in the size
of the input and the output~\citep{JohnsonPY88,Strozecki10}\footnote{%
Note that the result for $\ideal$ is not published,
but immediate by the fact that $\Ver{\ideal}$ is $\coNP$-complete~\citep{Dunne09}
and therefore the ideal extension is not computable in polynomial time.}.
Containment in $\DelayP$ on the other hand means that
the extensions can be enumerated with a delay which is polynomial
in the size of the input.

\begin{table}[t]
\renewcommand{\arraystretch}{1.25}
\centering
\caption{Complexity of reasoning with \af{}s. $\mathcal{C}$-c means that the problem is complete for class $\mathcal{C}$.}
\label{tab:complexity_af}
\label{tab:complexity}
\begin{tabular}{l|*{6}{c} }
$\sigma$ & $\Cred{\sigma}$ & $\Skept{\sigma}$ & $\Ver{\sigma}$ & $\Exists{\sigma}$ & $\Exists{\sigma}^{\neg\emptyset}$ & $\Enum{\sigma}$\\
\hline
%$\cf$ & in $L$ & trivial & in $L$ \\
%$\naive$ & in $L$ & in $L$ & in $L$ \\
%\hline
%$\adm$ & $\NP$-c & trivial & in $L$ \\
$\com$ & $\NP$-c & $\P$-c & in $\L$ & trivial & $\NP$-c & $\nOP$ \\
$\prf$ & $\NP$-c & $\PiP{2}$-c & $\coNP$-c & trivial & $\NP$-c & $\nOP$ \\
$\stb$ & $\NP$-c & $\coNP$-c & in $\L$ & $\NP$-c & $\NP$-c & $\nOP$ \\
$\grd$ & $\P$-c & $\P$-c & $\P$-c & trivial & in $\L$ & in $\DelayP$ \\
\hline
$\stage$ & $\SigmaP{2}$-c & $\PiP{2}$-c & $\coNP$-c & trivial & in $\L$ & $\nOP$ \\
$\sem$ & $\SigmaP{2}$-c & $\PiP{2}$-c & $\coNP$-c & trivial & $\NP$-c & $\nOP$ \\
$\ideal$ & in $\ThetaP{2}$ & in $\ThetaP{2}$ & in $\ThetaP{2}$ & trivial & in $\ThetaP{2}$ & $\nOP$ \\
%$\rbg$ & $\NP$-c & $\coNP$-c & $\P$-c \\
\end{tabular}
\renewcommand{\arraystretch}{1.0}
\end{table}

%%% Local Variables:
%%% TeX-master: "aij-iccma17-report"
%%% End:

%\input{format}
\section{Format of ICCMA'17}
\label{sec:format}
%[{\bf TODO: Tasks, Tracks, Scoring system, verification os answers ...} ]
%[{\bf TODO: example of tasks? see 3.4}]

This section presents the main design of the competition. The competition is organized into tracks, which are divided into tasks.
Two sub-sections are devoted to their definitions. A third sub-section then presents  the scoring system, which changed from ICCMA'15 in order to focus more on correctness of answers.
Related to this issue, a fourth sub-section outlines how we verified correctness of answers.
Finally, information about I/O requirements is given.

%%
%% Put In Previous Section.
%%
%%  The competition features seven main tracks, one for each semantics defined in Section~\ref{sec:back}. Each of these tracks is composed of 4 (resp. 2 for single-status semantics) tasks, one for each reasoning task. A special track, Dung's Triathlon, will be conducted in order to have a reasoning task including several semantics.

%% Each solver participating in the competition can support, i.e. compete in, an arbitrary set of tasks. If a solver supports all tasks of a track, it also participates in the track.

\subsection{Tasks}
\label{sec:format:tasks}

A \emph{task} is a reasoning problem under a particular semantics.
%while all tasks for a particular semantics constitute a \emph{track}.
We consider the semantics $\com$, $\prf$, $\stb$, and $\grd$ which have already
been employed in the first edition,
and additionally the semantics $\sem$, $\stg$, and $\ideal$; 
the motivation to add these three semantics is due to the fact that their complexity
differs from the semantics already considered. 
%which are among the most prominent semantics of AFs.
Following ICCMA'15 we consider four different problems:

\begin{description}
	\item[$\dc$-$\sigma$:] \Given $\AFC$ and $a \in \args$, \decide whether $a$ is credulously accepted in $\F$ under $\sigma$,
	\item[$\ds$-$\sigma$:] \Given $\AFC$ and $a \in \args$, \decide whether $a$ is skeptically accepted in $\F$ under $\sigma$,
	\item[$\se$-$\sigma$:] \Given $\AFC$, \return some set $E\subseteq\args$ that is a $\sigma$-extension of $\F$,
	\item[$\ee$-$\sigma$:] \Given $\AFC$, \Enumerate all sets $E\subseteq\args$ that are $\sigma$-extensions of $\F$,
\end{description}
%
% \begin{itemize}
% \item $\se$: given an AF, compute one extension
% \item $\ee$: given an AF, compute all extensions
% \item $\dc$: given an AF and argument $a$, check whether $a$ is credulously accepted
% \item $\ds$: given an AF and argument $a$, check whether $a$ is skeptically accepted
% \end{itemize}
%
for the seven semantics $\sigma \in \{\com,\pr,\st,\sst,\stg,\gr,\id\}$. 

For single-status semantics ($\gr$ and $\id$) some problems collapse,
i.e.\ $\se$ and $\ee$ require to compute the unique extension; and 
$\dc$ and $\ds$ are equivalent.
Thus, for $\gr$ and $\id$ only the problems %consitute a track.
$\se$ and $\dc$
are considered. At this point, we also recall the well known fact that 
$\ds$-$\com$ coincides with $\dc$-$\gr$ and $\dc$-$\pr$ coincides with $\dc$-$\com$.

The combination of problems with semantics amounts to a total number of $24$ tasks.

%\todo[inline]{TL: should we present the task groups here? TL: no}

\subsection{Tracks}
\label{sec:format:tracks}

All tasks for a particular semantics constitute a \emph{track}.
Therefore, there is one track for each semantics.

Moreover, the competition features an eighth special track, the Dung's Triathlon.
It is named after Phan Minh Dung, and involves enumerating three of the main semantics
(grounded, stable, and preferred)
from his seminal paper~\citep{Dung:1995}.
The aim of this track is to evaluate solvers also with respect to their capability of exploiting interrelationships between different semantics. 
%Although grounded semantics is a single-status semantics,
%below we talke about grounded extensions for the sake of uniformity. 

More concretely, the problem to solve in this track is defined as follows:

\begin{description}
\item[$\dt$:] \Given $\AFC$, \Enumerate 
\begin{itemize}
\item all sets $E\subseteq\args$ that are $\gr$-extensions\footnote{% 
Although grounded semantics is a single-status semantics,
we treat it here like a multi-status semantics for the sake of uniformity.} 
of $\F$, followed by
\item all sets $E\subseteq\args$ that are $\st$-extensions of $\F$, followed by
\item all sets $E\subseteq\args$ that are $\pr$-extensions of $\F$.
\end{itemize}
\end{description}

%The winner of each track will be awarded.

\subsection{Scoring system}

Each solver can  compete in an arbitrary set of tasks. If a solver supports all tasks of a track, it also participates in the track.

To compute the score for a solver, we start by defining the number of points a solver can get for each instance: 
%For each instance, a solver gets
\begin{itemize}
\item $1$ point, if it delivers a \textit{correct} result;
\item $-5$ points, if it delivers an \textit{incorrect} result; or
\item $0$ points otherwise.
\end{itemize}

The precise understanding of what is a \textit{correct}, or an \textit{incorrect}, answer will be given in the next sub-section. Here, we focus on explaining how the solvers are ranked.

But before going into these details, we would like to stress a difference to ICCMA'15: in this edition wrong answers are penalized, while in ICCMA'15 they were treated as being neither correct nor incorrect, and got $0$ points. The objective, as already stated before, is to put focus on solvers' correctness.

The \textit{score} of a solver for a particular task is the sum
of points over all instances.
The ranking of solvers for a task is then based on the scores in descending order.
Ties between solvers with the same score are broken by the total time it took the solver to return correct results.

The ranking of solvers for a track is based on the sum of scores over all tasks
of the track, where each task is guaranteed to have the same impact on the evaluation of the track by all having the same number of instances (see Section~\ref{sec:bench-sel} for details about the number of instances). Again, ties are broken by the total time it took the solver to return correct results.

As far as the Dung's triathlon in concerned, scoring and ranking follow the same method as for the single tasks.

%The winner of each track is awarded.

\subsection{Verification of answers}

In this sub-section we discuss how the solvers' answers have been verified.
Before going into the details,
in the following we precisely define the concepts of \textit{correct} and \textit{incorrect} answers:

\begin{itemize}
\item
$\dc$-$\sigma$ (resp.\ $\ds$-$\sigma$): 
if the queried argument is credulously (resp.\ skeptically) accepted in the given AF under $\sigma$,
the result is
\textit{correct} if it is $\yes$ and
\textit{incorrect} if it is $\no$;
if the queried argument is not credulously (resp.\ not skeptically) accepted in the given AF under $\sigma$,
the result is 
\textit{correct} if it is $\no$ and
\textit{incorrect} if it is $\yes$.
%if it is any other than the correct one (?).
\item
$\se$-$\sigma$: the result is \textit{correct} if
it is a $\sigma$-extension of the given AF and
\textit{incorrect} if it is a set of arguments that is not a  $\sigma$-extension
of the given AF.
If the given AF has no $\sigma$-extensions,
then the result is \emph{correct} if it is \texttt{NO}
and \emph{incorrect}
if it is any set of arguments.
\item
%Finally, 
$\ee$-$\sigma$: the result is \textit{correct}
if it is the set of all $\sigma$-extensions of the given AF and
\textit{incorrect} if it contains a set of arguments that is not a $\sigma$-extension of the given AF.
\item
$\dt$: the result is \emph{correct} if it is the set of all $\gr$-extensions,
followed by the set of all $\st$-extensions,
followed by the set of all $\pr$-extensions, and
\emph{incorrect}
if the first set contains a set of arguments that is not the $\gr$-extension,
the second set contains a set of arguments that is not a $\st$-extension, or
the third set contains a set of arguments that is not a $\pr$-extension.
\end{itemize}

Intuitively, a result is neither correct nor incorrect
(and therefore gets $0$ points)
if
$(i)$ it is empty (e.g.\ the timeout was reached without answer) or
$(ii)$ it is not parsable with respect to the required output format (e.g.\ due to some unexpected error message).
For $\ee$-$\sigma$ there is also the case that the result $(iii)$
contains $\sigma$-extensions, but not all of them.
%
%For the Dung's triathlon, the result of 
Case $(iii)$ applies also to the Dung's triathlon, recursively on the three sub-problems.

To verify the correctness of results,
we employ the following checking procedure.
First, we generate reference solutions
by running
ASPARTIX-D~\citep{EglyGW10,GagglMRWW15},
%\todo{reason why we used ASPARTIX-D}
extended by the encodings for the new semantics,\footnote{%
The ICCMA'15 version can be found at \url{https://iccl.inf.tu-dresden.de/web/Sarah_Alice_Gaggl/ASPARTIX-D};
the additional encodings are available at \url{https://www.dbai.tuwien.ac.at/proj/argumentation/systempage}.
The choice of this particular solver is due to
(i) its declarative nature,
(ii) its good results in 2015,
(iii) the fact that it is ``third-party'' in 2017 given that it does not participate, and
(iv)
%the familiarity of the organizers with the system.}
its reputation in the community
(``state of the
art of ASP-based solvers''~\cite{BistarelliRS14}).}
on all benchmarks selected for the competition (see Section~\ref{sec:bench-sel}).
For the instances that ASPARTIX-D is able to solve,
we compare the solutions with the reference solutions in order to
assess correctness.
For the other instances, 
we then use dedicated ASP encodings to check single extensions
%as well as existence of stable extensions
(available at \url{http://argumentationcompetition.org/2017/SE_encodings.zip})
to verify answers for the $\se$ and $\ee$ reasoning problems.
These ASP encodings are directly derived from
the ASPARTIX encodings -- the part for guessing an extension
is replaced by the given extension which is to be checked.
For the other tasks as well as these cases where also checking all single extensions was not feasible,
we then consider the solution provided by the majority of solvers as correct
(other solutions could always be checked to be wrong though).
The detailed number of uniquely solved instances by a certain solver will be given in Section\ref{sec:res}, also including the number
of instances for each track and solver which could not be verified. In total only approx. $0.1\%$ out of the $105350$ solutions could not be verified and thus have been rated
with $1$ point. In none of the tracks these had an influence on the ranking of the solvers.

%The number of instances uniquely solved by a certain solver without being formally checked
%are indicated in the detailed results in Appendix~A. %\ref{sec:app1}.
%\todo{check if we really do this.}

\subsection{Solver requirements}

Participant systems were required to support the same input-output format as used in 2015. %Input programs are compli-
%ant with the ASP-Core-2 standard, and the expected system output depends on the kind of a problem, i.e. Decision, Query,
%or Optimization. 
Details on the input and output formats can be found in~\citep{iccma17-solverreq}. % \url{http://www.argumentationcompetition.org/2017/SolverRequirements.pdf}.

%%% Local Variables:
%%% TeX-master: "aij-iccma17-report"
%%% End:

%% \input{format}
%\input{benchmarks}
\section{Benchmark Suite}\label{sec:bench}
%% [{\bf TODO: Divided into two parts: from the 1st event, and new. Descriptions for the new; what setting have been used for the old. Stress variety (SW: done at end of section).} ]

In this section we outline the benchmark suite available for ICCMA'17, which has been the starting point for the selection phase (described in the next section). The suite is composed both by domains employed in ICCMA'15 and by new domains, the latter received in response to a dedicated {\sl call for benchmarks}. The next two sub-sections are devoted to the presentation of these two sets of domains. 

\subsection{Previous domains}

ICCMA'15 introduced three new AF generators, called GroundedGenerator, StableGenerator, and SccGenerator,
each of them aiming to produce challenging AFs addressing
certain aspects of computational difficulty.
They have been implemented~\citep{CeruttiOSTV14} and employed to generate
the AFs that constituted the benchmark suite of ICCMA'15.
In the following, we briefly describe the generators,
but refer to~\citep{ThimmV17} for more details.

\begin{description}
\item[GroundedGenerator]
This generator aims at producing AFs with large grounded extensions.
It takes the number of arguments $n$ and probability $\probAttacks$
as parameters, linearly orders the arguments and adds an attack
from argument $a$ to argument $b$ in case $a < b$ with probability $\probAttacks$.
%Finally, it connects isolated arguments to one of the components obtained so far.
Finally, it adds random attacks between the arguments not yet connected and
the graph component obtained in the first part.
\item[SccGenerator]
This generator aims at producing AFs such that the graph
features many Strongly Connected Components (SCCs).
It first partitions the arguments (the number of which is given by parameter $n$)
into $\nSCCs$ (also given as parameter) components which are linearly ordered.
Within each component, attacks between any pair of arguments are added with
probability given by parameter $\innerAttackProb$.
Among arguments of different components, attacks are added with probability
given by parameter $\outerAttackProb$, but under the condition that the component
of the attacking arguments is ranked lower with respect to the linear order on components
than the component of the attacked argument.
\item[StableGenerator]
This generator aims at producing AFs with a large number of stable extensions.
It first identifies a set of arguments to form an acyclic subgraph of the AF and, consequently,
to contain the grounded extension.
Among the other arguments, subsets are iteratively singled out to form stable extensions
by attacking all other arguments.
Besides the parameter $n$ for the number of arguments,
the algorithm is further guided by the parameters
$\texttt{minNumExtensions}$,
$\texttt{maxNumExtensions}$,
$\texttt{minSizeOfExtensions}$,
$\texttt{maxSizeOfExtensions}$,
$\texttt{minSizeOfGroundedExtension}$, and \\
$\texttt{maxSizeOfGroundedExtension}$,
which determine heuristic values for
the minimum and maximum number of stable extensions,
the minimum and maximum size of stable extensions, and
the minimum and maximum size of grounded extensions, respectively.
\end{description}

\subsection{New Domains}
\label{subsec:newd}

ICCMA'17
%has taken
takes
advantage, for the first time, of a dedicated {\sl call for benchmarks}, which is customary in other competitions. The goal of this call has been to enlarge the set of domains that are considered in the competition, and thus possibly having a more heterogeneous set of benchmarks %(e.g., random, crafted, application-oriented) 
in the evaluation.  Contributors were asked to provide an instance set for the benchmark they submitted, and/or an instance generator, possibly with an indication about %which instances are expected to be "hard", or "easy". 
the estimated difficulty of the instances.
We have received 6 submissions, among them AF generators
%(AdmBuster, AFBenchGen2, SemBuster)
as well as concrete sets of AFs, thus meeting our desiderata to have a heterogeneous set of benchmarks, i.e.\ random, crafted, and application-oriented, as a benchmark suite of the competition. \\

%(ABA2AF, Planning2AF, Traffic).
%Six sets of new benchmarks, that are listed in the following along with their contributors and number of instances, have been submitted to ICCMA'17:

%[{\bf TODO: Add short description for each domains.}]
\noindent
Herewith we briefly describe the domains that were submitted:

\begin{description}
\item[``ABA2AF''] by Tuomo Lehtonen (\uni{} of Helsinki, Finland), Johannes P. Wallner (TU Wien, Austria), Matti J\"arvisalo (\uni{} of Helsinki, Finland), are assumption-based argumentation (ABA) 
benchmarks translated to AFs. ABA problems are one of the prevalent forms of structured argumentation in which, differently from AFs, the internal structure of arguments is made explicit
through derivations from more basic structure \citep{DBLP:journals/argcom/Toni14}. The translation employed is described in~\citep{LehtonenWJ17}. The original ABA set contains randomly generated cyclic and acyclic ABAs that, after a selection from the authors, resulted in a total of 426 instances.
\item[AdmBuster] by Martin Caminada (Cardiff \uni{}, UK), Mikolaj Podlaszewski (Talkwalker), is a crafted benchmark example for (strong) admissibility. It is made of a fixed structure composed of 4 sets of arguments and predetermined sets of attacks. The number $n$ is a parameter of the generator. Two ``starting'' and ``terminal'' sets are composed of only one element, one having only outgoing edges and the other only incoming edges.  The two ``intermediate'' sets have cardinality $n-2$, and their attack relations are constructed in order to have only one complete labelling. Details can be found in~\citep{Caminada14}.
At the competition, 13 instances generated with different values of $n$ are considered.
\item[AFBenchGen2] by Federico Cerutti (Cardiff \uni{}, UK), Mauro Vallati (\uni{} of Huddersfield, UK), Massimiliano Giacomin (\uni{} of Brescia, Italy), is a generator of random AFs of three different graph classes, with a configurable number of arguments~\citep{CeruttiGV16}. The three classes correspond to Erd\"os-R\'enyi~\citep{ErdosR59}, which selects attacks randomly, Watts-Strogatz~\citep{WattsS98}, which aims for a {\sl small-world} topology of networks being not completely random nor regular, and Barabasi-Albert~\citep{BarabasiA99} for large networks. For each graph class, the generator takes the number of arguments $n$ as parameter. $1400$ instances have been generated, of which $500$ are from Barabasi-Albert class, $500$ are from Erd\"os-R\'enyi class, and $400$ are from Watts-Strogatz class. In the following, we provide some more details for such three classes: 
\begin{itemize}
\item Barabasi-Albert: This graph class is motivated by a common property of many large networks, i.e. that
the node connectivities follow a scale-free power-law distribution.
Therefore, the generator
of a Barabasi-Albert graph iteratively connects a new node
by preferring sites that are already well connected.
In addition, a postprocessing procedure adds attacks in order
to ensure a certain amount of cycles in the graph.
This amount is controlled by the parameter $\probCycles$.
An attack is added as long as the number of SCCs of the AF
is higher than $n \cdot (1 - \probCycles)$.
\item Erd\"os-R\'enyi: Graphs are generated by randomly selecting attacks between arguments. For any two distinct arguments, the probability of an attack between them is given by the parameter $\probAttacks$. The direction of the attack is chosen randomly.
%parameter
\item Watts-Strogatz:
First, a ring of $n$ arguments is generated where each argument is connected
to its $k$ (a parameter of the generator)
nearest neighbors in the ring.
Then, each argument is connected to the remaining arguments with a probability
$\beta$ (another parameter of the generator).
Finally, as in Barabasi-Albert, random
attacks are added as long as the number of SCCs of the AF
is higher than $n \cdot (1 - \probCycles)$.
%where $\probCycles$ is another parameter.
\end{itemize}
\item[``Planning2AF''] by Federico Cerutti (Cardiff \uni{}, UK), Massimiliano Giacomin (\uni{} of Brescia, Italy), Mauro Vallati (\uni{} of Huddersfield, UK), are AFs obtained from translating the well-known Blocksworld and Ferry planning domains. Each planning instance is first encoded as a propositional formula, by using the method in~\citep{SiderisD10}; then, each clause is transformed into a material implication; and, finally, to each material implication the transformation in~\citep{WynerBDC15} is applied. This domain comprises 385 instances. 
\item[SemBuster] by Martin Caminada (Cardiff \uni{}, UK), Bart Verheij (Rijks\-universiteit Groningen, Netherlands), is a crafted benchmark example for semi-stable semantics. It has a fixed structure composed by $3$ sets of arguments of equal cardinality, and predetermined sets of attacks. Given a parameter $n$, attack relations are defined in a way that each instance has exactly $n+1$ complete labellings that correspond also to preferred labellings, but only one among those corresponds to a semi-stable extension. Details can be found in~\citep{CaminadaV10}. %% One set Two ``starting'' and ``terminal'' sets are made of only one node, one having only outgoing edges and another only incoming. The two ``intermediate'' set set have cardinality $n$ and their attack relations are constructed in order to have only one complete labelling. Details can be found in~\citep{Caminada14}. 
At the competition, 16 instances generated with different values of $n$ are considered.
\item[``Traffic''] by Martin Diller (TU Wien, Austria), are graphs obtained from real world traffic networks data available at \url{https://transitfeeds.com/} expressed as AFs. Given a graph, the corresponding AF contains the same set of vertices as the graph, and the attack relation is defined as follows: Given an existing edge, and a probability for the attack of being symmetric, the generator decides whether there are both attacks, or randomly selects the attack.  A total of 600 instances are provided, $200$ for each of the probabilities $0.2$, $0.5$, and $0.8$.
Although these instances do not directly relate to argumentation applications, we decided to include them in the competition, in order to have an orthogonal class of sparse graphs with  certain structural features.
\end{description}

More detailed descriptions for such domains can be found in the ICCMA'17 home page at~\citep{iccma17-solverdescr}.\\

%With the generators, we produced instances aiming at covering a possibly broad range of difficulty. [DETAILS?] \\

Table~\ref{tab:collection} gives details on the
collected benchmarks by stating, for each domain, the number of
instances as well as the parameters used for generating the instances.
If the benchmark submission consists of a set of instances,
we simply considered them all.
For domains emerging from submissions of benchmark generators,
we produced instances randomly with the aim of
covering a possibly broad range of difficulty.
The exact parameters used for generating the instances
can be read off from Table~\ref{tab:collection}.
In some cases, parameters are chosen randomly from an interval.
This is denoted by $\random[a,b]$.
In other cases, all values in a set are considered,
denoted by $\{v_1, v_2, \dots, v_n\}$.

Thus, the benchmark suite of ICCMA'17 is finally composed of 3990 instances over 11 domains.
This yields a healthy mixture of benchmarks ranging from random instances to more
structured AFs which are either handcrafted or instantiated from different application domains.

\afterpage{%
    \clearpage% Flush earlier floats (otherwise order might not be correct)
    \thispagestyle{empty}% empty page style (?)
    \begin{landscape}% Landscape page
        \centering % Center table
{
\small
\setlength\tabcolsep{3pt}
\begin{tabularx}{\linewidth}{|l|r|L|}
\hline
{\bf Domain} &     {\bf Inst.} & {\bf Parameters} \\
\hline
ABA2AF &               426 & all submitted instances \\
\hline
AdmBuster &            13  & $n$ in $\{1000, 2000, 4000, \dots, 10000, 20000, 50000, 100000, 200000, 500000, 1000000, 2000000\}$ \\
\hline
Barabasi-Albert &      500 & $5$ random instances for each ($n$, \probCycles) in $\{20, 40, \dots, 200\} \times \{0, 0.1, \dots, 0.9\}$ \\
\hline
Erd\"os-R\'enyi &      500 & $10$ random instances for each ($n$, \probAttacks) in $\{100, 200, \dots, 500\} \times \{0.1, 0.2, \dots, 1.0 \}$  \\
\hline
GroundedGenerator &    50  & $n = \random[100,1500]$; $10$ random instances for each \probAttacks{} in $\{0.01, 0.02, \dots, 0.05\}$ \\
\hline
Planning2AF &          385 & all submitted instances\\
\hline
SccGenerator &         600 & $n=\random[100,1500]$; $\nSCCs=\random[1,50]$; $25$ random instances for each (\innerAttackProb, \outerAttackProb) in $\{0.3,0.4,\dots,0.7\} \times \{0.05,0.1,0.15,0.2\}$.

$n=\random[5000,10000]$; no.\ SCCs $\random[40,50]$; $5$ random instances for each (\innerAttackProb, \outerAttackProb) in $\{0.3,0.4,\dots,0.7\} \times \{0.05,0.1,0.15,0.2\}$. \\
\hline
SemBuster &            16  & $n$ in $\{60, 150, 300, 600, \dots, 1800, 2400, 3000, 3600, 4200, 4800, 5400, 6000, 7500\}$ \\
\hline
StableGenerator &      500 & $n=\random[100,800]$; $500$ random instances with parameters
$\texttt{minNumExtensions}=5$,
$\texttt{maxNumExtensions}=30$,
$\texttt{minSizeOfExtensions}=5$,
$\texttt{maxSizeOfExtensions}=40$,
$\texttt{minSizeOfGroundedExtension}=5$,
$\texttt{maxSizeOfGroundedExtension}=40$ \\
\hline
Traffic &              600 & all submitted instances \\ %($200$ for each attack probability in $\{0.2,0.5,0.8\}$)\\
\hline
Watts-Strogatz &       400 & ($n$, $k$, $\beta$, \probCycles) in $\{100,200,\dots,500\} \times  \{\log_2(n), 2 \cdot \log_2(n), 3 \cdot \log_2(n), 4 \cdot \log_2(n)\} \times \{0.1, 0.3, \dots, 0.9\} \times \{0.1, 0.3, 0.5, 0.7\}$ \\
\hline
\end{tabularx}
}
\captionof{table}{Description of (generated) benchmarks constituting the benchmark suite.}
\label{tab:collection}
    \end{landscape}
    \clearpage% Flush page
}

%%% Local Variables:
%%% TeX-master: "aij-iccma17-report"
%%% End:

%\input{benchmarks-selection}
\section{Benchmark Selection}\label{sec:bench-sel}
%[{\bf TODO: Structure from the note.}]

With the benchmark suite described in the previous section,
the goal of this phase is to select the instances
%``for each domain''
that are indeed run in the competition.
%Then, a set of instances is collected (or generated)
%for each domain.
%These instances are subsequently
In order to guide this selection,
the instances are
classified into hardness categories according to the performance
of a set of solvers from the previous competition.
Finally, the instances to be run at the competition are selected
based on this classification,
following a predefined distribution over hardness categories.

As the tasks of the competition span over a wide range of complexity (cf.\ Table~\ref{tab:complexity}),
a single set of benchmarks for the whole competition
might not be suitable.
Therefore we aim to adjust the benchmarks to the complexity of the tasks,
while keeping the total amount of different benchmarks manageable.
%assign each task group defined in Section~\ref{sec:format:tasks}
%its own set of benchmarks.
%As each group of tasks has its ,
To this end,
we introduce a grouping of tasks according to their difficulty, % of the respective tasks.
such that each of the groups gets a dedicated set of benchmarks.
The classification into groups A to E is based on known complexity results and corroborated
by the analysis of the
%percentage of solved instances in
results of ICCMA'15.
The applied grouping is the following:

%Tasks are 

\begin{description}
\item[{\normalfont Group A:}] $\ds$-$\prf$, $\ee$-$\prf$, $\ee$-$\com$.
\item[{\normalfont Group B:}] $\dc$-$\stb$, $\ds$-$\stb$, $\ee$-$\stb$, $\se$-$\stb$, $\dc$-$\prf$, $\se$-$\prf$, $\dc$-$\com$.
\item[{\normalfont Group C:}] $\ds$-$\com$, $\se$-$\com$, $\dc$-$\grd$, $\se$-$\grd$.
\item[{\normalfont Group D:}] $\dc$-$\ideal$, $\se$-$\ideal$.
\item[{\normalfont Group E:}] $\dc$-$\sem$, $\ds$-$\sem$, $\ee$-$\sem$, $\se$-$\sem$, $\dc$-$\stage$, $\ds$-$\stage$, $\ee$-$\stage$, $\se$-$\stage$.
\end{description} 

Hence,
the classification and selection has to be done for each group.
However, since there are no reference solvers for the tasks of groups D and E
(these are the ones newly employed in this edition),
we do not perform a dedicated selection for these groups.
Instead, the tasks of these groups
are assigned the same benchmark set as group A,
because they are of high complexity and
we expect solvers to be less mature since ICCMA'15 did not feature these tasks yet.

%Note that groups D and E include the newly employed semantics.
%Each group A to C gets its own set of benchmarks,
%while for groups D and E the same set of benchmarks as for group A is used.

The following sub-sections present
%1) how instances have been collected,
%1) how 2015 solvers have been selected,
how instances are classified,
%3)
how instances are selected,
and, finally, how the query arguments
for the $\dc$ and $\ds$ tasks are selected.

\subsection{Benchmark Classification}
\label{subsec:bench-class}
%In the next step the collected instances
%have to be
%This 
%are
%classified with respect to
%their expected level of difficulty.

To classify the hardness of instances,
%This selection in
competitions in other research fields such as SAT \citep{sat2009,JarvisaloBRS12,BalintBJS15}, ASP \citep{GebserMR17}, and IPC for automated planning \citep{VallatiCGMRS15},
employ best solvers from the most recent competition in the series.
%\todo{references here? SW: Yes}
We follow this idea by also doing a classification of benchmarks
based on the performance of solvers from ICCMA'15.
However, in ICCMA the situation shows two significant differences.
On the one hand, the number of tasks and tracks employed in ICCMA
(significantly) exceeds the number of tasks and tracks in other competitions.
On the other hand, ICCMA'17 features new semantics (and, consequently, new tasks and tracks),
so no reference results are at disposal.

Due to the second point,
the option of selecting the best solvers from the previous edition for each task is not feasible.
But, even considering only tasks which are being conducted for the second time, % 14 of it
this option would lead to a very high number of solvers to run for the classification.
Instead, %to ``group'' tasks, and
%Instead of classifying instances for every single tasks,
we identify
%to select
``representative'' tasks for each task group A, B, and C
%containing tasks 
which have also been conducted in ICCMA'15.
Moreover, as mentioned earlier,
we
abstain from
%pass on
classifying instances for tasks in groups D and E,
but merge these tasks with the ones from group A
and employ the same set of benchmarks.
We identify the following representative tasks
which will be used for classification:

\begin{itemize}
\item Group A: $\ee$-$\prf$
\item Group B: $\ee$-$\stb$
\item Group C: $\se$-$\grd$
\end{itemize}

All task groups contain enumeration as well as decision tasks.
We select enumeration tasks as representative,
as the performance of solvers on decision tasks
highly depends on the argument for which acceptance is to be decided.
Therefore, enumeration tasks can give a better estimate
of the difficulty of instances.

%\subsubsection{(Best) Solvers selection}

\paragraph{(Best) Solver selection}
For each representative task we aim to select
``representative'' solvers from ICCMA'15,
to get a proper estimate of the instances' hardness.
Solvers to run for each group are thus selected by
(i) considering best performing solvers from 2015 for the tasks, and
(ii) ensuring that the selected solvers are based on different solving approaches, in order not to have results biased through a single solving approach.
The following solvers from ICCMA'15 are selected
(see~\citep{ThimmV15}
for system descriptions):

\begin{itemize}
\item Group A: Cegartix, CoQuiAAS, Aspartix-V    % 1, 3, 4
\item Group B: Aspartix-D, ArgSemSAT, ConArg     % 1, 2, 5
\item Group C: CoQuiAAS, LabSATSolver, ArgSemSAT % 1, 3, 4
\end{itemize}

Both Cegartix~\citep{DvorakJWW14} and ArgSemSAT~\citep{CeruttiGV14a} implement (iterative) SAT based approaches;
CoQuiAAS~\citep{LagniezLM15} makes use of Partial Max-SAT;
Aspartix-V and Aspartix-D~\citep{EglyGW10,GagglMRWW15} employ a translation to ASP;
ConArg~\citep{BistarelliS11} is based on Constraint Programming; and
LabSATSolver~\citep{BeierleBP15} implements a direct approach
%the algorithm by~\cite{ModgilC09}
(for $\se$-$\grd$).
All of the solvers have been among the top 5 solvers in the respective
tasks in ICCMA'15.
Hence, the selection is in line with (i) and (ii).

%\todo{justify the selection based on (i) and (ii),
%i.e.\ rankings and techniques of selected solvers}

\paragraph{Hardness categories}
The obtained performance results of the 3 selected solvers in each group are then taken to classify instances
into hardness categories by picking the upmost category such that the
following conditions apply:

\begin{description}
\item[\textbf{[very easy]}] Instances completed by all systems in less than 6 seconds solving time.
\item[\textbf{[easy]}] Instances completed by all systems in less than 60 seconds solving time.
\item[\textbf{[medium]}] Instances completed by all systems in less than 10 minutes solving time.
\item[\textbf{[hard]}] Instances completed by at least one system in 20 minutes (twice the timeout) solving time.
\item[\textbf{[too hard]}] Instances such that none of the systems finished solving in 20 minutes.
\end{description}

The results of the classification are summarized in
Tables~\ref{tab:classificationA},~\ref{tab:classificationB}, and~\ref{tab:classificationC}
for task groups A, B, and C\footnote{AdmBuster domain in Table~\ref{tab:classificationC} contains two additional instances with $n$ of $1500000$ and $2500000$.
}, respectively.
%in Table~\ref{tab:classification}.
It can be seen that almost every combination of
domain and difficulty category contains instances.
Only for the ``too hard'' category we are not
able to obtain instances for every domain
(even for no domain for task group C).
If at least two of the representative solvers crashes for an instance,
the instance is not classified (abbreviated by ``n.\ c." in the tables),
and therefore not considered for selection.

% details about this:
% if one gave no result,
% we took the average of the runtime of the other two as its runtime
% if two or more gave no result,
% we said ``not classified''

%\todo{is this interesting? if yes, add results of classification for B and C,
%\url{benchmarks/results_first_run/Group{A,B,C}.statistics.new}}

\begin{table}
\caption{Classification results for task group A.}
\label{tab:classificationA}
{\small
\begin{tabular}{l|*{7}{r|}}
A: $\ee$-$\prf$                & total  & very easy   & easy  &  medium  & hard  & too hard & n.\ c. \\
\hline
ABA2AF                & 426    & 381         & 19    &  16      & 10    &  0      &  0 \\
AdmBuster             & 13     & 4           & 3     &  2       & 4     &  0      &  0 \\
Barabasi-Albert       & 500    & 267         & 25    &  20      & 42    &  145    &  1 \\
Erd\"os-R\'enyi       & 500    & 180         & 109   &  43      & 46    &  122    &  0 \\
Watts-Strogatz        & 400    & 264         & 28    &  10      & 12    &  86     &  0 \\
GroundedGenerator     & 50     & 9           & 8     &  6       & 27    &  0      &  0 \\
Planning2AF           & 385    & 95          & 35    &  34      & 187   &  33     &  1 \\
SccGenerator          & 600    & 398         & 78    &  44      & 79    &  0      &  1 \\
SemBuster             & 16     & 2           & 1     &  3       & 9     &  1      &  0 \\
StableGenerator       & 500    & 260         & 34    &  24      & 182   &  0      &  0 \\
Traffic               & 600    & 164         & 11    &  11      & 284   &  127    &  3 \\
\hline
Total                 & 3990   & 2024        & 351   &  213     & 882   &  514    &  6
\end{tabular}
}
\end{table}

\begin{table}
\caption{Classification results for task group B.}
\label{tab:classificationB}
{\small
\begin{tabular}{l|*{7}{r|}}
B: $\ee$-$\stb$    & total  & very easy   & easy  &  medium  & hard  & too hard & n.\ c. \\
\hline
ABA2AF             &  426    &  407       &  18    &  1       &  0     &  0        &  0 \\
AdmBuster          &  13     &  9         &  1     &  1       &  2     &  0        &  0 \\
Barabasi-Albert    &  500    &  262       &  19    &  5       &  122   &  92       &  0 \\
Erd\"os-R\'enyi    &  500    &  247       &  102   &  31      &  49    &  71       &  0 \\
Watts-Strogatz     &  400    &  201       &  39    &  26      &  76    &  58       &  0 \\
GroundedGenerator  &  50     &  19        &  25    &  5       &  1     &  0        &  0 \\
Planning2AF        &  385    &  117       &  5     &  5       &  159   &  99       &  0 \\
SccGenerator       &  600    &  248       &  66    &  65      &  218   &  3        &  0 \\
SemBuster          &  16     &  6         &  6     &  4       &  0     &  0        &  0 \\
StableGenerator    &  500    &  225       &  26    &  37      &  73    &  139      &  0 \\
Traffic            &  600    &  275       &  7     &  2       &  70    &  245      &  1 \\
\hline
Total              &  3990   &  2016      &  314   &  182     &  770   &  707      &  1
\end{tabular}
}
\end{table}

\begin{table}
\caption{Classification results for task group C.}
\label{tab:classificationC}
{\small
\begin{tabular}{l|*{7}{r|}}
C: $\se$-$\grd$    & total  & very easy   & easy  &  medium  & hard  & too hard & n.\ c. \\
\hline

ABA2AF             &  426    &  404       &  21    &  1       &  0     &  0        &  0 \\
AdmBuster          &  15
                             &  7         &  1     &  1       &  6     &  0        &  0 \\
Barabasi-Albert    &  500    &  500       &  0     &  0       &  0     &  0        &  0 \\
Erd\"os-R\'enyi    &  500    &  424       &  44    &  11      &  21    &  0        &  0 \\
Watts-Strogatz     &  400    &  296       &  36    &  21      &  47    &  0        &  0 \\
GroundedGenerator  &  50     &  20        &  25    &  1       &  4     &  0        &  0 \\
Planning2AF        &  385    &  359       &  23    &  3       &  0     &  0        &  0 \\
SccGenerator       &  600    &  485       &  84    &  31      &  0     &  0        &  0 \\
SemBuster          &  16     &  3         &  1     &  0       &  12    &  0        &  0 \\
StableGenerator    &  500    &  308       &  62    &  42      &  88    &  0        &  0 \\
Traffic            &  600    &  459       &  42    &  51      &  50    &  0        &  0 \\
\hline
Total              &  3992   &  3265      &  339   &  162     &  228   &  0        &  0
\end{tabular}
}
\end{table}

\subsection{Benchmark selection}
\label{subsec:bench-sel}

The final benchmark set 
for each task group
is made up of $350$ instances,
%Moreover,
%The distribution
distributed
%of these instances 
over the difficulty categories
%is
as follows:
\begin{itemize}
\item $50$ very easy,
\item $50$ easy,
\item $100$ medium,
\item $100$ hard,
\item $50$ too hard.
\end{itemize}

Due to the lack of ``too hard'' instances
for group C (cf.\ Table~\ref{tab:classificationC}),
the number of ``hard'' instances
is increased to $150$ there.

We aim for an even distribution
of benchmarks over levels of difficulty,
but also among domains.
Now, in order to select $n$ instances for a certain
task group and a certain class of difficulty,
we apply the following procedure:
for each domain $d$,
%we start with a
we are given the
set $I_d$ of instances and want to select a subset $S_d$ of these instances.
Now for each domain such that $I_d$ is non-empty,
we select one element of $I_d$ at random,
i.e.\ remove it from $I_d$ and add it to $S_d$.
We repeat this process until we have selected $n$ instances,
i.e.\ the sum over all $|S_d|$ is $n$.
In the last iteration,
when the number of domains where $I_d$ is non-empty
is higher than the number of instances that remains to be selected,
the domains to be chosen from are determined randomly.
%The procedure is described more rigorously in Algorithm~\ref{alg:selection}.
A more rigorous description of this procedure can be found at
\url{http://argumentationcompetition.org/2017/benchmark-selection-algorithm.pdf}.

%\begin{algorithm}[t]
%\caption{Algorithm for selecting instances based on classification}
%\label{alg:selection}
%\begin{algorithmic}[1]
%\small
%\Require $D$: set of domains.
%\Require $\{I_d\}_{d \in D}$: set of instances for each domain $d$.
%\Require $n$: desired number of instances.
%\Ensure $\{S_d\}_{d\in D}$: set of selected instances for each domain $d$.
%
%\State $S_d = \emptyset$ for each domain $d \in D$
%\State $D = \{d \in D \mid I_d \neq \emptyset\}$
%  
%\While{$\sum_{d\in D}{|I_d|} < n$}
%    \If{$n - \sum_{d\in D}{|I_d|} \geq |D|$}
%        \State $D' = D$
%    \Else
%        \State select $D' \subseteq D$ randomly such that $|D'| = n - \sum_{d\in D}{|I_d|}$
%    \EndIf
%    
%    \ForAll{$d \in D'$}
%       \State randomly select $i \in I_d$
%       \State $S_d = S_d \cup \{i\}$
%       \State $I_d = I_d \setminus \{i\}$
%    \EndFor
%
%    \State $D = \{d \in D \mid I_d \neq \emptyset\}$
%\EndWhile
%
%\State \Return $\{ S_d \}_{d\in D}$
%\end{algorithmic}
%\end{algorithm}

\begin{example}
\label{ex:selection}
Assume domains $D=\{\alpha,\beta,\gamma,\delta\}$
such that
we have
1 instance for domain $\alpha$,
2 for $\beta$,
4 for $\gamma$, and
11 for $\delta$,
i.e.\ $|S_\alpha|=1$, $|S_\beta|=2$, $|S_\gamma|=4$, and $|S_\delta|=11$.
Further assume that we want to select $n=10$ instances.
The selection algorithm
will return
all instances from $\alpha$ and $\beta$, $3$ instances from $\gamma$ and $\delta$, and
$1$ additional instance randomly selected from either $\gamma$ or $\delta$.
\end{example}

The numbers of selected instances for
every domain, task group, and difficulty category
can be read off from Table~\ref{tab:selection}.

\begin{table}
\centering
\setlength\tabcolsep{2.25pt}
\caption{Number of selected instances for each
task group, difficulty class, and domain,
where difficulty classes 1 to 5 stand for very easy, easy, medium, hard, and too hard, respectively. ``T'' indicates the total number of selected instances.}
\label{tab:selection}
{%\small\footnotesize
\footnotesize
\begin{tabular}{l|*{6}{r}|*{6}{r}|*{6}{r}}
Task group        & \multicolumn{6}{c|}{A} & \multicolumn{6}{c|}{B} & \multicolumn{6}{c}{C} \\
Difficulty class    & 1 & 2 & 3  & 4  & 5 &T& 1 & 2 & 3 & 4 & 5 &T& 1 & 2 & 3 & 4 & 5 & T\\
\hline
ABA2AF            & 5 & 5 & 12 & 10 & 0 &32& 5 & 5 & 1 & 0 & 0 &11& 5 & 6 & 1 & 0 & 0 & 12\\
AdmBuster         & 4 & 3 & 2  & 4  & 0 &13& 4 & 1 & 1 & 2 & 0 &8& 4 & 1 & 1 & 6 & 0 & 12\\
Barabasi-Albert   & 5 & 5 & 11 & 10 & 10 &41& 5 & 5 & 5 & 14 & 8 &37& 5 & 0 & 0 & 0 & 0 & 5\\
Erd\"os-R\'enyi   & 5 & 5 & 11 & 10 & 9 &40& 5 & 5 & 19 & 13 & 7 &49& 5 & 6 & 11 & 21 & 0 & 43\\
Watts-Strogatz    & 5 & 5 & 10 & 10 & 10 &40& 5 & 5 & 20 & 14 & 8 &52& 5 & 6 & 21 & 36 & 0 & 68\\
GroundedGenerator & 4 & 5 & 6  & 9  & 0 &24& 4 & 4 & 5 & 1 & 0 &14& 5 & 6 & 1 & 4 & 0 & 16\\
Planning2AF       & 5 & 6 & 12 & 10 & 10 &43& 5 & 5 & 5 & 14 & 8 &37& 5 & 6 & 3 & 0 & 0 & 14\\
SccGenerator      & 5 & 5 & 11 & 9  & 0 &30& 4 & 5 & 19 & 14 & 3 &45& 4 & 6 & 21 & 0 & 0 & 31\\
SemBuster         & 2 & 1 & 3  & 9  & 1 &16& 4 & 5 & 4 & 0 & 0 &13& 3 & 1 & 0 & 12 & 0 & 16\\
StableGenerator   & 5 & 5 & 11 & 9  & 0 &30& 4 & 5 & 19 & 14 & 8 &50& 4 & 6 & 20 & 35 & 0 & 65\\
Traffic           & 5 & 5 & 11 & 10 & 10 &41& 5 & 5 & 2 & 14 & 8 &34& 5 & 6 & 21 & 36 & 0 & 68\\
\hline
Total             & 50& 50& 100& 100 & 50 &350& 50 & 50 & 100 & 100 & 50 &350& 50 & 50 & 100 & 150 & 0 & 350\\
\end{tabular}
}
\end{table}

The instances for Dung's triathlon are selected based on
the classification for task group A,
but by a separate process.
That means that the numbers of instances
per domain coincide with group A,
but instances are not necessarily the same.

\paragraph{No stable extensions}
Semi-stable and stage extensions coincide with
stable extensions if at least one of the latter exists.
In this case, the complexity of the reasoning tasks drops to
the level of the corresponding tasks for stable semantics (cf.\ Table~\ref{tab:complexity}).
Therefore, in order to force solvers to deal with the
``full hardness'' of semi-stable and stage semantics,
we want to make sure that the selection for these
semantics contains a %considerable
sufficient
amount of benchmarks possessing
no stable extensions.
To this end, we checked the selected instances
on existence of stable extensions
by running ASPARTIX-D from ICCMA'15
(winning solver for all tasks involving stable semantics).
The numbers are shown in Table~\ref{tab:stable}:
for $22$ instances no answer is provided by ASPARTIX-D.
%We assess the share of instances
%without stable extensions to be sufficient.
We consider
 the number of instances without stable extensions (114) to be satisfactory.

\begin{table}[h]
\centering
%\caption{Share of instances without stable extensions.}
\caption{Analysis of the existence of stable extensions.}
\label{tab:stable}
%\begin{tabular}{l|r|r|r|r}
%hardness category & $\stb(F)\neq\emptyset$ & $\stb(F)=\emptyset$ & unknown & share\\
\begin{tabular}{l|r|r|r}
hardness category & $\stb(F)\neq\emptyset$ & $\stb(F)=\emptyset$ & unknown \\
\hline
very easy & 34 & 16 & 0 \\% &  $32 \%$ \\
easy & 34 & 16 & 0 \\%& $32 \%$ \\
medium & 60 & 40 & 0 \\%& $40 \%$ \\
hard & 56 & 33 & 11 \\%& $> 33 \%$ \\
too hard & 30 & 9 & 11 \\%& $> 18 \%$ \\
\hline
total & 214 & 114 & 22 \\%& $> 34 \%$ 
\end{tabular}
\end{table}

\subsection{Argument Selection}
\label{subsec:arg-sel}

Due to the joint evaluation of all tasks for a semantics,
making up a track,
the number of benchmarks has to be constant among the tasks.
Therefore,
for the acceptance tasks we cannot select multiple arguments
for every instance.
Instead, we select only one argument for each instance,
with the exception that
we dropped the ``very easy'' instances
for acceptance tasks
and selected two arguments to be queried
for the ``too hard'' instances,
which again amounts to $350$ instances in total.

For each task group except group D
the query arguments are selected at random,
maintaining a minimum number of yes- and no-instances, respectively.
For group A and E, the same arguments are used.
%\todo{mention even distribution between yes/no instances?
%see \url{benchmarks/script/ratios.csv}}

%\subsection{Further issues}
%\label{sec:further}

%\todo{statistics to be found in \url{benchmarks/results_first_run/GroupA.stable}}

\paragraph{Ideal Semantics}

While the selection of arguments for the decision tasks $\dc$ and $\ds$ in all task groups except D
was done randomly,
for the task $\dc$-$\id$ we were aiming for a more sophisticated selection
in order to select
the ``interesting'' arguments for the acceptance task.

That selection was based on the following insights:
%on the one hand,
\begin{itemize}
\item if the query argument is contained in the grounded extension,
then the answer to $\dc$-$\ideal$ is always yes;
\item if the query argument is not contained in every preferred extension,
then the answer to $\dc$-$\ideal$ is always no.
\end{itemize}
Hence, we aimed for a considerable number of instances
for which we select an argument contained in all preferred extensions,
but not in the grounded extension.
%i.e.\ an argument in $\bigcap{\prf(F)} \setminus \grd(F)$.

We did so by considering the following strategy:
Given an AF $F=(A,R)$, let $G \in \grd(F)$ be its grounded extension.
Moreover, let $\alpha$ and $\beta$ be random variables
with a uniform distribution in the interval $[0,1]$.
\begin{enumerate}
\item if $\bigcap{\prf(F)} \setminus G \neq \emptyset$ and $\alpha < 0.9$,
select an argument randomly
from \\ $\bigcap{\prf(F)} \setminus G$;
\item otherwise, if $G \neq \emptyset$ and $\beta < 0.6$,
select an argument randomly
from $G$;
\item otherwise,
select an argument randomly
from $A \setminus \bigcap{\prf(F)}$.
\end{enumerate}

That is, if arguments that we consider ``interesting'' as described before exist,
we select one of them with a high probability ($0.9$).
Otherwise we give a slight preference (probability of $0.6$)
to the arguments
contained in the grounded extension,
given that the grounded extension is not empty.

This strategy is applied to the selection of query arguments
for instances in the easy and medium hardness category.
The obtained
distributions of the selected arguments
is given in Table~\ref{tab:ideal}.
We randomly select the arguments for the hard and too hard instances.

\begin{table}
\centering
\caption{Distribution of selected arguments for $\dc$-$\ideal$, with $F$ being the AF and $G$ its grounded extension.}
\label{tab:ideal}
\begin{tabular}{l|c|c|c}
 & $G$ & $\bigcap{\prf(F)} \setminus G$ & $A \setminus \bigcap{\prf(F)}$  \\
\hline
easy & $14$ &  $\mathbf{15}$ &  $21$  \\
medium & $21$ &  $\mathbf{21}$ &  $58$ 
\end{tabular}
\end{table}

%\noindent

%%% Local Variables:
%%% TeX-master: "aij-iccma17-report"
%%% End:

%\input{participants}
\section{Participants}
\label{sec:part}

%% [{\bf TODO: List of participants, with structure from the AI Mag. Table from webpage with solvers's participation to tasks and tracks. Stress variety.}]

Sixteen solvers participate in the competition, and are listed in Table~\ref{tab:part}, together with the list of contributors and their institutions, and a main reference in the last column.  New entries compared to the previous edition are marked by $^\star$.

System descriptions for all solvers can be found on the competition webpage
at \url{http://argumentationcompetition.org/2017/submissions.html}.
The set of participants is characterized by a great variety of solving approaches.
We provide a grouping based on these approaches and provide some highlights
for each group.
Detailed results will be presented in Section~\ref{sec:res}. % and Appendix~A.

\begin{table}
 \scriptsize
 \begin{tabular}{|l|l|c|}
 \hline
  {\bf Solver} & {\bf Contributors} & {\bf Reference} \\ 
 \hline
  argmat-clpb$^\star$ & Fuan Pu (Tsinghua \uni{}, China) & \cite{PuLJ17} \\
   & Guiming Luo (Tsinghua \uni{}, China) & \url{https://sites.google.com/site/argumatrix/} \\
   & Yucheng Chen (Tsinghua \uni{}, China) &  \\
\hline
  argmat-dvisat$^\star$ & Fuan Pu (Tsinghua \uni{}, China) & \cite{PuLJ17} \\
  & Guiming Luo (Tsinghua \uni{}, China) & \url{https://sites.google.com/site/argumatrix/} \\
  & Ya Hang (Tsinghua \uni{}, China) &  \\
\hline
 argmat-mpg$^\star$ & Fuan Pu (Tsinghua \uni{}, China) & \cite{PuLJ17} \\
 & Guiming Luo (Tsinghua \uni{}, China) & \url{https://sites.google.com/site/argumatrix/} \\
 & Ya Hang (Tsinghua \uni{}, China) &  \\ %; 24 tasks, 8 tracks.
\hline
 argmat-sat$^\star$ & Fuan Pu (Tsinghua \uni{}, China) & \cite{PuLJ17} \\
 & Guiming Luo (Tsinghua \uni{}, China) & \url{https://sites.google.com/site/argumatrix/} \\ 
 & Ya Hang (Tsinghua \uni{}, China) &  \\ %; 24 tasks, 8 tracks.
\hline
 ArgSemSAT & Federico Cerutti (Cardiff \uni{}, UK) & \\
 & Mauro Vallati (\uni{} of Huddersfield, UK) & \cite{CeruttiGV14a}\\
 & Massimiliano Giacomin (\uni{} of Brescia, Italy) & \url{https://sourceforge.net/projects/argsemsat/} \\
 & Tobia Zanetti (\uni{} of Brescia, Italy) & \\ %; 18 tasks, 5 tracks.
\hline
 ArgTools & Samer Nofal (German Jordanian \uni{}, Jordan) & \cite{NofalAD16} \\
 & Katie Atkinson (\uni{} of Liverpool, UK) & \url{https://sourceforge.net/projects/argtools} \\
 & Paul E. Dunne (\uni{} of Liverpool, UK) & \\ %;  24 tasks, 7 tracks.
\hline
 ASPrMin$^\star$ & Wolfgang Faber (\uni{} of Huddersfield, UK) & \\
 & Mauro Vallati (\uni{} of Huddersfield, UK) & \cite{FaberVCG16} \\ 
 & Federico Cerutti (Cardiff \uni{}, UK) & {\tiny \url{https://helios.hud.ac.uk/scommv/storage/ASPrMin-v1.0.tar.gz}} \\
 & Massimiliano Giacomin (\uni{} of Brescia, Italy) & \\ %; 1 task, no track.
\hline
 cegartix & Wolfgang Dvo\v{r}\'ak (TU Wien, Austria) & \cite{DvorakJWW14} \\
 & Matti J\"arvisalo (\uni{} of Helsinki, Finland) & {\tiny \url{http://www.dbai.tuwien.ac.at/proj/argumentation/cegartix/}} \\
 & Johannes P. Wallner (TU Wien, Austria) & \\ %; 24 tasks, 8 tracks.
 \hline
 Chim\ae{}rarg$^\star$ & Federico Cerutti (Cardiff \uni{}, UK) & \cite{CeruttiVG18} \\
 & Mauro Vallati (\uni{} of Huddersfield, UK) & \url{https://github.com/federicocerutti/Chimaerarg} \\
 & Massimiliano Giacomin (\uni{} of Brescia, Italy) & \\ %; 2 tasks, no track.
\hline
 ConArg & Stefano Bistarelli (\uni{} of Perugia, Italy) & \cite{BistarelliS11} \\
 & Fabio Rossi (\uni{} of Perugia, Italy) & \url{http://www.dmi.unipg.it/conarg/} \\ 
 & Francesco Santini (\uni{} of Perugia, Italy) & \\ %; 24 tasks, 8 tracks.
\hline
 CoQuiAAS & Jean-Marie Lagniez (\uni{} of Artois, France) & \cite{LagniezLM15}\\
 & Emmanuel Lonca (\uni{} of Artois, France) & \url{http://www.cril.univ-artois.fr/coquiaas}\\
 & Jean-Guy Mailly (\uni{} of Artois, France) & \\ %; 24 tasks, 8 tracks.
\hline
 EqArgSolver$^\star$ & Odinaldo Rodrigues (King's College London, UK) & \cite{GabbayR16} \\
 & & {\tiny \url{http://nms.kcl.ac.uk/odinaldo.rodrigues/eqargsolver}} \\%; 15 tasks, 5 tracks.
\hline
 gg-sts$^\star$ & Tomi Jahunen (Aalto \uni{}, Finland) & \cite{BogaertsJT16} \\
 & Shahab Tasharrofi (Aalto \uni{}, Finland) & {\tiny \url{https://research.ics.aalto.fi/software/sat/gg-sts/}} \\ % 24 tasks, 8 tracks.
\hline
 goDIAMOND & Stefan Ellmauthaler (Leipzig \uni{}, Germany) & \cite{EllmauthalerS14} \\
 & Hannes Strass (Leipzig \uni{}, Germany) & {\tiny \url{https://sourceforge.net/p/diamond-adf/code/ci/go/tree/go/}} \\ %; 24 tasks, 8 tracks. 
\hline
 heureka$^\star$ & Nils Geilen (\uni{} of Koblenz-Landau, Germany) & \cite{GeilenT17}\\
 & Matthias Thimm (\uni{} of Koblenz-Landau, Germany) & \url{https://github.com/nilsgeilen/heureka} \\ %; 14 tasks, 4 tracks.
\hline
 pyglaf$^\star$ & Mario Alviano (\uni{} of Calabria, Italy) & \cite{Alviano17}\\
 & & \url{http://alviano.com/software/pyglaf/} \\%; 24 tasks, 8 tracks.
\hline
 \end{tabular}
  \caption{List of participants, with contributors, main reference paper, and link to the solver home page. $\star$ means newly submitted in the ICCMA series.}
 \label{tab:part}
 \end{table}

\begin{itemize}
\item Reductions to SAT: argmat-dvisat, argmat-sat, ArgSemSAT, cegartix, CoQuiAAS, gg-sts.
All of these systems are implemented in C++.
argmat-dvisat, argmat-sat, ArgSemSAT, and cegartix rely
on reductions to SAT or (iterative) calls to SAT solvers.
Two of them are among the top five solvers for each track except $\gr$.
While the backbone of both ArgSemSAT and cegartix is \mbox{MiniSAT}~\citep{EenS03},
argmat-dvisat and argmat-sat use CryptoMiniSat (\url{https://github.com/msoos/cryptominisat}) for \\ SAT solving.
gg-sts does not use SAT directly, but a reduction to an extension
of the second-order logic system presented in~\citep{BogaertsJT16}.
Finally,
CoQuiAAS uses various constraint programming techniques such as MaxSAT and Maximal Satisfiable Sets extraction.
% argmat-dvisat
% SAT -- CyptoMiniSat5
% SCC decomposition
% C++
%
% argmat-sat
% SAT -- CyptoMiniSat5
% C++
% 
% ArgSemSAT
% SAT -- MiniSAT, AllSAT
% C++
%
% cegartix
% SAT -- miniSAT
% C++
%
% gg-sts
% Beyond-SAT -- SAT-TO-SAT
% 2nd order encodings
% C++
%
% CoQuiAAS
% Constraint Programming
% Partial Max-SAT
% C++
%
\item Reductions to CSP: argmat-clpb, argmat-mpg, ConArg.
All of these systems are implemented in C++.
argmat-clpb employs
Constraint Logic Programming over Boolean variables in Prolog,
while
argmat-mpg uses a reduction to CSP using Gecode (\url{http://www.gecode.org/}).
Both are based on formulations of argumentation problems in Boolean matrix algebra.
Also ConArg implements a CSP approach using Gecode.

% argmat-clbp
% ``Constrain Logic Programming over Boolean variables'' -- Constraint Programming
% C++, Prolog
%
% argmat-mpg
% CSP Gecode
% C++
%
% ConArg
% Constrain Programming -- Gecode
% C++
%
\item Reductions to circumscription: pyglaf.
pyglaf
is  implemented  in  Python  and  uses
a circumscription solver extending the SAT solver
glucose \citep{AudemardS09}. pyglaf participated in all tracks and is 
one of the most successful participants (see below).
% pyglaf
% Circumscriptino
% python
% 
\item Reductions to ASP: ASPrMin, goDIAMOND.
Both systems rely on the state-of-the-art ASP system clingo \citep{DBLP:journals/corr/GebserKKS14}. While 
goDIAMOND consists of a suite of different encodings for all the considered semantics
(plus some native implementation for $\gr$ and $\id$), ASPrMIN makes use of a 
particular feature of clingo to control the heuristics such that only a certain form of 
subset-maximal answer-sets are delivered. This can be used to enumerate prefererred extensions.
Consequently, ASPrMIN only participated in the $\ee$-$\pr$ task (and, in fact, was the best solver for this single task) , whereas goDIAMOND entered all tracks (and reached the 2nd place in $\st$).
% ASPrMin
% ASP with heuristic maximization -- clingo
%
% goDIAMOND
% ASP -- clingo
% go
% ADF solver
% some direct procedures
% 
\item Direct approaches: ArgTools, EqArgSolver, heureka.
All of these solvers implement genuine algorithms in C++.
EqArgSolver is an enhancement of GRIS (submitted to ICCMA'15, \citep{ThimmV15}) and
uses the discrete version of the Gabbay-Rodrigues iteration schema~\citep{GabbayR16}.
ArgTools and heureka use various forms of backtracking algorithms
on the basis of labellings of arguments.
% ArgTools
% C++
%
% EqArgSolver
% Gabbay-Rodrigues Iteration Schema
% enhancement of GRIS
% SCC
% C++
%
% heureka
% backtracking algorithm
% C++
%
\item Portfolio-based approaches: Chim\ae{}rarg.
% Chimaerarg
% Static portfolio
% EE-PR: cegartix 450s, GRIS 150s
% EE-ST: LabSATSolver 300s, ArgTools 300s
This system uses
all  the  solvers  that  took  part  in  the  $\ee$-$\pr$,  and respectively, $\ee$-$\st$  tasks of
ICCMA'15, for generating a static schedule of solvers, whose performance are measured in terms of PAR10 score. Chim\ae{}rarg participated in these two tasks in ICCMA'17, running Cegartix, GRIS, LabSATSolver and ArgTools. Unfortunately, Chim\ae{}rarg delivered some
wrong results and thus did not rank very well. Checking the number of solved instances however shows the potential of this system. 
We shall provide a separate analysis of comparing best solvers from ICCMA'15 and ICCMA'17 in Section~\ref{sec:improv}.
\end{itemize}

%\begin{table}[h!]
\afterpage{%
    \clearpage
    \begin{landscape}% Landscape page
        \centering % Center table
{
\small
 \tabcolsep=.1cm
 \begin{tabular}{|l||r|r|r|r|r|r|r|r|r|r|r|r|r|r|r|r|r|r|r|r|r|r|r|r|r||r|}
 \hline
 & $\dt$ & \multicolumn{4}{l|}{$\com$} & \multicolumn{4}{l|}{$\pr$} & \multicolumn{4}{l|}{$\st$} & \multicolumn{4}{l|}{$\sst$} & \multicolumn{4}{l|}{$\stg$} & \multicolumn{2}{l|}{$\gr$} & \multicolumn{2}{l||}{$\id$} & \\     
 \hline
     {\bf Solver}& & \multicolumn{1}{r|}{$\dc$} & \multicolumn{1}{r|}{$\ds$} & \multicolumn{1}{r|}{$\se$} & \multicolumn{1}{r|}{$\ee$} & \multicolumn{1}{r|}{$\dc$} & \multicolumn{1}{r|}{$\ds$} & \multicolumn{1}{r|}{$\se$} & \multicolumn{1}{r|}{$\ee$} &\multicolumn{1}{r|}{$\dc$} & \multicolumn{1}{r|}{$\ds$} & \multicolumn{1}{r|}{$\se$} & \multicolumn{1}{r|}{$\ee$} &\multicolumn{1}{r|}{$\dc$} & \multicolumn{1}{r|}{$\ds$} & \multicolumn{1}{r|}{$\se$} & \multicolumn{1}{r|}{$\ee$} &\multicolumn{1}{r|}{$\dc$} & \multicolumn{1}{r|}{$\ds$} & \multicolumn{1}{r|}{$\se$} & \multicolumn{1}{r|}{$\ee$} & \multicolumn{1}{r|}{$\dc$} & \multicolumn{1}{r|}{$\ds$} &  \multicolumn{1}{r|}{$\dc$} & \multicolumn{1}{r||}{$\se$} & {\bf \#Task}\\
\hline
\hline
argmat-clpb & & $\surd$ & $\surd$ & $\surd$ & $\surd$ & & & & & $\surd$ & $\surd$ & $\surd$ & $\surd$ & & & & & & & & & $\surd$ & $\surd$ & & & 10 \\ 
 \hline
 argmat-dvisat & $\surd$ & $\surd$ & $\surd$ & $\surd$ & $\surd$ & $\surd$ & $\surd$ & $\surd$ & $\surd$ & $\surd$ & $\surd$ & $\surd$ & $\surd$ & & & & & & & & & $\surd$ & $\surd$ & $\surd$ & $\surd$ & 17 \\
 \hline
 argmat-mpg & $\surd$ & $\surd$ & $\surd$ & $\surd$ & $\surd$ & $\surd$ & $\surd$ & $\surd$ & $\surd$ & $\surd$ & $\surd$ & $\surd$ & $\surd$ & $\surd$ & $\surd$ & $\surd$ & $\surd$ & $\surd$ & $\surd$ & $\surd$ & $\surd$ & $\surd$ & $\surd$ & $\surd$ & $\surd$ & 25 \\
 \hline
 argmat-sat & $\surd$ & $\surd$ & $\surd$ & $\surd$ & $\surd$ & $\surd$ & $\surd$ & $\surd$ & $\surd$ & $\surd$ & $\surd$ & $\surd$ & $\surd$ & $\surd$ & $\surd$ & $\surd$ & $\surd$ & $\surd$ & $\surd$ & $\surd$ & $\surd$ & $\surd$ & $\surd$ & $\surd$ & $\surd$ & 25 \\
 \hline
 ArgSemSAT & & $\surd$ & $\surd$ & $\surd$ & $\surd$ & $\surd$ & $\surd$ & $\surd$ & $\surd$ & $\surd$ & $\surd$ & $\surd$ & $\surd$ & $\surd$ & $\surd$ & $\surd$ & $\surd$ & & & & & $\surd$ & $\surd$ & & & 18 \\
 \hline
 ArgTools & & $\surd$ & $\surd$ & $\surd$ & $\surd$ & $\surd$ & $\surd$ & $\surd$ & $\surd$ & $\surd$ & $\surd$ & $\surd$ & $\surd$ & $\surd$ & $\surd$ & $\surd$ & $\surd$ & $\surd$ & $\surd$ & $\surd$ & $\surd$ & $\surd$ & $\surd$ & $\surd$ & $\surd$ & 24 \\
 \hline
 ASPrMin & & & & & & & & & $\surd$ & & & & & & & & & & & & & & & & & 1 \\
 \hline
 cegartix & $\surd$ & $\surd$ & $\surd$ & $\surd$ & $\surd$ & $\surd$ & $\surd$ & $\surd$ & $\surd$ & $\surd$ & $\surd$ & $\surd$ & $\surd$ & $\surd$ & $\surd$ & $\surd$ & $\surd$ & $\surd$ & $\surd$ & $\surd$ & $\surd$ & $\surd$ & $\surd$ & $\surd$ & $\surd$ & 25 \\
 \hline
 Chim\ae{}rarg & & & & & & & & & $\surd$ & & & & $\surd$ & & & & & & & & & & & & & 2 \\
 \hline
 ConArg & $\surd$ & $\surd$ & $\surd$ & $\surd$ & $\surd$ & $\surd$ & $\surd$ & $\surd$ & $\surd$ & $\surd$ & $\surd$ & $\surd$ & $\surd$ & $\surd$ & $\surd$ & $\surd$ & $\surd$ & $\surd$ & $\surd$ & $\surd$ & $\surd$ & $\surd$ & $\surd$ & $\surd$ & $\surd$ & 25 \\
 \hline
 CoQuiAAS & $\surd$ & $\surd$ & $\surd$ & $\surd$ & $\surd$ & $\surd$ & $\surd$ & $\surd$ & $\surd$ & $\surd$ & $\surd$ & $\surd$ & $\surd$ & $\surd$ & $\surd$ & $\surd$ & $\surd$ & $\surd$ & $\surd$ & $\surd$ & $\surd$ & $\surd$ & $\surd$ & $\surd$ & $\surd$ & 25 \\
 \hline
 EqArgSolver & $\surd$ & $\surd$ & $\surd$ & $\surd$ & $\surd$ & $\surd$ & $\surd$ & $\surd$ & $\surd$ & $\surd$ & $\surd$ & $\surd$ & $\surd$ & & & & & & & & & $\surd$ & $\surd$ & & & 15 \\
 \hline
 gg-sts & $\surd$ & $\surd$ & $\surd$ & $\surd$ & $\surd$ & $\surd$ & $\surd$ & $\surd$ & $\surd$ & $\surd$ & $\surd$ & $\surd$ & $\surd$ & $\surd$ & $\surd$ & $\surd$ & $\surd$ & $\surd$ & $\surd$ & $\surd$ & $\surd$ & $\surd$ & $\surd$ & $\surd$ & $\surd$ & 25 \\
 \hline
 goDIAMOND & $\surd$ & $\surd$ & $\surd$ & $\surd$ & $\surd$ & $\surd$ & $\surd$ & $\surd$ & $\surd$ & $\surd$ & $\surd$ & $\surd$ & $\surd$ & $\surd$ & $\surd$ & $\surd$ & $\surd$ & $\surd$ & $\surd$ & $\surd$ & $\surd$ & $\surd$ & $\surd$ & $\surd$ & $\surd$ & 25 \\
 \hline
 heureka & & $\surd$ & $\surd$ & $\surd$ & $\surd$ & $\surd$ & $\surd$ & $\surd$ & $\surd$ & $\surd$ & $\surd$ & $\surd$ & $\surd$ & & & & & & & & & $\surd$ & $\surd$ & & & 14 \\
 \hline
 pyglaf & $\surd$ & $\surd$ & $\surd$ & $\surd$ & $\surd$ & $\surd$ & $\surd$ & $\surd$ & $\surd$ & $\surd$ & $\surd$ & $\surd$ & $\surd$ & $\surd$ & $\surd$ & $\surd$ & $\surd$ & $\surd$ & $\surd$ & $\surd$ & $\surd$ & $\surd$ & $\surd$ & $\surd$ & $\surd$ &  25 \\
 \hline
 \hline
 {\bf \#Solver} & 10 & 14 & 14 & 14 & 14 & 13 & 13 & 13 & 15 & 14 & 14 & 14 & 15 & 10 & 10 & 10 & 10 & 9 & 9 & 9 & 9 & 14 & 14 & 10 & 10 & \\
 \hline
\end{tabular}
}
\captionof{table}{Tasks supported by solvers.}
\label{tab:task-supp}
    \end{landscape}
    \clearpage% Flush page
}
%\caption{Tasks supported by solvers}\label{tab:task-supp}
%\end{table}

 In Table~\ref{tab:task-supp} %, instead, 
we also provide information about the participation to tasks of each solver. The table contains the solvers in its rows, and the tasks in its columns: a ``$\surd$'' indicates that a solver competes in a task. The table is completed by a last row reporting the number of solvers participating to each task, and a last column with the number of tasks supported by each solver. Without taking into account ASPrMin and Chim\ae{}rarg, which are specifically designed for enumeration and focus on very few semantics, all other solvers participate in at least 10 tasks. Half of the submitted solvers participate in all 25 tasks. %\todo{SW: I don't understand this sentence} 
The number of participants in single tasks ranges from 9 to 15 solvers. As far as participation in tracks is concerned, each track includes between 9 ($\stg$ semantics) and 14 ($\com$, $\st$, and $\gr$ semantics) solvers.

\section{Results and Awards}
\label{sec:res}

In this section we present the results of our experiments, run on a cluster of Intel Xeon (Haswell) with 2.60GHz, where time and memory limits have been set to 10 minutes and 4 GB for all tasks but $\dt$, and to 30 minutes and 6.5 GB for $\dt$. The first sub-section is devoted to announce the winners. In the second sub-section we compare the award winners of this year and the best solvers from the ICCMA'15 competition on this year's benchmarks, on common tracks. %Full results will be added in Appendix~A.

\subsection{Award winners}

In this sub-section we outline the winners of the competition. We remind that
the winner of each track has been awarded.

\begin{figure}
% \begin{minipage}[c]{0.46\linewidth}
\begin{tabular}{|l|r|r|r|r|r|r|c|}
\hline
 {\bf Solver} & {\bf Points} & Time & Correct & Wrong & TO & Other & USC (u) \\ % DC+DS+EE+SE
\hline
 pyglaf & 1229 & \footnotesize{28774.77} & 1229 & 0 & 168  & 3 & 0\\ % 5127.66+5446.25+12747.90+5452.96
\hline
 cegartix & 1188 & \footnotesize{19846.86} & 1188 & 0 & 205 & 7 & 1 (0)\\ % 7206.73+658.29+11374.57+607.27
\hline
 argmat-sat & 1167 & \footnotesize{10472.57} & 1167 & 0 & 204 & 29 & 0\\ % 2777.64+252.72+7192.75+249.46 
\hline
 goDIAMOND & 1156 & \footnotesize{18166.98} & 1176 & 4 & 181 &  39 & 2 (0)\\ % 4224.86+306.23+9636.08+3999.81
\hline
 argmat-dvisat & 1151 & \footnotesize{15259.38} & 1151 & 0 & 226 &  23 & 0\\ % 3566.01+105.50+11398.27+189.60 
\hline
 CoQuiAAS & 1132 & \footnotesize{10785.98} & 1132 & 0 & 149 & 119 & 0 \\ % 2594.77+247.73+6884.05+1059.43
\hline
 argmat-mpg & 1126 & \footnotesize{15133.06} & 1126 & 0 & 227 & 47 & 2 (2)\\ % 7922.23+298.74+6633.08+279.01 
\hline
 heureka & 1018 & \footnotesize{9869.94} & 1018 & 0 & 309 & 73 & 0\\ % 4955.97+520.03+3651.08+742.86
\hline
 ConArg & 1017 & \footnotesize{51015.41} & 1037 & 4 & 130 & 229 & 19 (11) \\ % 7836.48+426.29+42340.63+412.01
\hline
ArgTools & 935 & \footnotesize{36134.08} & 935 & 0 & 444 & 21 & 0\\ % 11592.28+7965.62+10190.85+6385.33
\hline
 ArgSemSAT & 900 & \footnotesize{20077.48} & 900 & 0 & 299 & 201 & 0 \\ % 2789.39+7445.12+6809.93+3033.04 
\hline
EqArgSolver & 401 & \footnotesize{5430.45} & 401 & 0 & 92 & 907 & 0\\ % 1697.47+837.13+1508.71+1387.14
\hline
 argmat-clpb &  40 & \footnotesize{4779.14} & 40 & 0 & 1109 & 251 & 0\\ % 1318.12+0.0+2858.38+602.64
\hline
 gg-sts & -1170 & \footnotesize{18203.86} & 834 & 402 & 107 & 57 & 12 (12)\\ % 983.99+7247.40+6290.68+3681.79
 \hline
 \end{tabular}
 % \caption{$\com$ track: Ranking of solvers.}\label{fig:co}
  \\
 % \end{figure}
 % \begin{figure}
  
% \end{minipage} 
% \hspace{0.05cm}
 %\begin{minipage}[c]{0.56\linewidth}
 \begin{center}
 \includegraphics[width=0.825\linewidth]{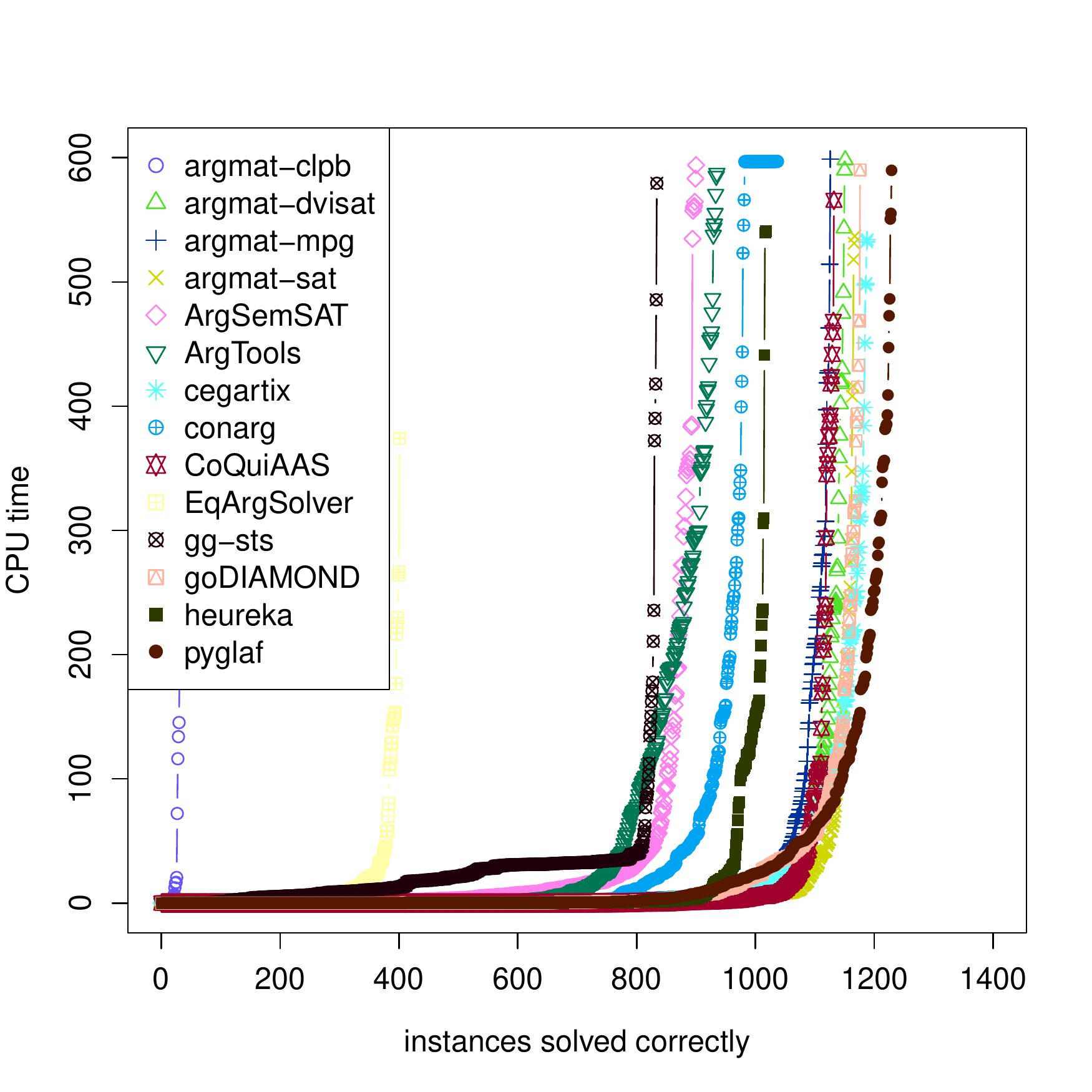}
 \end{center}
\vspace{-1cm}
% \end{minipage} 
 \caption{$\com$ track: Ranking of solvers (top). Cactus plot of runtimes (bottom).}\label{fig:co}
 \end{figure}

\begin{figure}[h!]
% \begin{minipage}[c]{0.46\linewidth}
%% \begin{table}
%% \center
\begin{tabular}{|l|r|r|r|r|r|r|c|}
\hline
 {\bf Solver} & {\bf Points} & Time & Correct & Wrong & TO & Other & USC (u)\\ % time
\hline
ArgSemSAT & 1146 & {\footnotesize 36607.37} & 1146 & 0 & 234 & 20 & 8 (0)\\ % 4697.26+7376.00+14678.70+9855.41
\hline
argmat-sat & 1139 & {\footnotesize 25110.57} & 1139 & 0 & 245 & 16 & 0\\ % 3921.02+5537.81+7717.52+7934.22
\hline
pyglaf & 1122 & {\footnotesize 43394.57} & 1127 & 1 & 272 & 1 & 5 (5)\\ % 4526.47+14751.76+13548.99+10567.35
\hline
argmat-dvisat & 1075 & {\footnotesize  28597.16} & 1075 & 0 &307 & 18 & 2 (2) \\ % 3591.10 % 8579.83 % 7956.65 % 8469.58
cegartix & 1075 & {\footnotesize  58263.31} & 1075 & 0 & 302  & 23 & 0\\ % 7349.59 % 20185.07 % 13850.56 % 16878.09
\hline
goDIAMOND & 1014 & {\footnotesize 51717.30} & 1069 & 11 & 289 & 31 & 0 \\ % 5882.46+13086.29+16723.17+16025.38
\hline
ArgTools & 898 & {\footnotesize 53147.54} & 898 & 0 & 501 & 1 & 0\\ % 11627.87+16478.36+10051.75+14989.56
\hline
ConArg & 773 & {\footnotesize 48197.84} & 773 & 0 & 433 & 194 & 1 (0)\\ % 9855.37+7865.16+24552.40+5924.91
\hline
heureka & 745 & {\footnotesize 19691.87} & 745 & 0 & 655 & 0 & 0 \\ % 5058.28 % 3669.74 % 5053.78 % 5910.07
argmat-mpg & 745 & {\footnotesize 30744.76} & 745 & 0 & 470 & 185 & 0 \\ % 5852.03 % 7947.32 % 6525.46 % 10419.95
\hline
EqArgSolver & 652 & {\footnotesize 6930.97} & 652 & 0 & 139 & 609 & 0\\ % 1403.64+1645.59+1788.27+2093.47
%ASPrMin & 285 & \\
%Chimaerarg & 92 & \\
\hline
CoQuiAAS & -863 & {\footnotesize 7756.35}& 477 & 268 & 228 & 427 & 0 \\ % 2574.27+4624.06+512.69+45.33
\hline
gg-sts & -1107 & {\footnotesize 32999.15} & 678 & 357 & 285 & 80 & 2 (1)\\ % 2832.85+10861.13+12402.93+6902.24
 \hline
\end{tabular}
 %\caption{$\pr$ track: Ranking of solvers. }\label{fig:pr_ranking}
%\end{figure}
\\
% \end{minipage} 
% \hspace{0.05cm}
% \begin{minipage}[c]{0.56\linewidth}
%\begin{figure}
\begin{center}
 \includegraphics[width=0.85\linewidth]{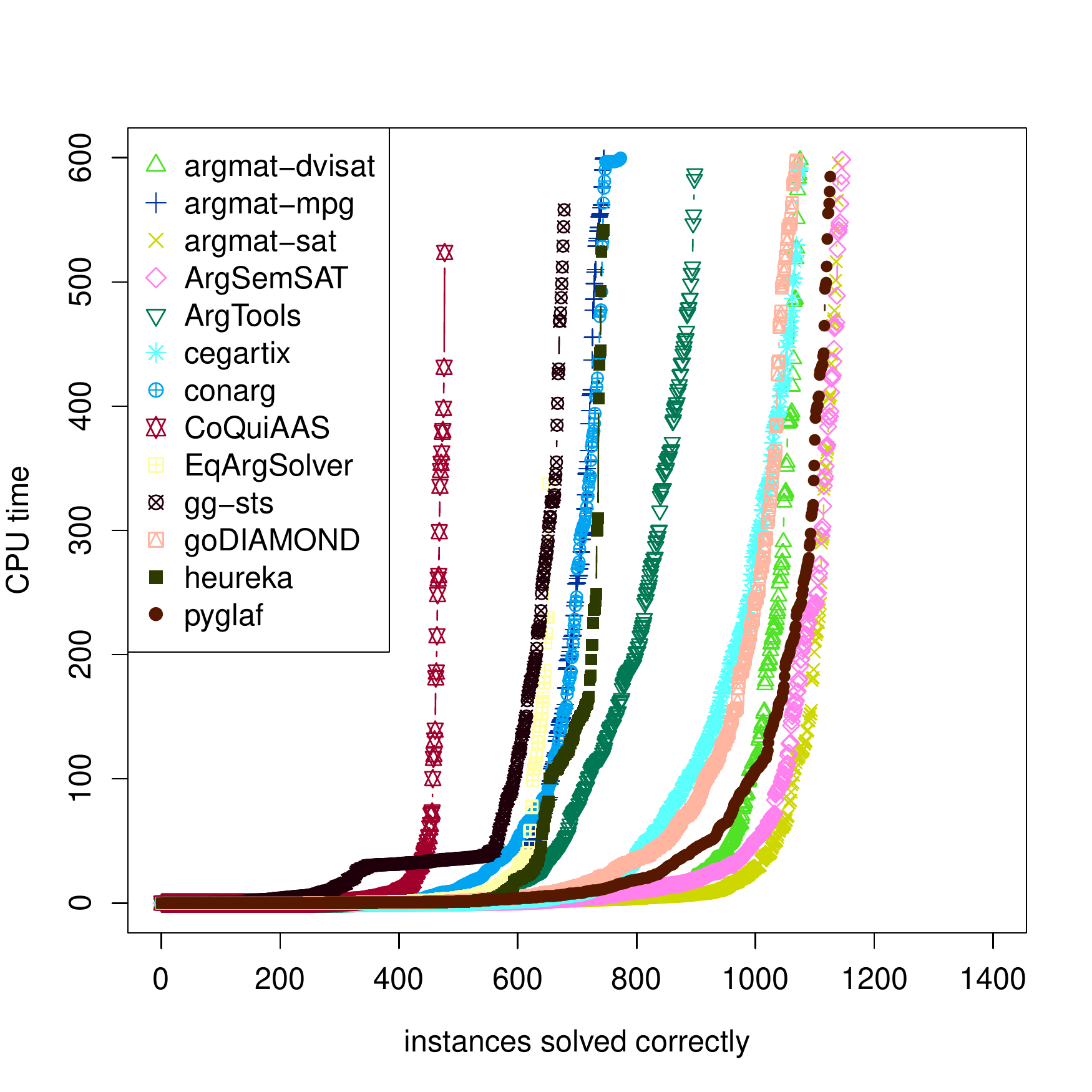}
 \end{center}
% \end{minipage} 
\vspace{-1cm}
 \caption{$\pr$ track: Ranking of solvers (top). Cactus plot of runtimes (bottom).}\label{fig:pr}
 \end{figure}

\begin{figure}[h!]
 %\begin{minipage}[c]{0.46\linewidth}
%\begin{table}
%\center
\begin{tabular}{|l|r|r|r|r|r|r|c|}
\hline
 {\bf Solver} & {\bf Points} & Time & Correct & Wrong & TO & Other & USC (u)\\ % time
\hline
 pyglaf & 1183 & {\footnotesize 47155.98} & 1183 & 0 & 217 & 0 & 0\\ % 5162.24+13561.09+14998.66+13433.99
\hline
 goDIAMOND & 1143 & {\footnotesize 30116.76} & 1143 & 0 & 224 & 33 & 5 (0)\\ % 3703.94+7492.59+9294.69+9625.54
\hline
 argmat-sat & 1129 & {\footnotesize 22087.70} & 1129 & 0 & 247 & 24 & 0\\ % 3449.19+5147.01+7057.86+6433.64
\hline
 cegartix & 1102 & {\footnotesize 33963.81} & 1102 & 0 & 283  & 15 & 1 (0) \\ % 6576.63+9170.14+9397.49+8819.55
\hline
 argmat-mpg & 1073 & {\footnotesize 52284.56} & 1073 & 0 & 311 & 16 & 1 (1)\\ % 4194.63+16365.51+17178.10+14546.32
\hline
 argmat-dvisat & 1039 & {\footnotesize 22591.20}& 1039 & 0 & 334 & 27 & 1 (0)\\ % 4162.23+4794.09+7649.85+5985.03
\hline
 ConArg & 1002 & {\footnotesize 58792.29} & 1002 & 0 & 348 & 50 & 0\\ % 3763.94+17001.48+21843.14+16183.73
\hline
 heureka & 938 & {\footnotesize 29417.69} & 938 & 0 & 439 & 23 & 0 \\ % 4923.77+8376.68+9830.58+6286.66
\hline
 ArgSemSAT & 888 & {\footnotesize 23200.99} & 888 & 0 & 291 & 221 & 1 (0)\\  % 2106.29+5781.30+7925.68+7387.72
\hline
 ArgTools & 687 & {\footnotesize 45465.87} & 917 & 46 & 316 & 121 & 0\\ % 11201.44+11634.59+18265.96+4363.88
\hline
 EqArgSolver & 558 & {\footnotesize 7820.17} & 558 & 0 & 118 & 724 & 0\\ % 1864.70+1866.23+2047.56+2041.68
\hline
 argmat-clpb & 135 & {\footnotesize 8840.31} & 135 & 0 & 1133 & 132 & 0  \\ % 1277.40+703.87+3402.92+3456.12
% Chimaerarg & -220 \\
\hline
 CoQuiAAS & -299 & {\footnotesize 13647.26} & 821 & 224 & 297 & 58 & 0 \\ % 2414.08+3507.63+3957.02+3768.53
\hline
 gg-sts & -1193 & {\footnotesize 19037.19} & 782 & 395 & 187 & 36 & 1 (0)\\ % 2836.55+5357.28+6525.92+4317.44
 \hline
\end{tabular}
% \caption{$\st$ track: Ranking of solvers.}\label{fig:st_ranking}
% \end{figure}
\\ 
% \end{minipage} 
% \hspace{0.05cm}
% \begin{minipage}[c]{0.56\linewidth}
%\begin{figure}
\begin{center}
 \includegraphics[width=0.85\linewidth]{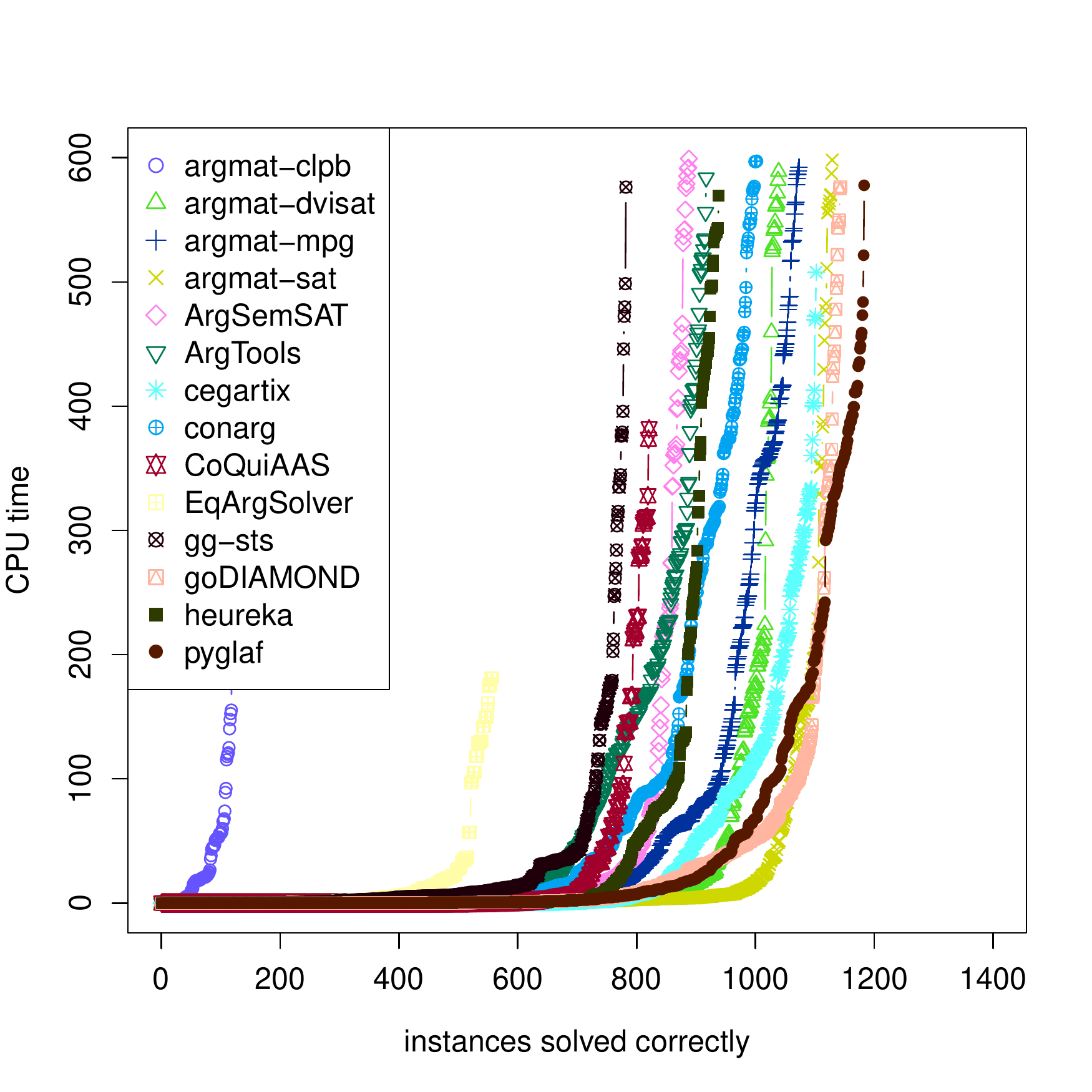}
 \end{center}
% \end{minipage} 
\vspace{-1cm}
 \caption{$\st$ track: Ranking of solvers (top). Cactus plot of runtimes (bottom).}\label{fig:st}
 \end{figure}

\begin{figure}[h!]
% \begin{minipage}[c]{0.46\linewidth}
%\begin{table}
%\center
\begin{tabular}{|l|r|r|r|r|r|r|c|}
\hline
 {\bf Solver} & {\bf Points} & Time & Correct & Wrong & TO & Other & USC (u)\\ % time
\hline
argmat-sat & 1164 & {\footnotesize 26043.50} & 1164 & 0 & 236 & 0 & 4 (1)\\ % 7246.49+7513.92+6700.27+4582.82
\hline
ArgSemSAT & 1113 & {\footnotesize 38816.07} & 1113 & 0 & 264 & 23 & 3 (0)\\ % 8925.05+7671.20+14042.56+8177.26
\hline
cegartix & 1091 & {\footnotesize 62543.78} & 1091 & 0 & 282 & 27 & 8 (0)\\ % 14631.76+19214.48+16219.86+12477.68
\hline
pyglaf & 1047 & {\footnotesize 41378.28} & 1047 & 0 & 349 & 4 & 1 (0)\\ % 13652.55+9350.64+10363.46+8011.63
\hline
goDIAMOND & 1032 & {\footnotesize 57957.15} & 1032 & 0 & 323 & 45 & 0\\ % 8163.27+15094.21+19624.99+15074.68
\hline
argmat-mpg & 755 & {\footnotesize 11464.36} & 755 & 0 & 419 & 226 & 3 (3)\\ % 2564.35+1908.36+3483.99+3507.66
\hline
ConArg & 668 & {\footnotesize 38572.13} & 668 & 0 & 437 & 295 & 24 (24)\\ % 5035.64+4842.81+25730.73+2962.95
\hline
ArgTools & 268 & {\footnotesize 52108.16} & 568 & 60 & 614 & 158 & 0\\ % 17601.22+12453.74+11039.45+11013.75
\hline
gg-sts & -1321 & {\footnotesize 22846.63} & 564 & 377 & 237 & 222 & 8 (2)\\ % 2134.87+4067.11+9331.95+7312.70
\hline
CoQuiAAS & -1642 & {\footnotesize 4855.65} & 218 & 372 & 215 & 595 & 0\\ % 4693.00+7.36+103.52+51.77
%% \hline
%% argmat-sat & 289 \\
%% ArgSemSAT & 284 \\
%% cegartix & 260 \\
%% goDiamond & 257 \\
%% pyglaf & 255 \\
%% argmat-mpg & 183 \\
%% ConArg & 163 \\
%% gg-sts & 66 \\
%% ArgTools & -177 \\
%% CoQuiAAS & -447 \\
 \hline
\end{tabular}
%\caption{$\sst$ track: Ranking of solvers.}\label{fig:sst_ranking}
%\end{figure}
\\
% \end{minipage} 
% \hspace{0.05cm}
% \begin{minipage}[c]{0.56\linewidth}
%\begin{figure}
 \includegraphics[width=1\linewidth]{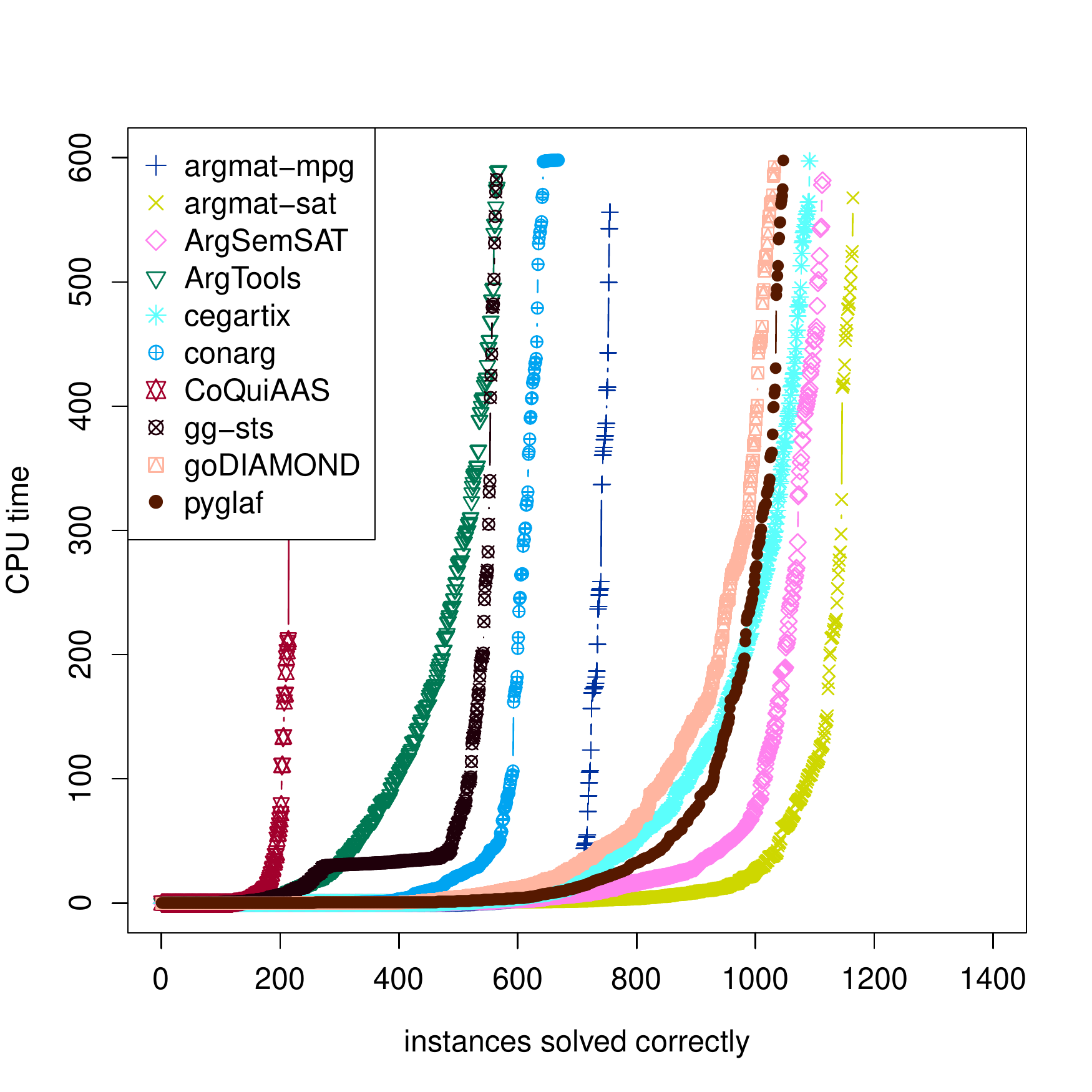}
% \end{minipage} 
 \vspace{-1cm}
 \caption{$\sst$ track: Ranking of solvers (top). Cactus plot of runtimes (bottom).}\label{fig:sst}
 \end{figure}

\begin{figure}[h!]
% \begin{minipage}[c]{0.46\linewidth}
%\begin{table}
%\center
\begin{tabular}{|l|r|r|r|r|r|r|c|}
\hline
 {\bf Solver} & {\bf Points} & Time & Correct & Wrong & TO & Other & USC (u) \\ % time
\hline
argmat-sat & 1065 & {\footnotesize 19948.06} & 1065 & 0 & 332 & 3 & 50 (1)\\ % 7331.12+8732.50+3142.80+741.64
\hline
pyglaf & 909 & {\footnotesize 32019.47} & 909 & 0 & 488 & 3 & 2 (0)\\ % 9551.98+8305.89+5761.45+8400.15
\hline
cegartix & 898 & {\footnotesize 62852.40} & 898 & 0 & 502 & 0 & 3 (0)\\ % 18725.14+14984.78+11545.91+17596.57
\hline
goDIAMOND & 724 & {\footnotesize 31394.75} & 724 & 0 & 629 & 47& 0\\ % 6605.11+6741.86+9167.36+8880.42
\hline
ConArg & 649 & {\footnotesize 43482.21} & 649 & 0 & 490 & 261 & 29 (29)\\ % 5378.28+5190.41+29879.38+3034.14
\hline
argmat-mpg & 618 & {\footnotesize 8381.57} & 618 & 0 & 396 & 386 & 4 (0)\\ % 1834.18+0.0+3869.08+2678.31
\hline
ArgTools & 67 & {\footnotesize 9558.97} & 172 & 21 & 1207 & 0 &  3 (3)\\ % 2046.04+1699.33+2090.21+3723.39
\hline
CoQuiAAS & -305 & {\footnotesize 4162.59} & 320 & 125 & 272 & 683 & 0\\ % 583.44+17.18+0.0+3561.97
\hline
gg-sts & -1325 & {\footnotesize 8242.35} & 185 & 302 & 654 & 259 & 4 (0)\\ % 3721.79+3941.86+27.26+551.44
%%  \hline
%% argmat-sat & 291 \\
%% pyglaf & 230 \\
%% cegartix & 226 \\
%% goDIAMOND & 179 \\
%% ConArg & 156 \\
%% gg-sts & 38 \\
%% ArgTools & 14 \\
%% argmat-mpg & 0? \\
%% CoQuiAAS & -236 \\
 \hline
\end{tabular}
% \caption{$\stg$ track:  Ranking of solvers.}\label{fig:stg_ranking}
% \end{figure}
 \\
% \end{minipage} 
% \hspace{0.05cm}
% \begin{minipage}[c]{0.56\linewidth}
%\begin{figure}
 \includegraphics[width=1\linewidth]{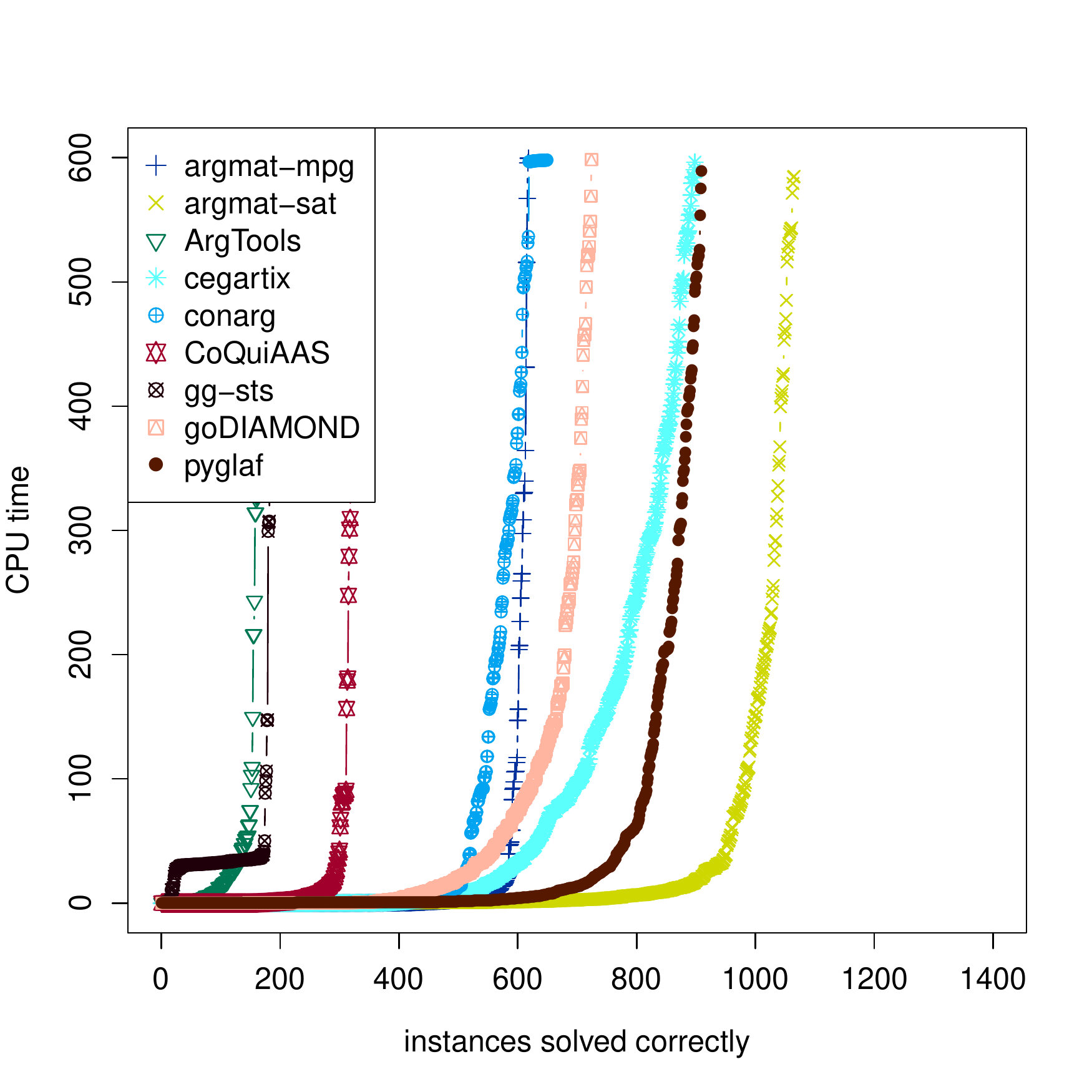}
 %\end{minipage} 
 \caption{$\stg$ track: Ranking of solvers (top). Cactus plot of runtimes (bottom).}\label{fig:stg}
 \end{figure}

\begin{figure}[h!]
% \begin{minipage}[c]{0.46\linewidth}
%\begin{table}
%\center
\begin{tabular}{|l|r|r|r|r|r|r|r|}
\hline
 {\bf Solver} & {\bf Points} & Time & Correct & Wrong & TO & Other & USC \\ % time
\hline
CoQuiAAS & 695 & {\footnotesize 335.85} & 695 & 0 & 3 & 2 & 0\\ % 200.96+134.79
cegartix & 695 & {\footnotesize 1152.51} & 695 & 0 & 0 & 5 & 0\\ % 660.73+491.78
\hline
heureka & 690 & {\footnotesize 671.37} & 690 & 0 & 8  & 2 & 0\\ % 599.16+72.21 
\hline
goDIAMOND & 688 & {\footnotesize 627.43} & 688 & 0 & 12 & 0 & 0\\ %305.09+322.34
\hline
%argmat-dvisat & ? \\ % 
pyglaf & 683 & {\footnotesize 11595.16} & 683 & 0 & 14 & 3 & 0\\ % 5825.87+5769.29
\hline
argmat-dvisat & 682 & {\footnotesize 163.80} & 682 & 0 & 4 & 14 & 0\\ % % 78.30+%85.50
argmat-clpb & 682 & {\footnotesize 263.21} & 682 & 0 & 4 & 14 & 0\\ % 128.55+134.66
EqArgSolver & 682 & {\footnotesize 502.80} & 682 & 0 & 18 & 0 & 0\\ %224.47+278.33
argmat-sat & 682 & {\footnotesize 504.75} & 682 & 0 & 4 & 14 & 0\\ % 253.15+251.60
\hline
ArgTools & 674 & {\footnotesize 15664.26} & 674 & 0 & 26 & 0 & 0\\ % 7963.62+7700.64
\hline
argmat-mpg & 662 & {\footnotesize 580.80} & 662 & 0 & 4 & 34 & 0 \\ % 304.95+275.85
\hline
ConArg & 588 & {\footnotesize 703.33} & 588 & 0 &  0 & 112 & 0\\ % 385.33+318.00
\hline
ArgSemSAT & 561 & {\footnotesize 11444.85} & 561 & 0 & 119 & 20 & 0\\ % 5887.68+5557.17
\hline
gg-sts & -1871 & {\footnotesize 4246.95} & 264 & 427 & 0 & 9 & 0\\ % 555.93+3691.02
%% \hline
%% heureka & 345 \\ % 72.21
%% CoQuiAAS & 345 \\ % 134.79
%% goDIAMOND & 345 \\ % 322.34
%% cegartix & 345 \\ % 491.78
%% pyglaf & 343 \\
%% argmat-clpb & 342 \\ % 134.66
%% argmat-sat &  342 \\ % 251.60
%% EqArgSolver & 342 \\ % 278.33
%% ArgTools & 338 \\
%% argmat-mpg & 332 \\
%% ConArg & 294 \\
%% ArgSemSAT & 289 \\
%% argmat-dvisat & ? \\
%% gg-sts & -247 \\
\hline
\end{tabular}
 %\caption{$\gr$ track:  Ranking of solvers.}\label{fig:gr_ranking}
 %\end{figure}
 \\
% \end{minipage} 
% \hspace{0.05cm}
% \begin{minipage}[c]{0.56\linewidth}
%\begin{figure}
\begin{center}
 \includegraphics[width=0.85\linewidth]{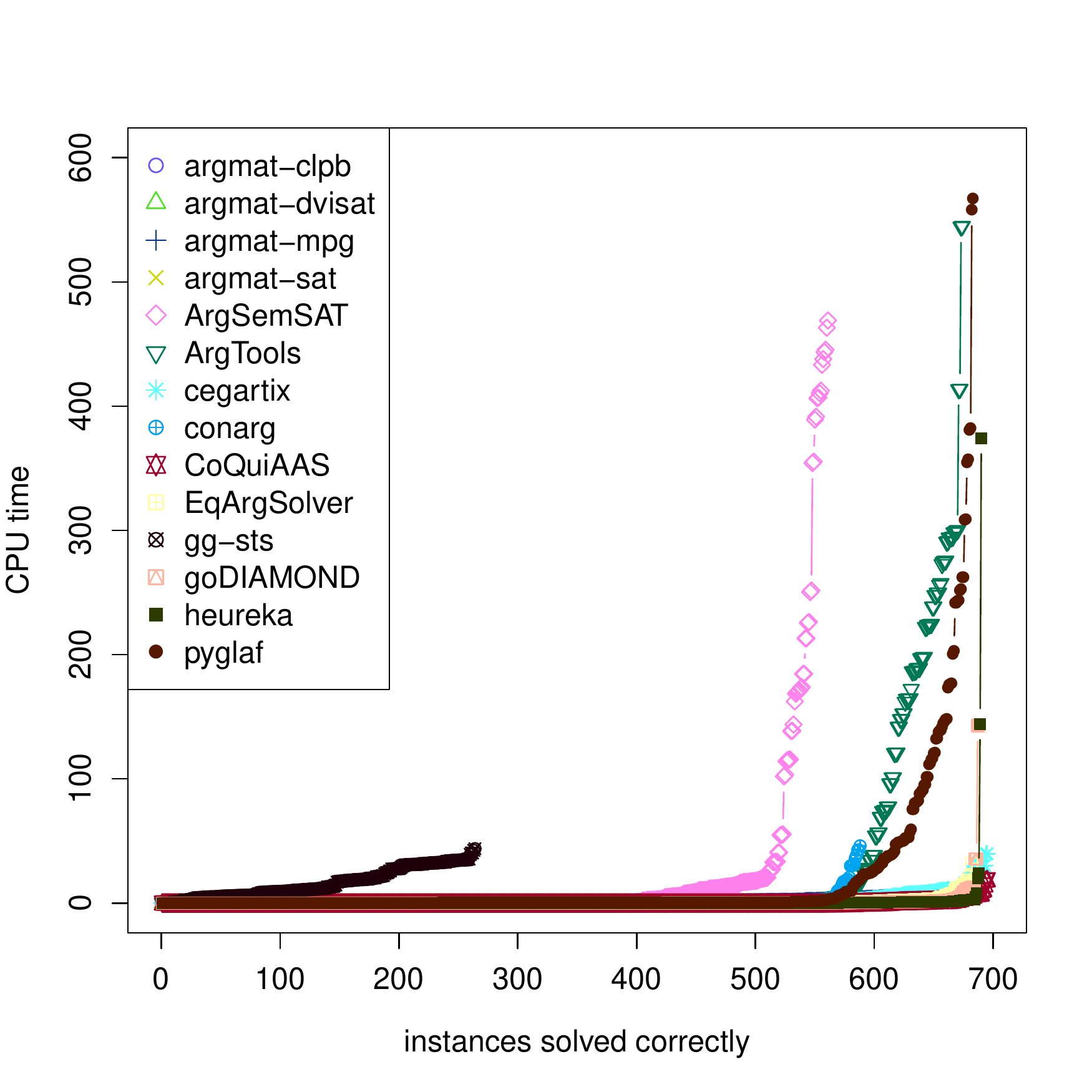}
 \end{center}
% \end{minipage} 
 \vspace{-1cm}
 \caption{$\gr$ track: Ranking of solvers (top). Cactus plot of runtimes (bottom).}\label{fig:gr}
 \end{figure}

\begin{figure}[h!]
% \begin{minipage}[c]{0.46\linewidth}
%\begin{table}
%\center
\begin{tabular}{|l|r|r|r|r|r|r|c|}
\hline
 {\bf Solver} & {\bf Points} & Time & Correct & Wrong & TO & Other & USC (u)\\ % time
\hline
 pyglaf & 585 & {\footnotesize 17341.50} &585 & 0 & 88 & 27 & 4 (0)\\ % 8712.36+8629.14
\hline
argmat-dvisat & 493 & {\footnotesize 17650.83} & 493 & 0 & 199 & 8 & 0\\ % 9199.92+8450.91
\hline
argmat-sat & 477 & {\footnotesize 16605.80} & 477 & 0 & 215 & 8 & 2 (0)\\ % 8841.36+7764.44
\hline
goDIAMOND & 414 & {\footnotesize 22496.34} & 414 & 0 & 270 & 16 & 0 \\ % 10713.49+11782.85
\hline
cegartix & 368 & {\footnotesize 25388.79} & 548 & 36 & 109 & 7 & 0 \\ % 13935.29+11453.50
\hline
ArgTools & 268 & {\footnotesize 20089.40} & 268 & 0 & 385 & 47 & 0\\ % 9981.80+10107.60
\hline
argmat-mpg & 217 & {\footnotesize 16031.89} & 217 & 0 & 396 & 87 & 0\\ % 7691.90+8339.99
\hline
ConArg & 181 & {\footnotesize 13254.90} & 181 & 0 & 434 & 85 & 1 (0)\\ % 6218.32+7036.58
\hline
CoQuiAAS & -794 & {\footnotesize 2597.28} & 156 & 190 & 94  & 260 & 1 (0)\\ % 2110.34+486.94
\hline
gg-sts & -1050 & {\footnotesize 13379.17} & 205 & 251 & 197 & 47 & 2 (0)\\ % 3500.05+9879.12
 \hline
\end{tabular}
% \caption{$\id$ track: Ranking of solvers.}\label{fig:id_ranking}
% \end{figure}
 \\
% \end{minipage} 
% \hspace{0.05cm}
% \begin{minipage}[c]{0.56\linewidth}
%\begin{figure}
 \includegraphics[width=1\linewidth]{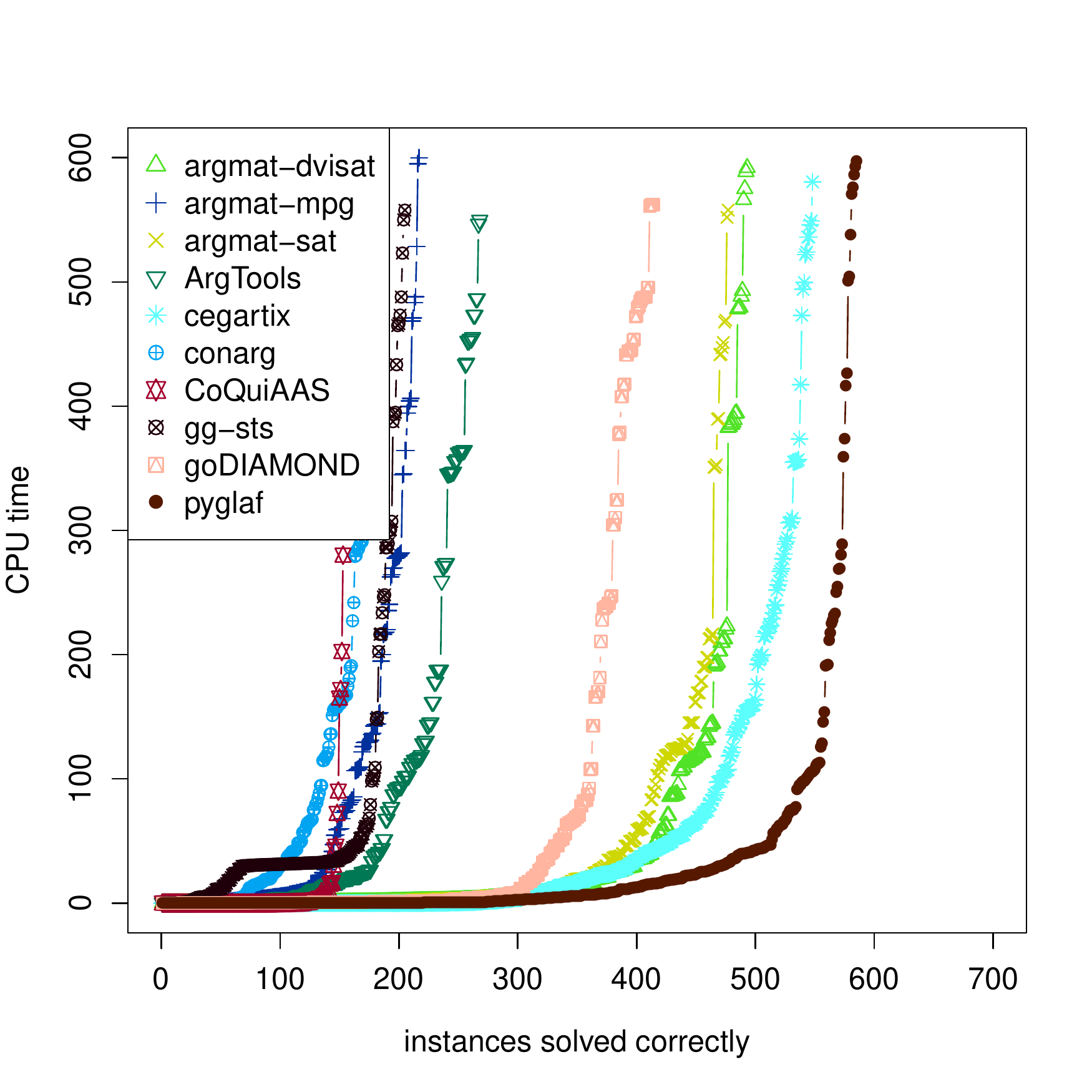}
% \end{minipage} 
 \vspace{-1cm}
 \caption{$\id$ track: Ranking of solvers (top). Cactus plot of runtimes (bottom).}\label{fig:id}
 \end{figure}

\begin{figure}[h!]
% \begin{minipage}[c]{0.46\linewidth}
%\begin{table}
%\center
\begin{tabular}{|l|r|r|r|r|r|r|c|}
\hline
 {\bf Solver} & {\bf Points} & Time & Correct & Wrong & TO & Other & USC (u) \\ % time
\hline
 argmat-dvisat & 276 & {\footnotesize 20222.07} & 276 & 0 & 68 & 6 & 5 (5)\\
\hline
pyglaf & 275 & {\footnotesize 25212.29} & 275 & 0 & 55 & 20 & 1 (1)\\
\hline
argmat-sat & 271 & {\footnotesize 22441.56} & 271 & 0 & 64 & 15 & 3 (3)\\
\hline
cegartix & 259 & {\footnotesize 35715.67} & 259 & 0 & 80 & 11 & 1 (0)\\
\hline
EqArgSolver & 192 & {\footnotesize 6577.89} & 192 & 0 & 32 & 126 & 0\\
ConArg & 192 & {\footnotesize 52007.99} & 192& 0 & 20 & 138 & 2 (2)\\ 
\hline
goDIAMOND & 179 & {\footnotesize 28857.58} & 179 & 0 & 52  & 119 & 0 \\
\hline
argmat-mpg & 164 & {\footnotesize 35916.74} & 164 & 0 & 158 & 28 & 0\\
\hline
gg-sts & -326 & {\footnotesize 25767.12} & 144 & 94 & 77 & 35 & 0\\
\hline
CoQuiAAS & -498 & {\footnotesize 441.22} & 32 & 106 & 43 & 169 & 0 \\

 \hline
\end{tabular}
 %\caption{$\dt$ track: Ranking of solvers. }\label{fig:d3_ranking}
 %\end{figure}
 \\
% \end{minipage} 
% \hspace{0.05cm}
% \begin{minipage}[c]{0.56\linewidth}
%\begin{figure}
 \includegraphics[width=1\linewidth]{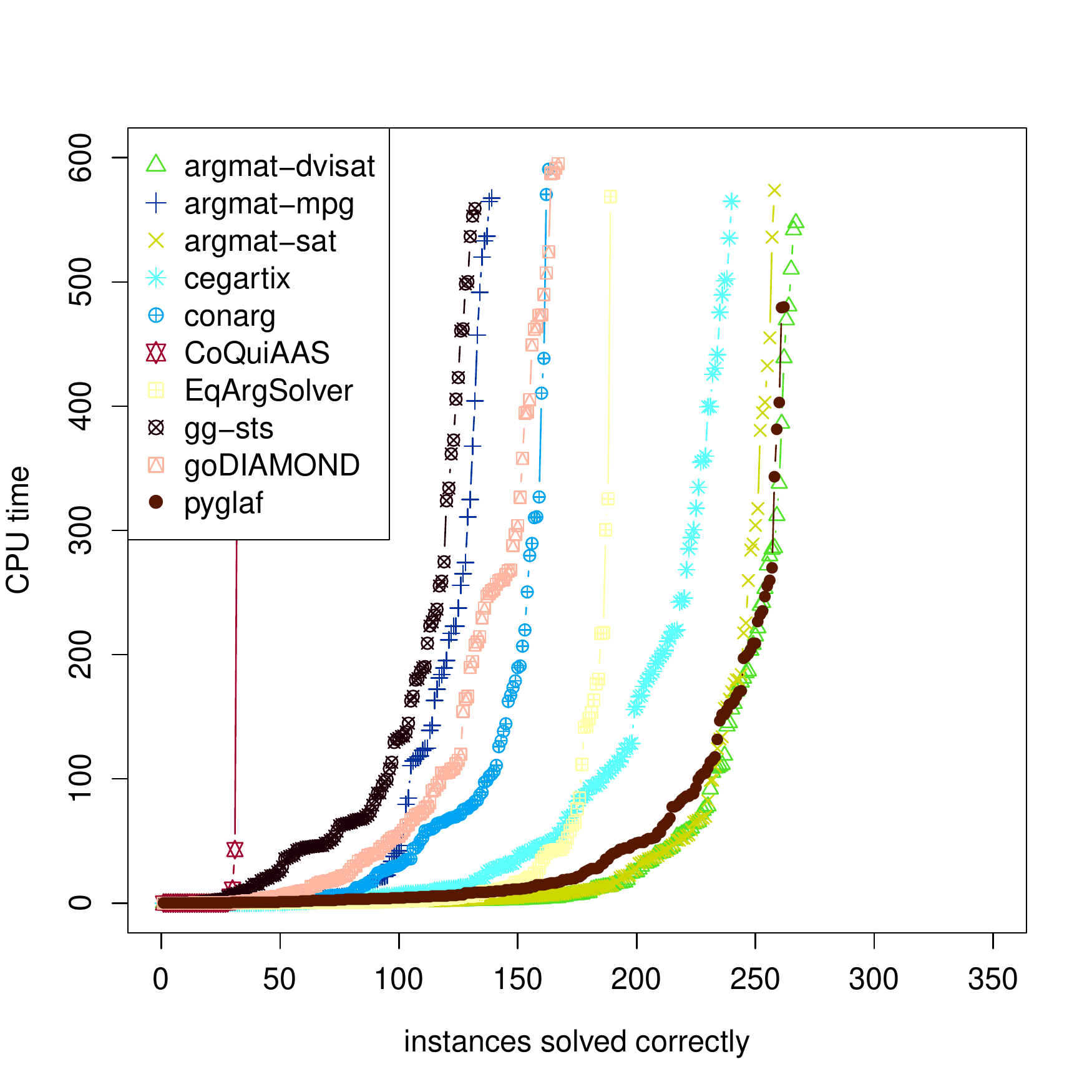}
% \end{minipage} 
\vspace{-1cm}
 \caption{$\dt$ track: Ranking of solvers (top). Cactus plot of runtimes (bottom).}\label{fig:d3}
 \end{figure}

Results are presented in Figures~\ref{fig:co}--\ref{fig:d3}, where at the top there is the ranking of solvers, and at the bottom the companion cactus plots. 
More specifically, the ranking of solvers is presented through tables organized as follows: the first column contains the name of the solver, the second column is the score of the respective solver (computed as defined in Section~\ref{sec:format}), while the third column reports the cumulative time of correctly solved instances. The fourth and fifth columns count the number of correct and wrong solutions given by each solver.
In the sixth column the number of instances reaching  timeout (TO) is given. The entries in seventh column (\emph{Other}) stand for all other 
instances which also got 0 points. These are incomplete, memory-out and not-parseable solutions including those where the solvers could only return some error messages.
The last column with \emph{USC (u)} shows the unique solver contributions (USC), being the number of instances where only one
solver could give a solution. The additional entries \emph{(u)} stand for \emph{unchecked}, that is the number of USC which could not
be verified (this is not specified when USC is 0). 
Solvers are ordered by score, and ties are broken by cumulative time, as defined already in Section~\ref{sec:format}. Cactus plots, instead, present another view of the results by showing the cumulative number of correctly solved instances ($x$-axis) within a given CPU time ($y$-axis).\\

\noindent
To sum up: 
\begin{itemize}
\item pyglaf has been the winner of the $\com$, $\st$, and $\id$ semantics;
\item argmat-sat has been the winner of the $\sst$ and $\stg$ semantics;
\item ArgSemSAT, CoQuiAAS and argmat-dvisat won the $\prf$, $\gr$, and $\dt$ semantics, respectively.
\end{itemize}

%% \caption{Results of the $\pr$ track.}\label{fig:pr}
%% \end{table}

%% \caption{Results of the $\st$ track.}\label{fig:st}
%% \end{table}

%% \caption{Results of the $\sst$ track.}\label{fig:sst}
%% \end{table}

%% \caption{Result of the $\stg$ track.}\label{fig:stg}
%% \end{table}

%\caption{Results of the $\gr$ track.}\label{tab:gr}
%\end{table}

%\caption{Results of the $\id$ track.}\label{tab:id}
%\end{table}

Interestingly, argmat-dvisat was not awarded as winner in any of the other track, but is the best solver in the $\dt$ track, where different semantics are considered. It is also worth to be noted that the set of winner solvers involves AF solvers based on different forms of reductions to SAT, CSP and circumscription. 

In the following we discuss the correctness of the solvers and the USC. 
The solvers argmat-clpb, argmat-dvisat, argmat-mpg, argmat-sat, ArgSemSAT, EqArgSolver and heureka always returned correct solutions in
all tracks. The solver pyglaf had only one incorrect solution in $\ds$-$\pr$, ConArg returned 4 incorrect answers in $\ee$-$\com$, goDIAMOND had in total 15 
wrong answers in tracks $\ee$-$\com$ and $\ee$-$\pr$. ArgTools had wrong solutions in tracks $\ds$-$\st$, $\dc$-$\sst$, $\ds$-$\sst$ and $\dc$-$\stg$, $\ds$-$\stg$  and $\ee$-$\stg$. 
Although the solver CoQuiAAS is the winner of the track $\gr$ and had no sanity problems in $\com$, in all other tracks many wrong answers were
given. Finally gg-sts had wrong answers in all tracks.
From the ranking of the solvers in all tracks it is easy to see that the penalty of -5 for each wrong answer had the desired effect to rank solvers with many wrong answers at the very end of the ranking. 

The solvers ChimaerArg and ASPrMin are not listed in the tables, as they did not contribute in all tasks of a track, thus we summarize 
the results for them in the following. 
ASPrMin was the winner of the task $\ee$-$\pr$ with 285 correct solutions, 0 wrong answers and thus obtained the score 285. The 2 USCs have
been verified and 63 instances resulted in timeouts while 2 fall into the category Other.
The solver ChimaerArg returned 255 correct solutions for the task $\ee$-$\st$ and 95 wrong answers, this results in the score -220. From the
21 USCs, 12 could not be verified. For $\ee$-$\pr$, ChimaerArg had 207 correct solutions and 23 wrong answers resulting in the score 92. All 120
answers with 0 points fall into the category Other.

%% \caption{Results of the $\dt$ track.}
%% \end{table}

\afterpage{%
    \clearpage% Flush earlier floats (otherwise order might not be correct)
    \thispagestyle{empty}% empty page style (?)
    \begin{landscape}% Landscape page
        \centering % Center table
{
\small
\begin{tabularx}{\linewidth}{l*{9}{|@{\hspace{1.2mm}}r@{\hspace{1.2mm}}|@{\hspace{1.2mm}}r@{\hspace{1.2mm}}}}
%\hline
     & \multicolumn{2}{c@{\hspace{1.2mm}}|@{\hspace{1.2mm}}}{\rotatebox{75}{CO -- pyglaf}} & \multicolumn{2}{c@{\hspace{1.2mm}}|@{\hspace{1.2mm}}}{\rotatebox{75}{PR -- ArgSemSAT}} & \multicolumn{2}{c@{\hspace{1.2mm}}|@{\hspace{1.2mm}}}{\rotatebox{75}{ST -- pyglaf}} & \multicolumn{2}{c@{\hspace{1.2mm}}|@{\hspace{1.2mm}}}{\rotatebox{75}{SST -- argmat-sat}} & \multicolumn{2}{c@{\hspace{1.2mm}}|@{\hspace{1.2mm}}}{\rotatebox{75}{STG -- argmat-sat}} & \multicolumn{2}{c@{\hspace{1.2mm}}|@{\hspace{1.2mm}}}{\rotatebox{75}{GR -- CoQuiAAS}} & \multicolumn{2}{c@{\hspace{1.2mm}}|@{\hspace{1.2mm}}}{\rotatebox{75}{ID -- pyglaf}} & \multicolumn{2}{c@{\hspace{1.2mm}}|@{\hspace{1.2mm}}}{\rotatebox{75}{D3 -- argmat-dvisat}} & 
     \rotatebox{90}{Total of points} &
     \rotatebox{90}{Max.\ number of points}  \\
%    & pyglaf    & ArgSemSAT & pyglaf    & argmat-sat & argmat-sat & CoQuiAAS & pyglaf & \\
\hline
\input{domainTable.txt}
\end{tabularx}
}
\captionof{table}{Points acquired by track winners for each domain.}
\label{table:domains}
    \end{landscape}
    \clearpage% Flush page
}

Finally, Table~\ref{table:domains} gives more details for the track winners. In particular it is given, for each track, the number of points acquired by the winning system in each domain. More in details, the table is organized as follows: the rows contain the domains and the columns the track winners. Each column is then divided in two sub-columns containing the number of points acquired by the solver and the maximum acquirable number of points in a domain, respectively. The table is complemented by a last row and a last column containing the total number of points acquired (or, acquirable) by each solver and in a domain, respectively. 

\subsection{Comparison to the results of ICCMA'15%Improvements to the 1st ICCMA
}\label{sec:improv}

By comparing the award winners of the 2017 event with those of the first edition, which cumulatively awarded CoQuiAAS, ArgSemSAT, and LabSATSolver in first, second and third place, respectively, we notice that CoQuiAAS and ArgSemSAT are in this year the winners of two tracks and ArgSemSAT is second-best in another track, while for the remaining semantics other AF solvers, mainly newcomers, have best performance. 

Goal of this sub-section is to (qualitatively) compare the award winners of this year's event to the best solvers in the past competition on common tracks. The comparison is done 
using the benchmarks from the current competition.

Given that the first competition awarded only global results, we applied the Borda count to the tracks of 2015 to get track winners. Thus, the 2015 (version of the) solvers ASPARTIX-D, ArgSemSAT, again ASPARTIX-D, and CoQuiAAs have been run for $\com$, $\pr$, $\st$, and $\gr$ semantics, respectively. Such additional experiments have been conducted on a separate machine, which is an Intel Xeon CPU E5345, 
2.33GHz; 2 processors with each 4 physical cores; no hyperthreading enabled.

\begin{figure}[h!t]
 \begin{minipage}[c]{0.5\linewidth}
 \includegraphics[width=1\linewidth]{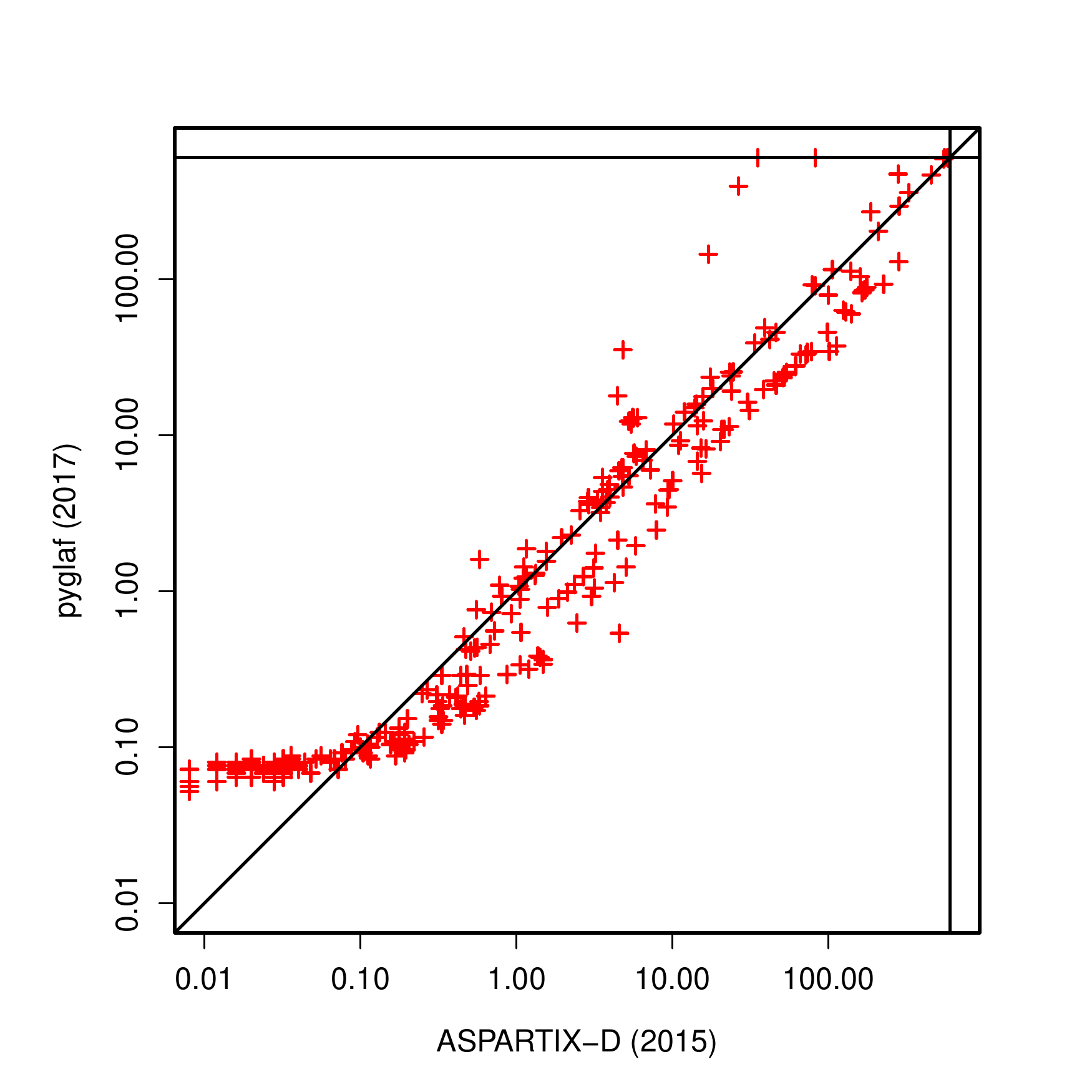}
 \end{minipage} 
 \begin{minipage}[c]{0.5\linewidth}
 \includegraphics[width=1\linewidth]{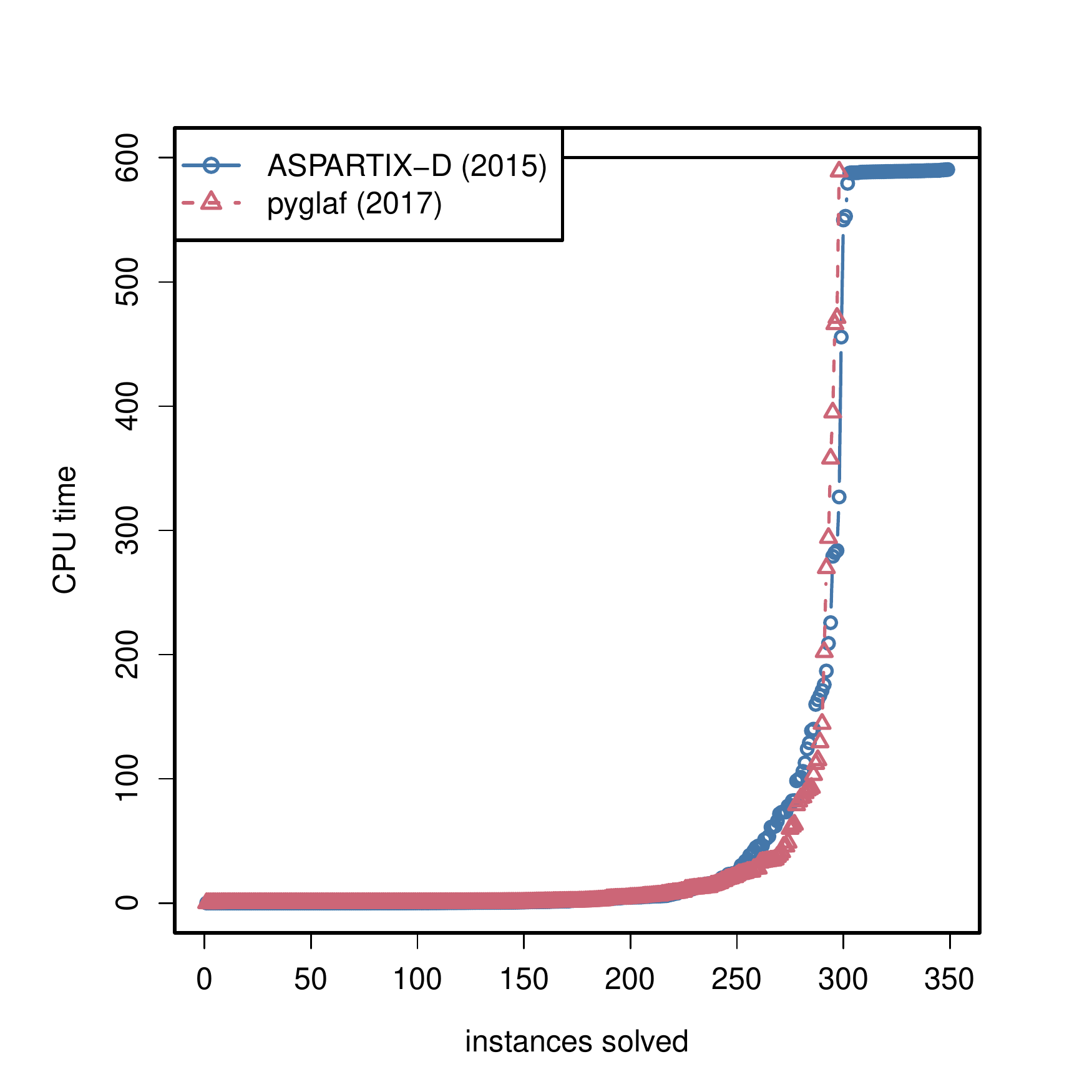} 
 \end{minipage}
 \\
 \begin{minipage}[c]{0.5\linewidth}
 \includegraphics[width=1\linewidth]{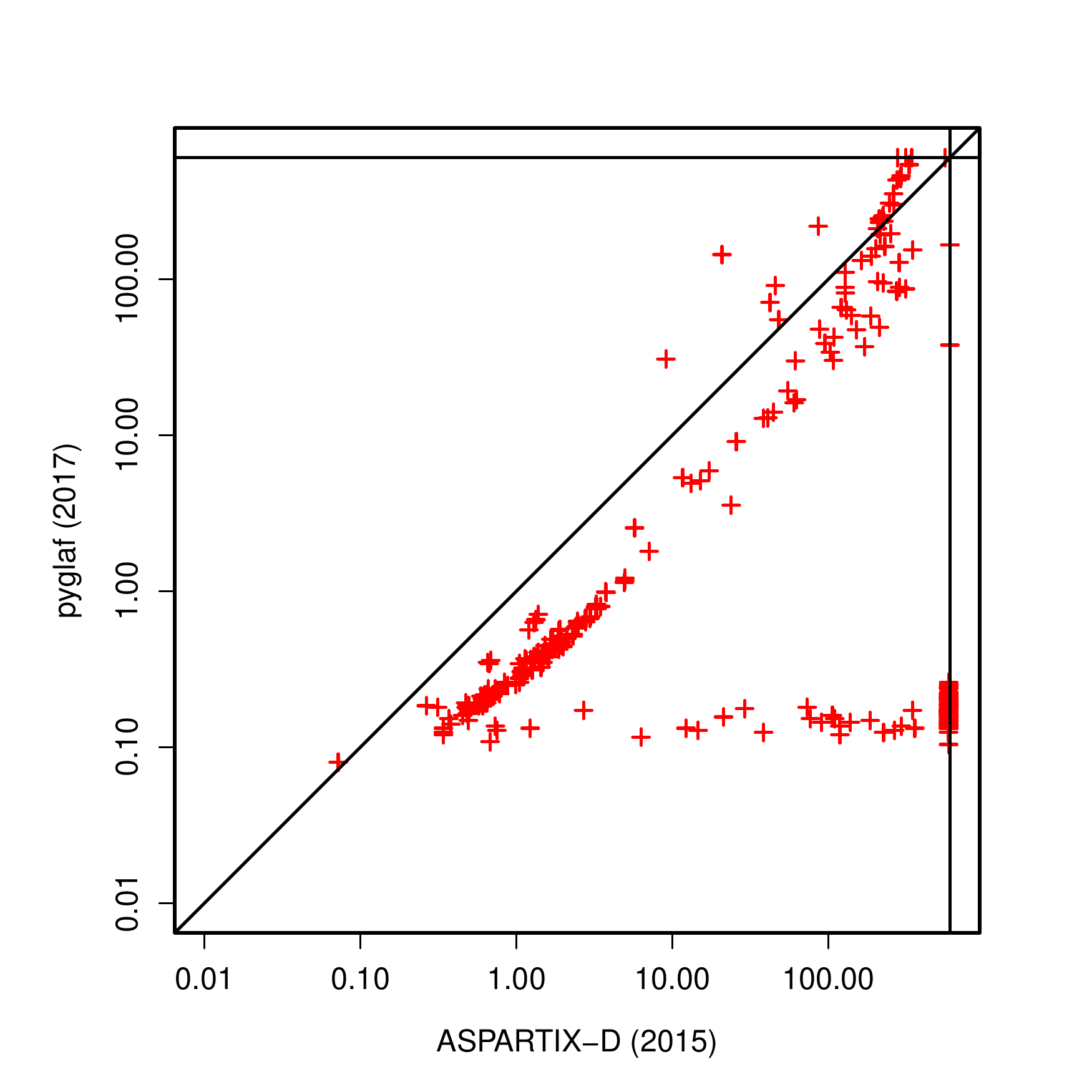}
 \end{minipage} 
 \begin{minipage}[c]{0.5\linewidth}
 \includegraphics[width=1\linewidth]{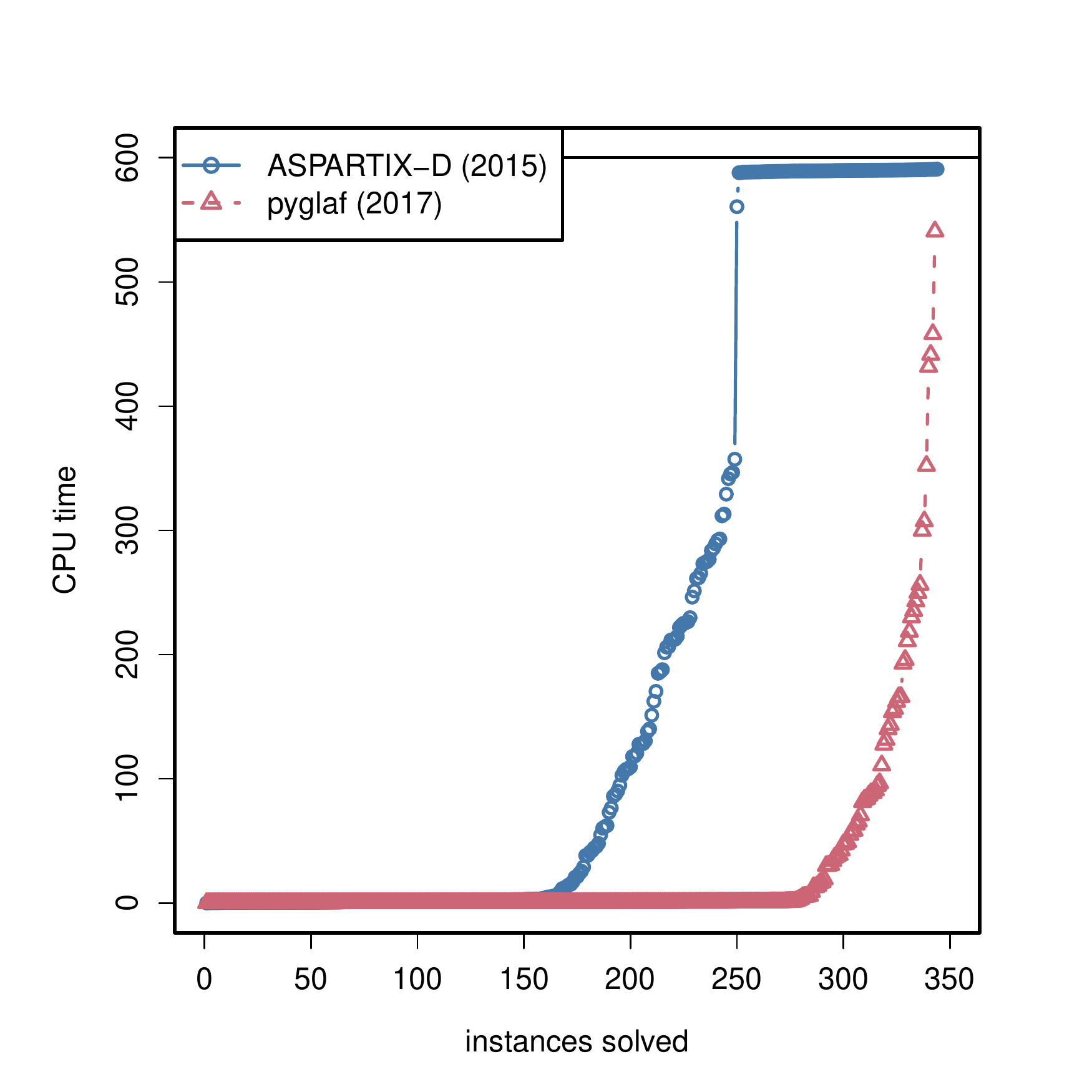} 
 \end{minipage}
 \caption{$\com$ track, $\dc$ and $\ds$ tasks: Comparison between ASPARTIX-D (2015) and pyglaf (2017).}\label{fig:co_dc_ds}
 \end{figure}

\begin{figure}[h!t]
 \begin{minipage}[c]{0.5\linewidth}
 \includegraphics[width=1\linewidth]{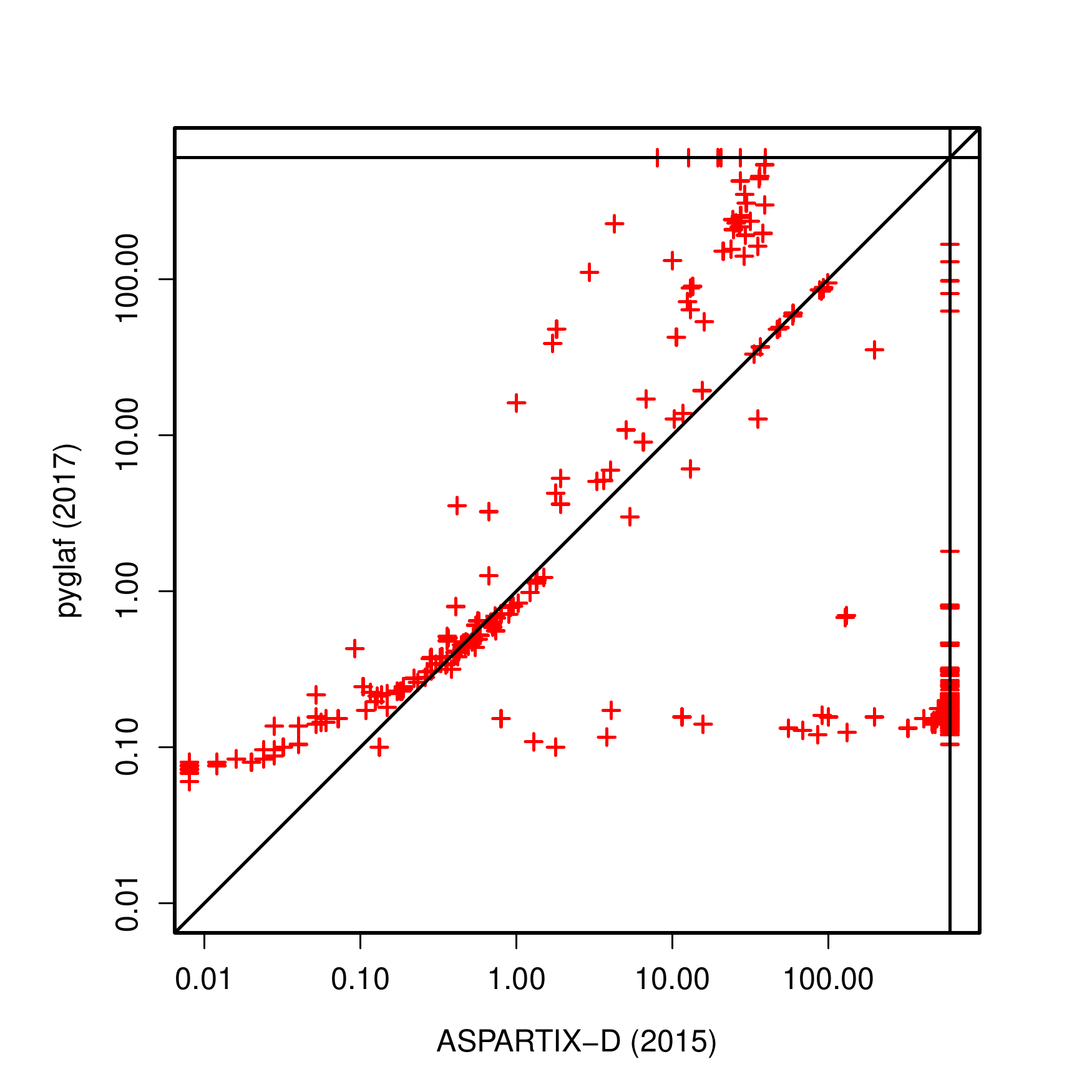}
 \end{minipage} 
 \begin{minipage}[c]{0.5\linewidth}
 \includegraphics[width=1\linewidth]{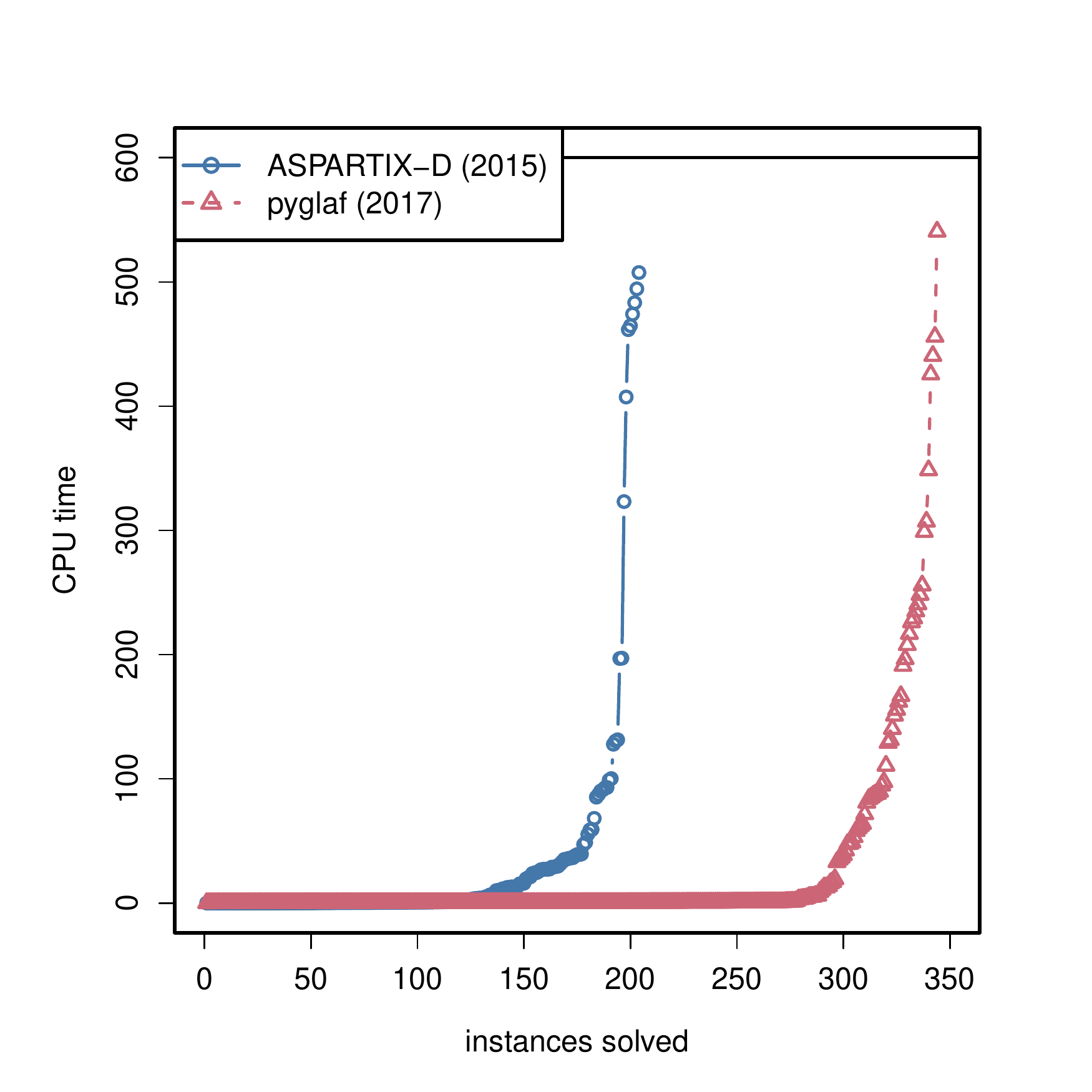} 
 \end{minipage}
 \\
 \begin{minipage}[c]{0.5\linewidth}
 \includegraphics[width=1\linewidth]{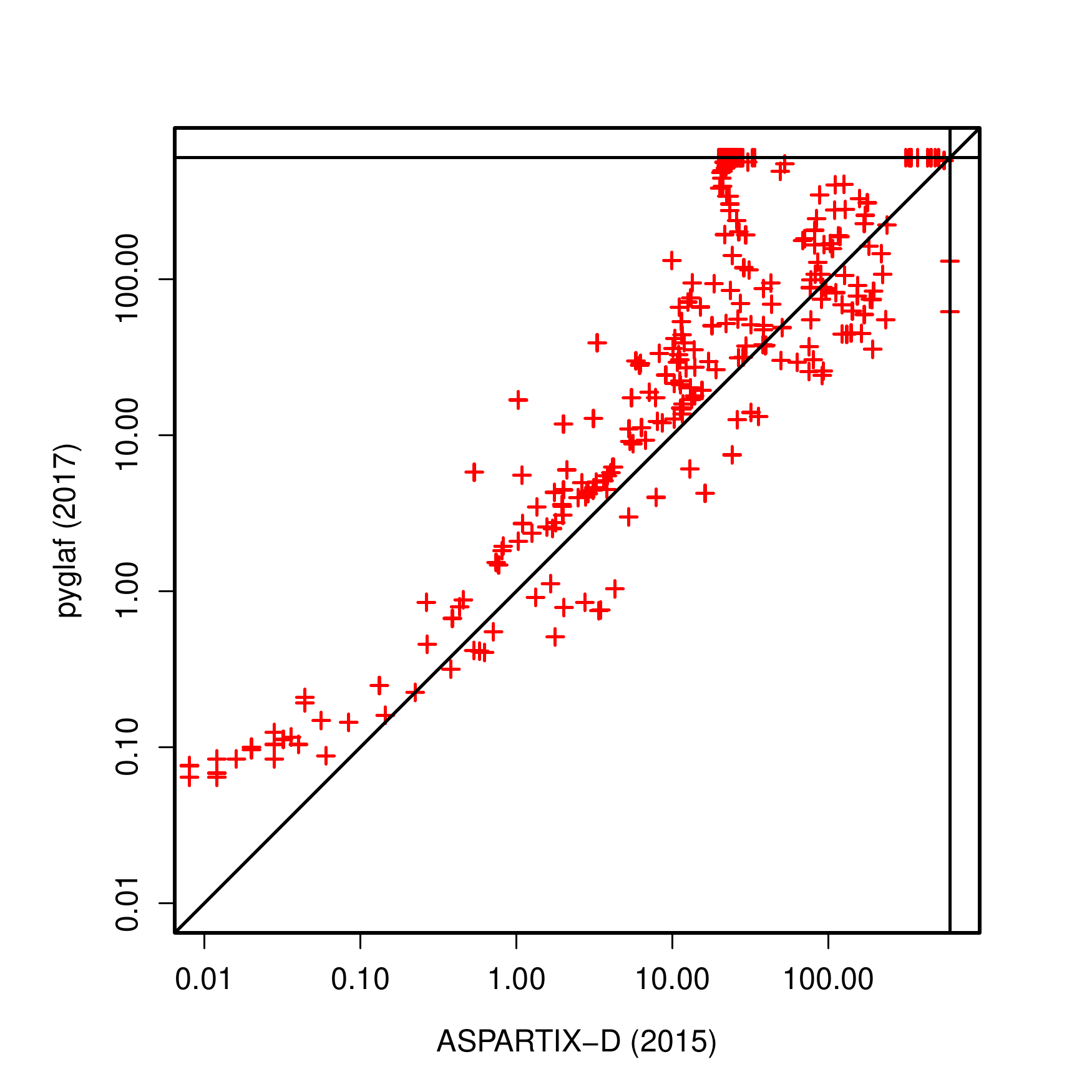}
 \end{minipage} 
 \begin{minipage}[c]{0.5\linewidth}
 \includegraphics[width=1\linewidth]{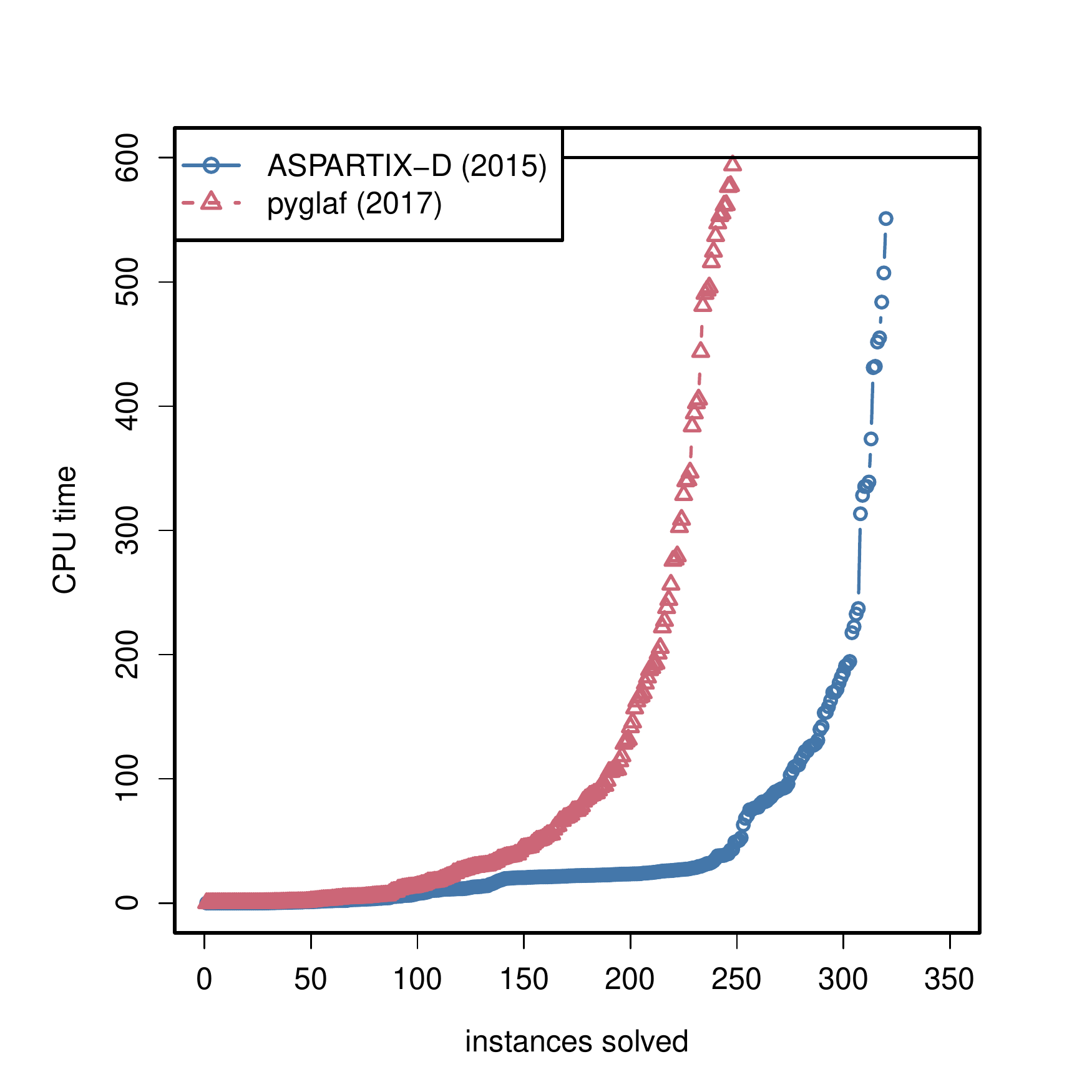} 
 \end{minipage}
 \caption{$\com$ track, $\se$ and $\ee$ tasks: Comparison between ASPARTIX-D (2015) and pyglaf (2017.)}\label{fig:co_se_ee}
 \end{figure}

\begin{figure}[h!t]
 \begin{minipage}[c]{0.5\linewidth}
 \includegraphics[width=1\linewidth]{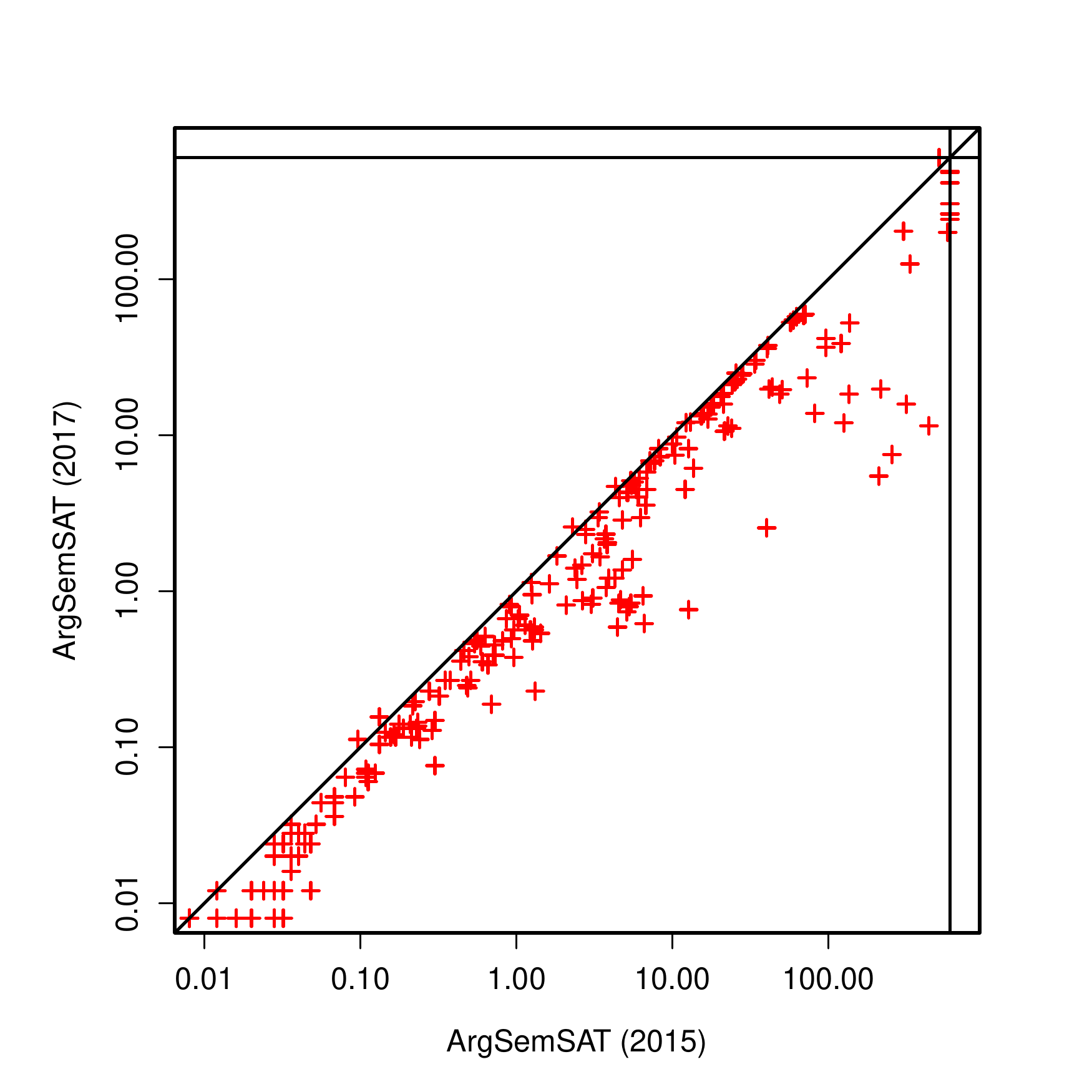}
 \end{minipage} 
 \begin{minipage}[c]{0.5\linewidth}
 \includegraphics[width=1\linewidth]{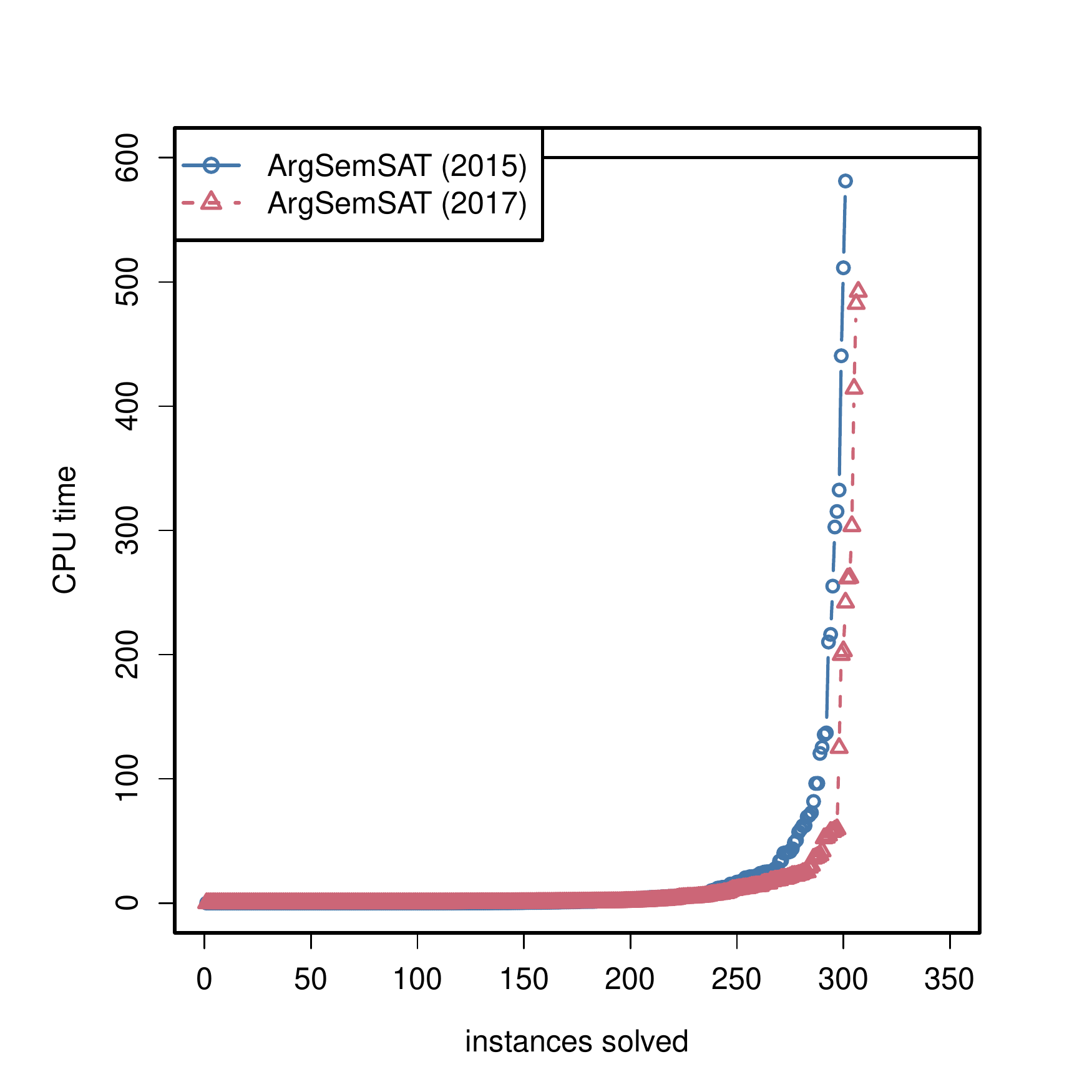} 
 \end{minipage}
 \\
 \begin{minipage}[c]{0.5\linewidth}
 \includegraphics[width=1\linewidth]{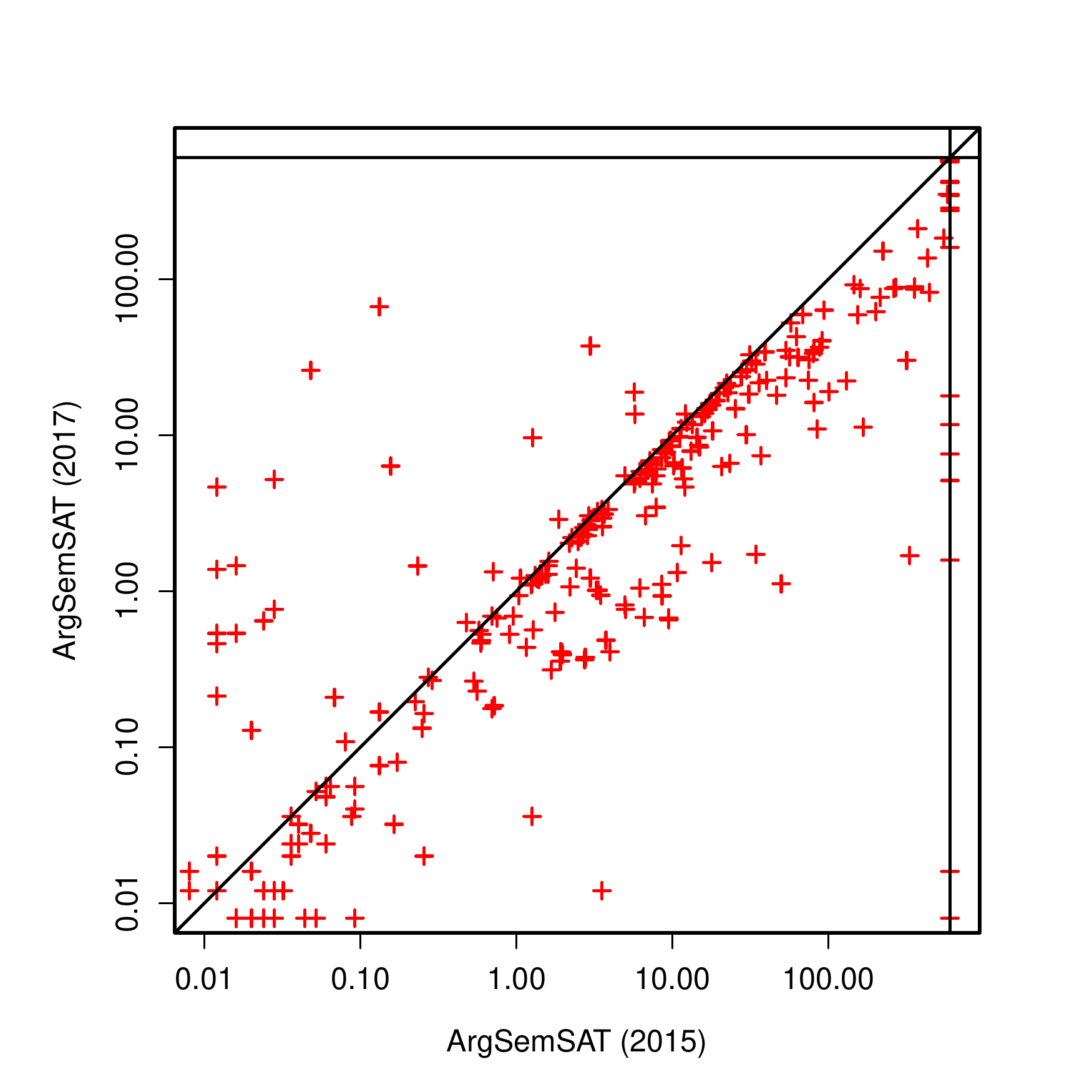}
 \end{minipage} 
 \begin{minipage}[c]{0.5\linewidth}
 \includegraphics[width=1\linewidth]{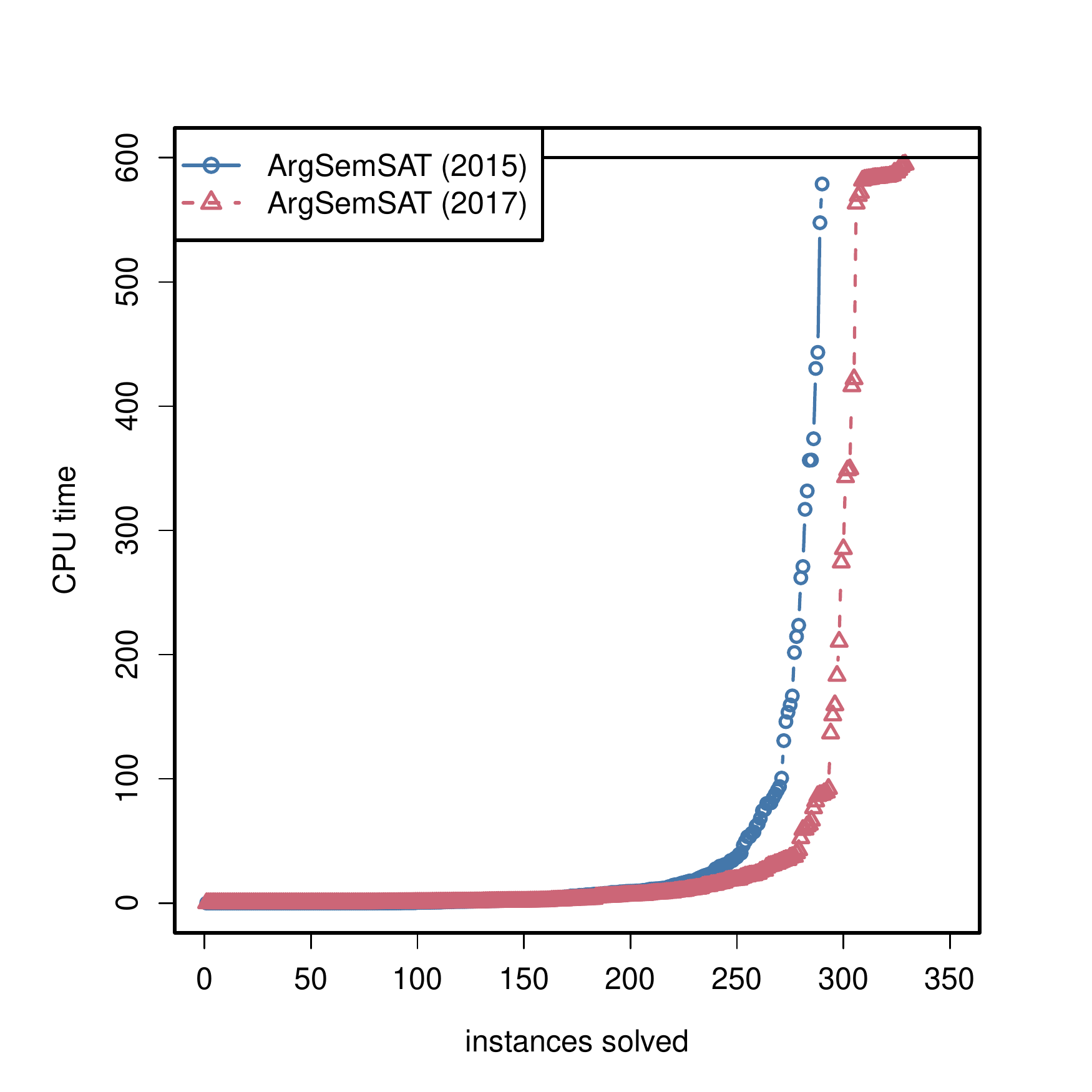} 
 \end{minipage}
 \caption{$\pr$ track, $\dc$ and $\ds$ tasks: Comparison between ArgSemSAT (2015) and ArgSemSAT (2017).}\label{fig:pr_dc_ds}
 \end{figure}

\begin{figure}[h!t]
 \begin{minipage}[c]{0.5\linewidth}
 \includegraphics[width=1\linewidth]{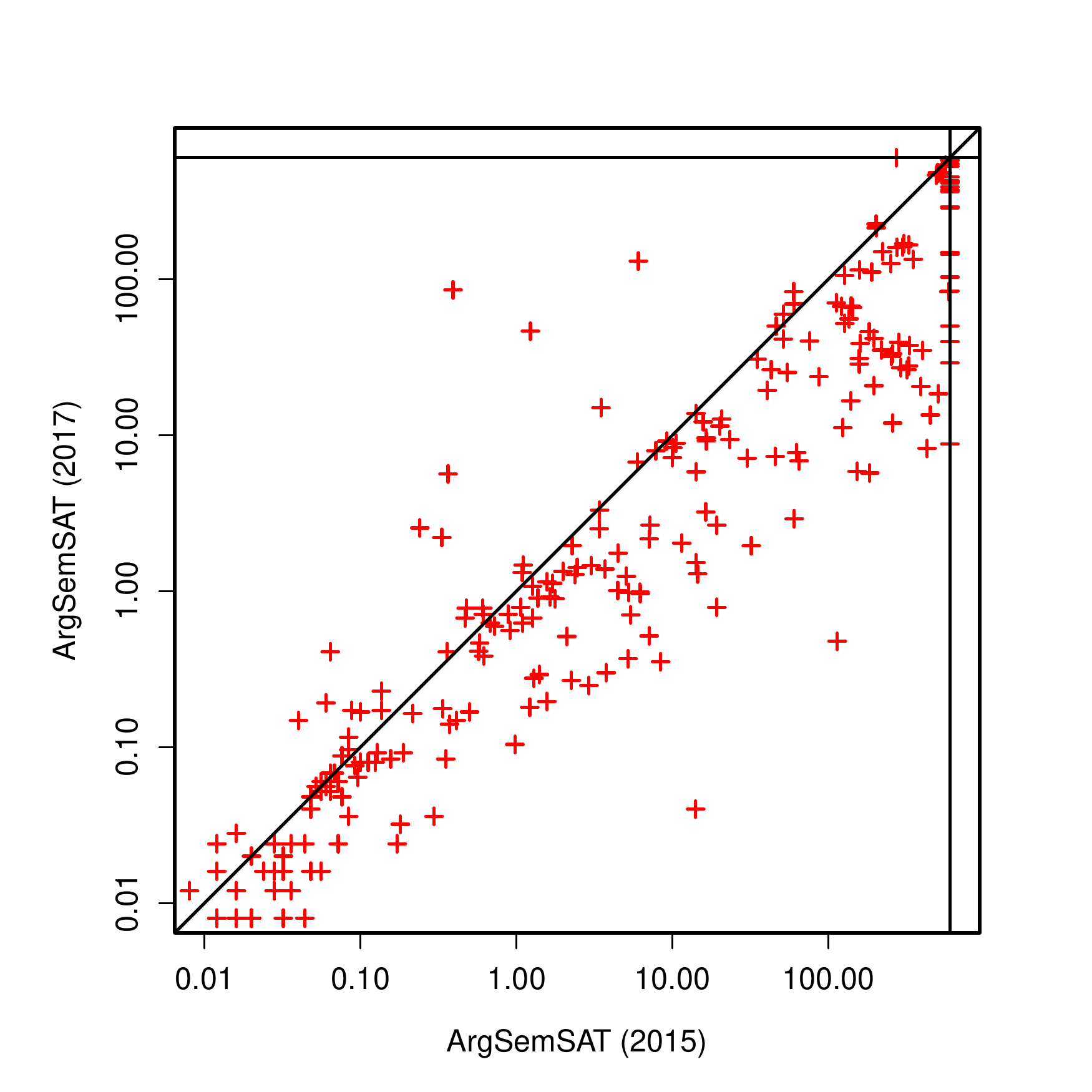}
 \end{minipage} 
 \begin{minipage}[c]{0.5\linewidth}
 \includegraphics[width=1\linewidth]{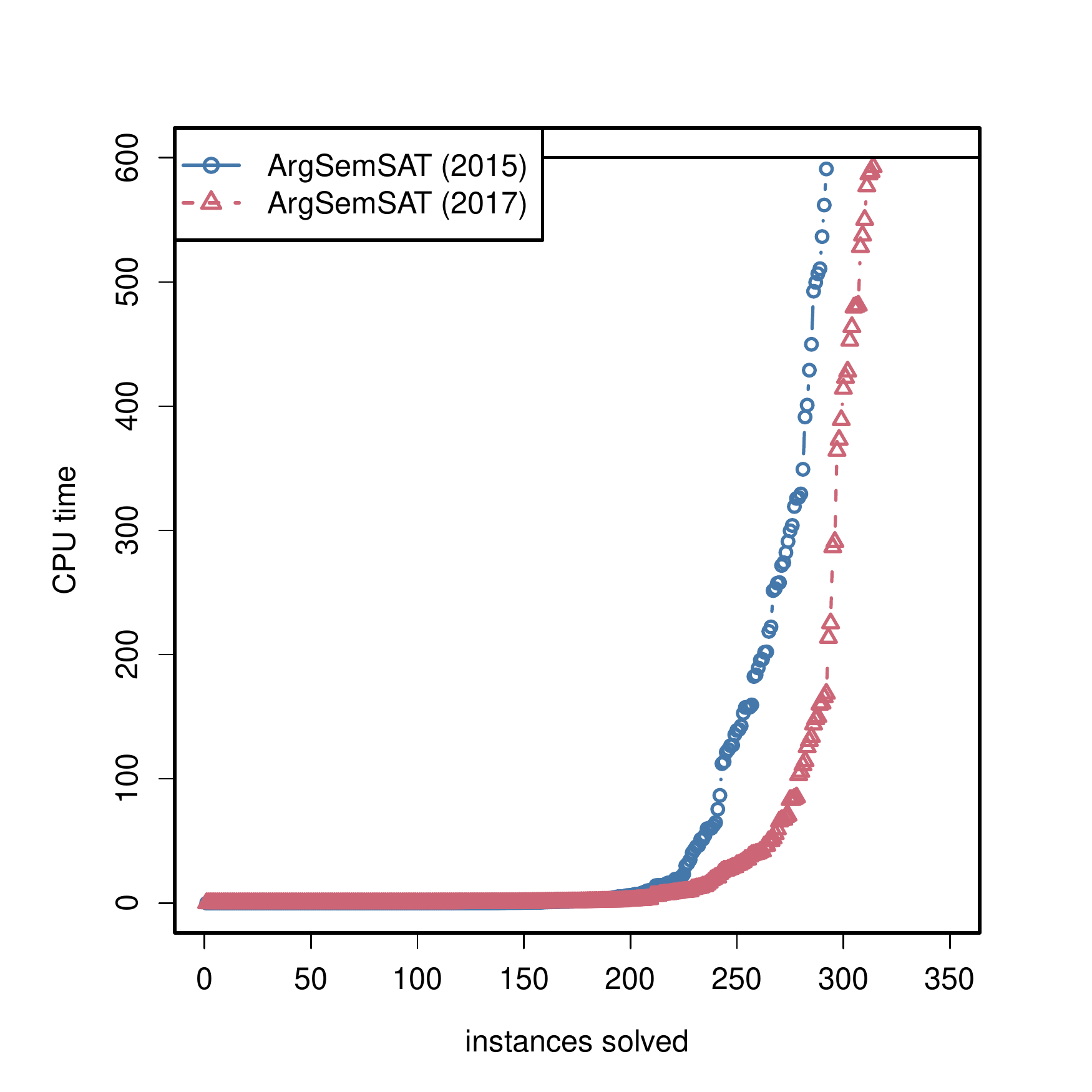} 
 \end{minipage}
 \\
 \begin{minipage}[c]{0.5\linewidth}
 \includegraphics[width=1\linewidth]{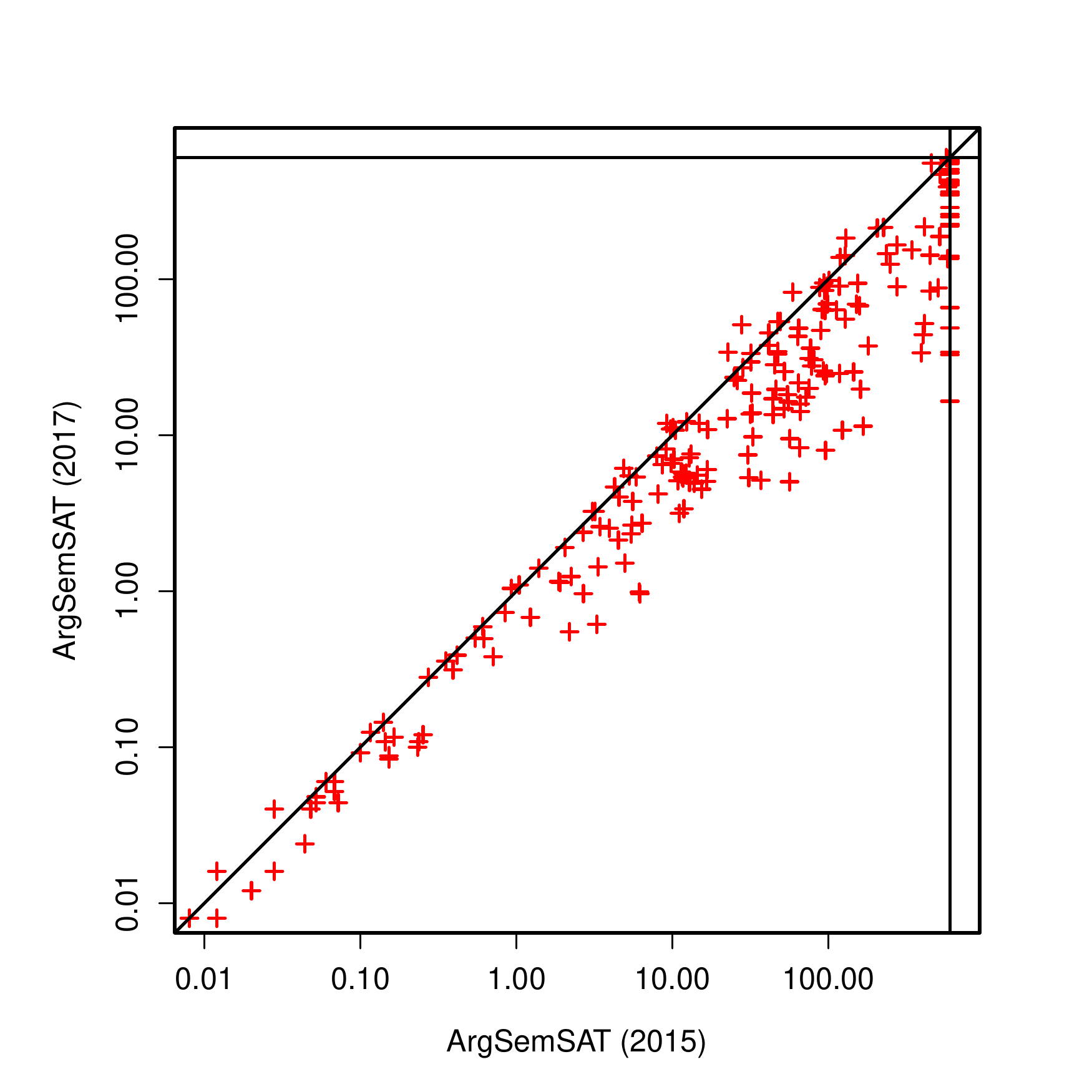}
 \end{minipage} 
 \begin{minipage}[c]{0.5\linewidth}
 \includegraphics[width=1\linewidth]{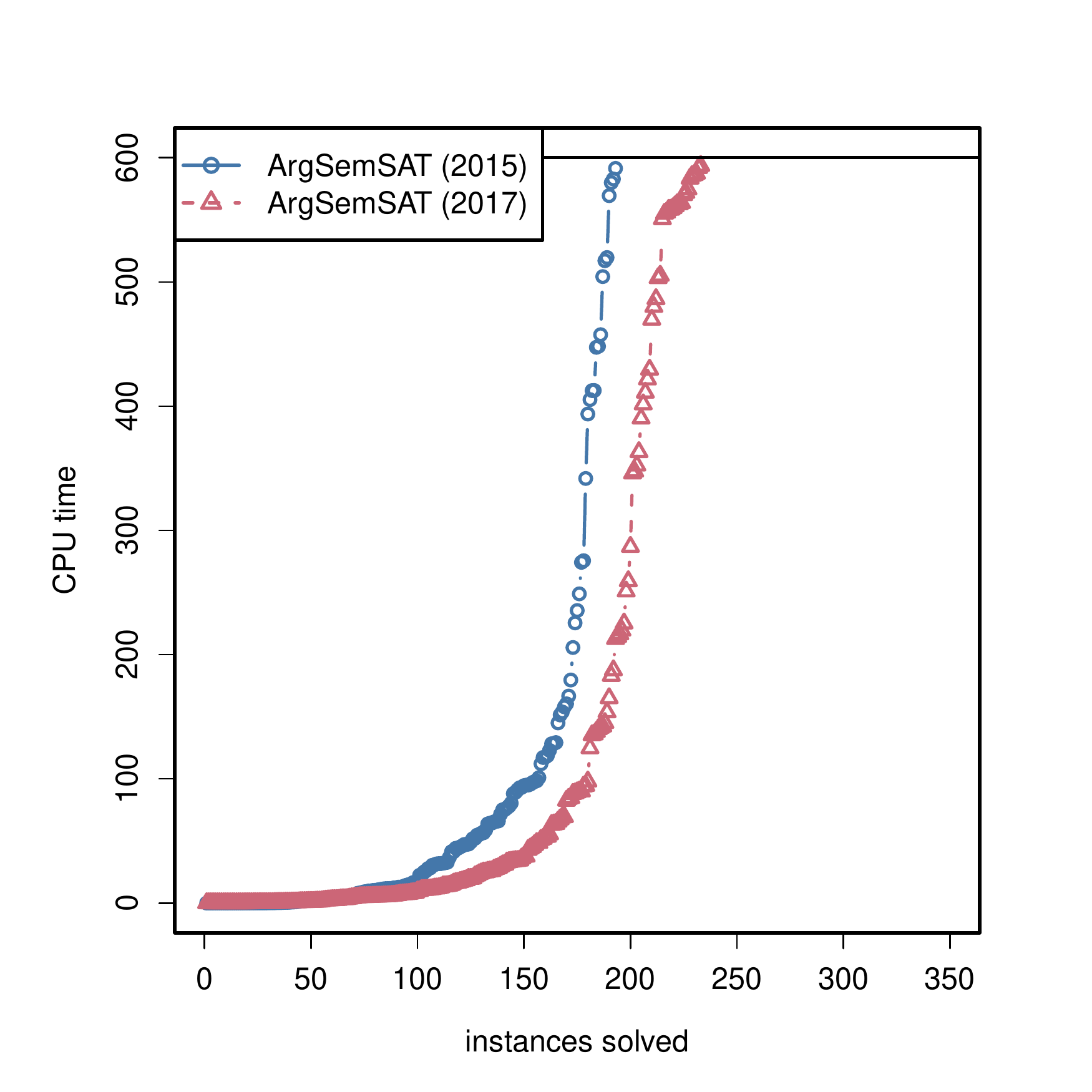} 
 \end{minipage}
 \caption{$\pr$ track, $\se$ and $\ee$ tasks: Comparison between ArgSemSAT (2015) and ArgSemSAT (2017).}\label{fig:pr_se_ee}
 \end{figure}

\begin{figure}[h!t]
 \begin{minipage}[c]{0.5\linewidth}
 \includegraphics[width=1\linewidth]{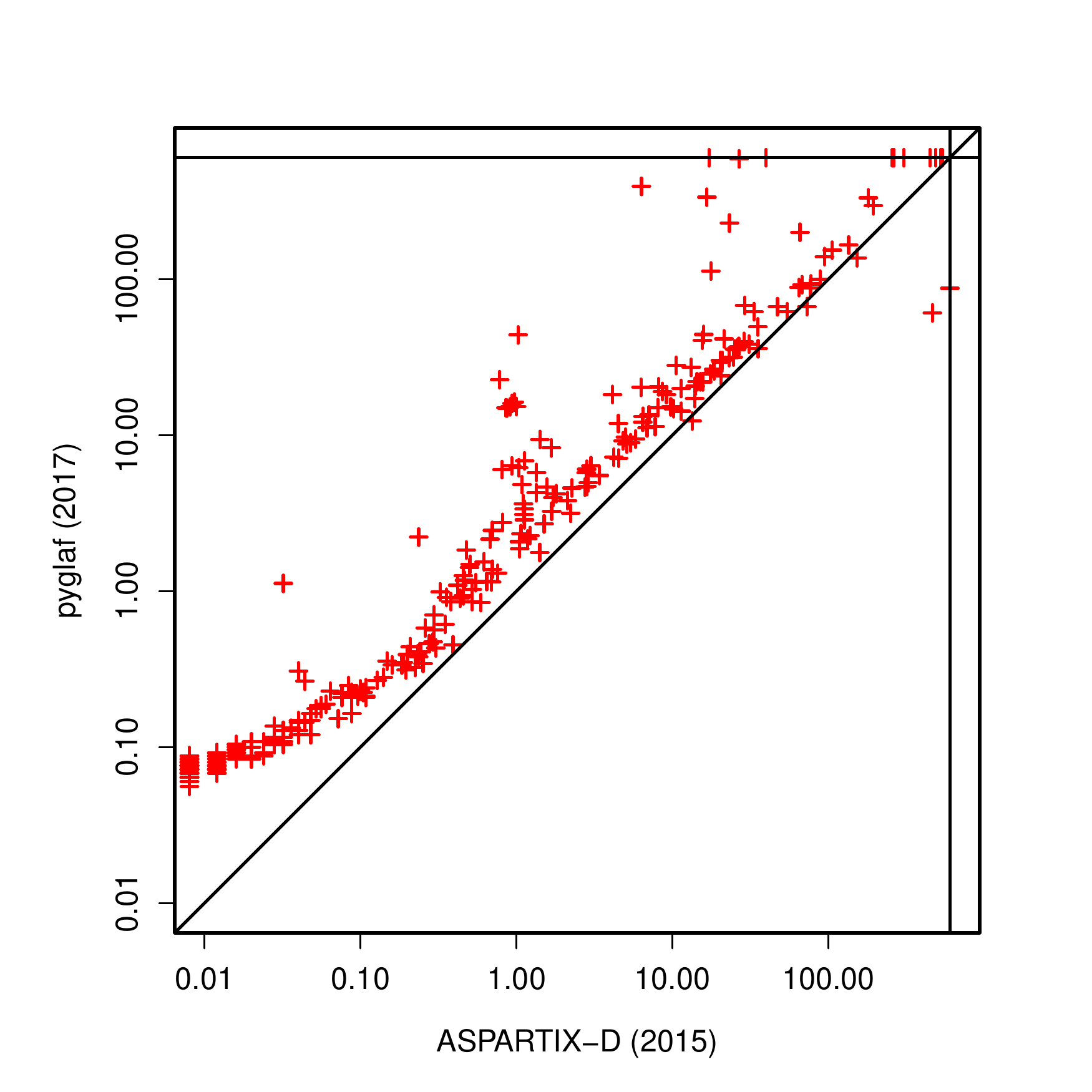}
 \end{minipage} 
 \begin{minipage}[c]{0.5\linewidth}
 \includegraphics[width=1\linewidth]{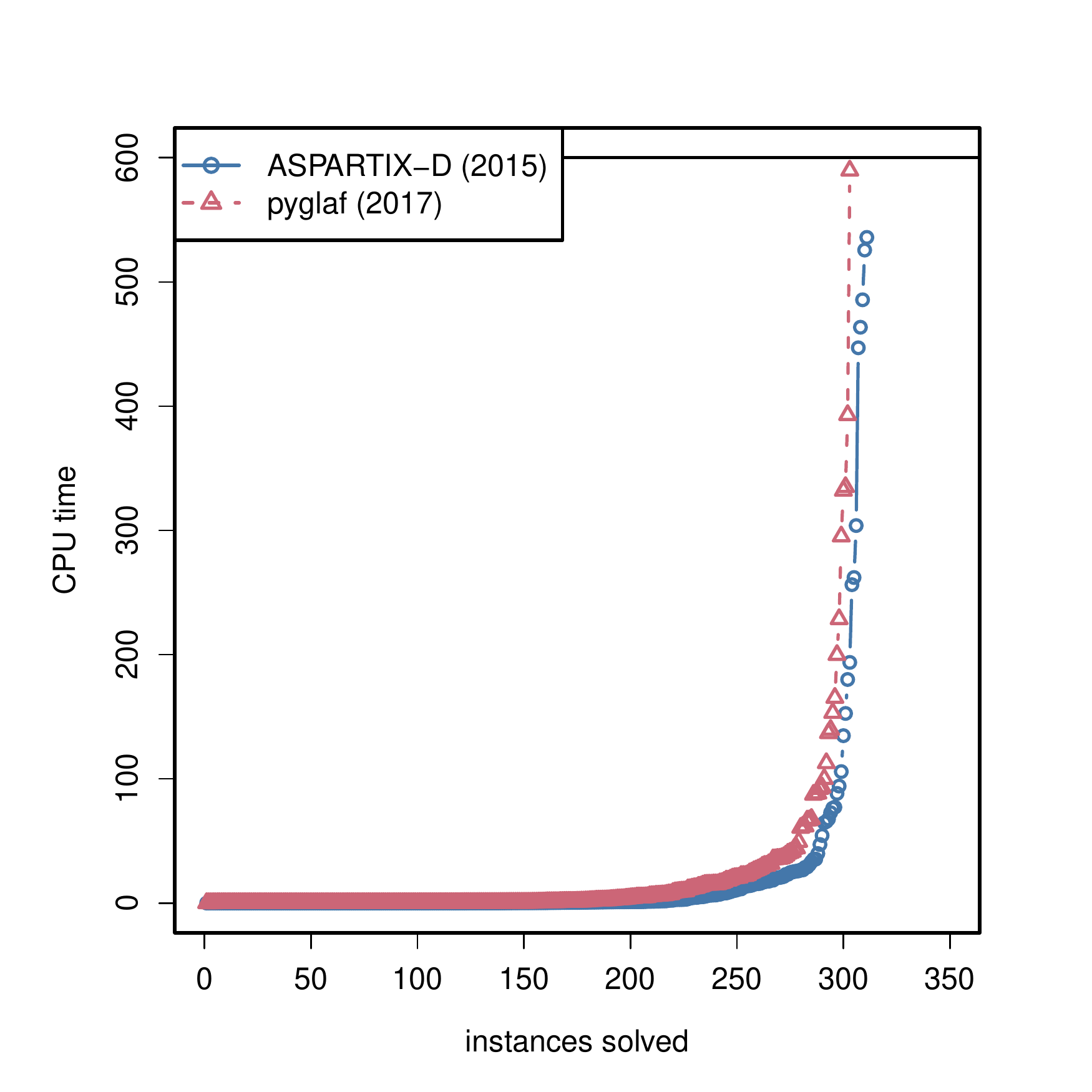} 
 \end{minipage}
 \\
 \begin{minipage}[c]{0.5\linewidth}
 \includegraphics[width=1\linewidth]{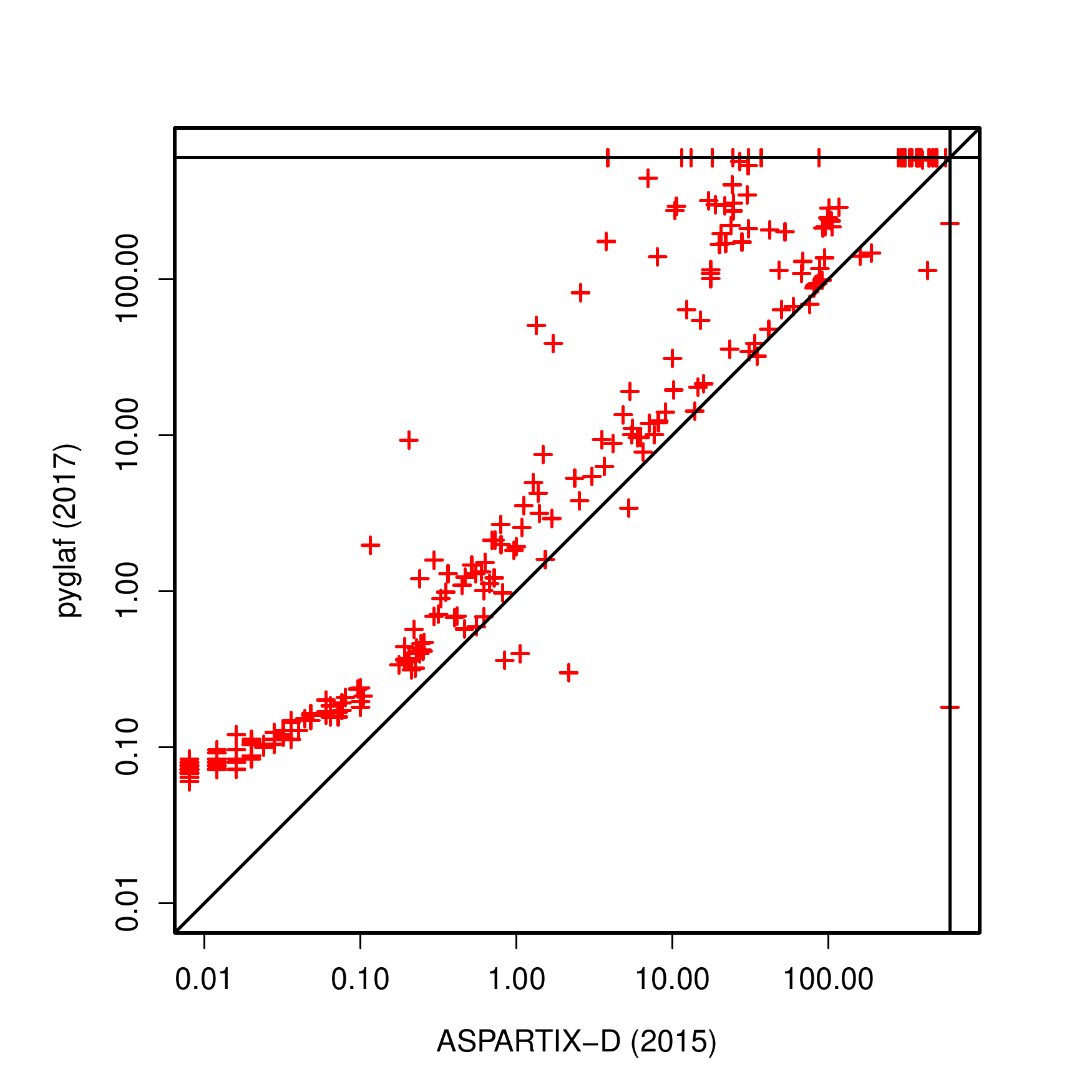}
 \end{minipage} 
 \begin{minipage}[c]{0.5\linewidth}
 \includegraphics[width=1\linewidth]{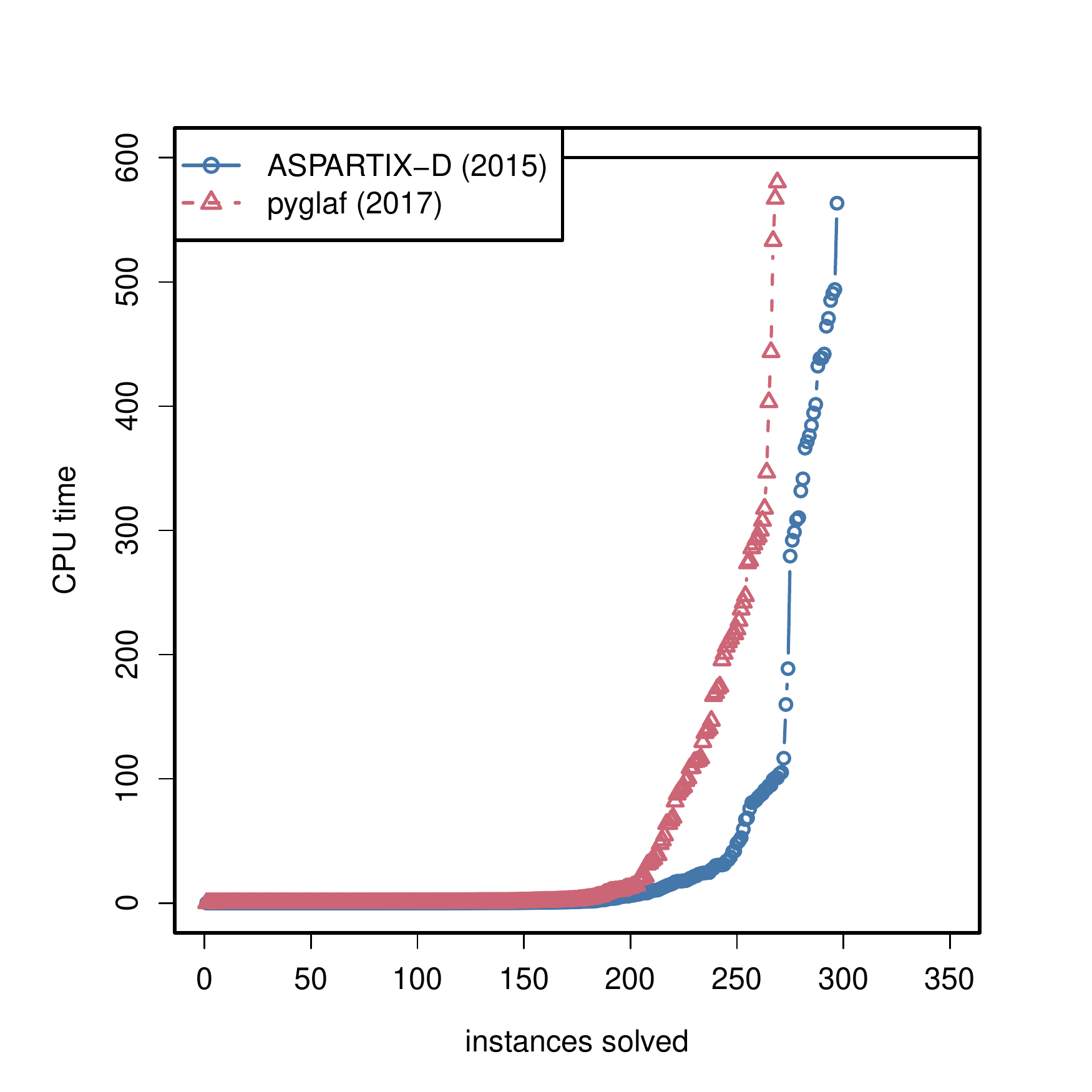} 
 \end{minipage}
 \caption{$\st$ track, $\dc$ and $\ds$ tasks: Comparison between ASPARTIX-D (2015) and pyglaf (2017).}\label{fig:st_dc_ds}
 \end{figure}

\begin{figure}[h!t]
 \begin{minipage}[c]{0.5\linewidth}
 \includegraphics[width=1\linewidth]{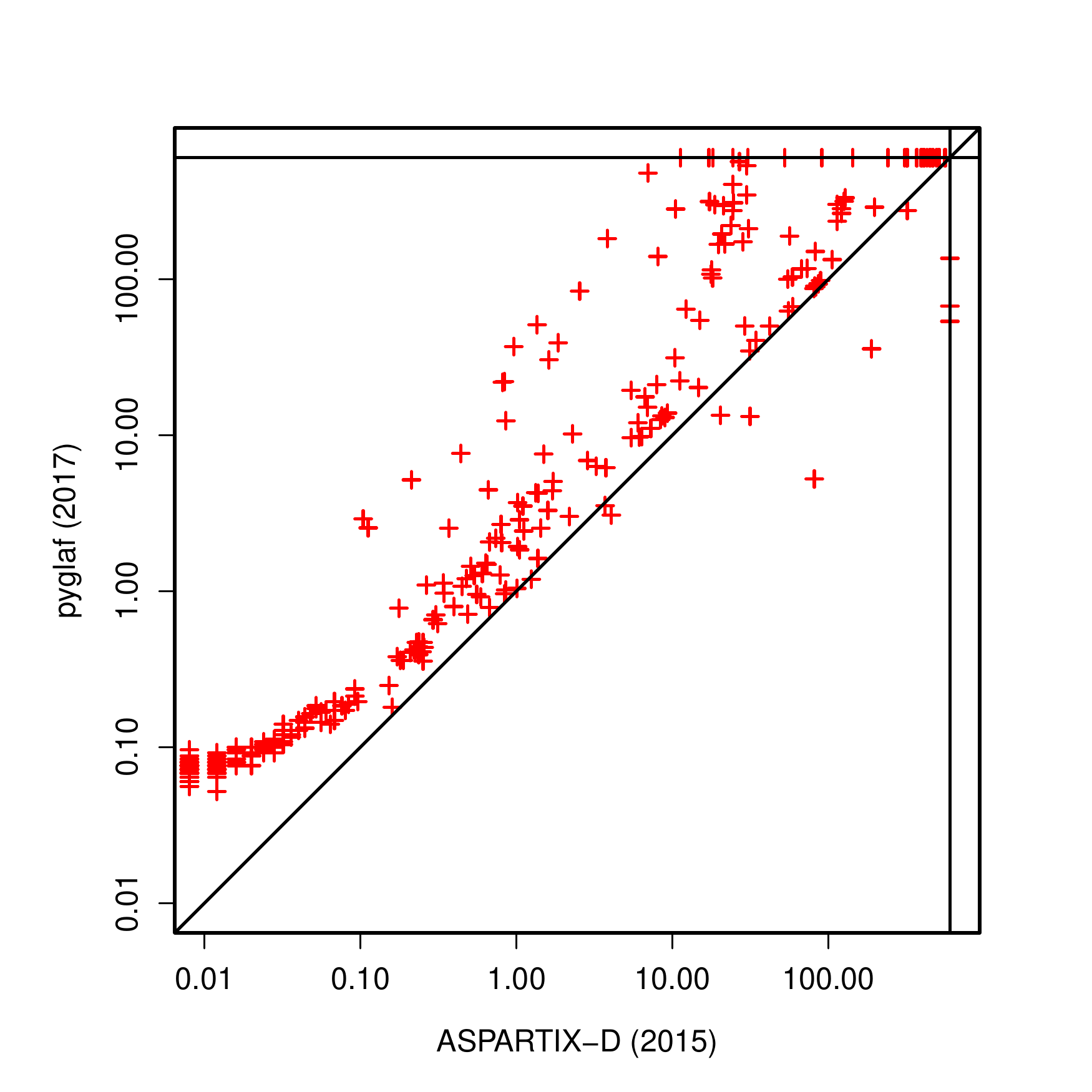}
 \end{minipage} 
 \begin{minipage}[c]{0.5\linewidth}
 \includegraphics[width=1\linewidth]{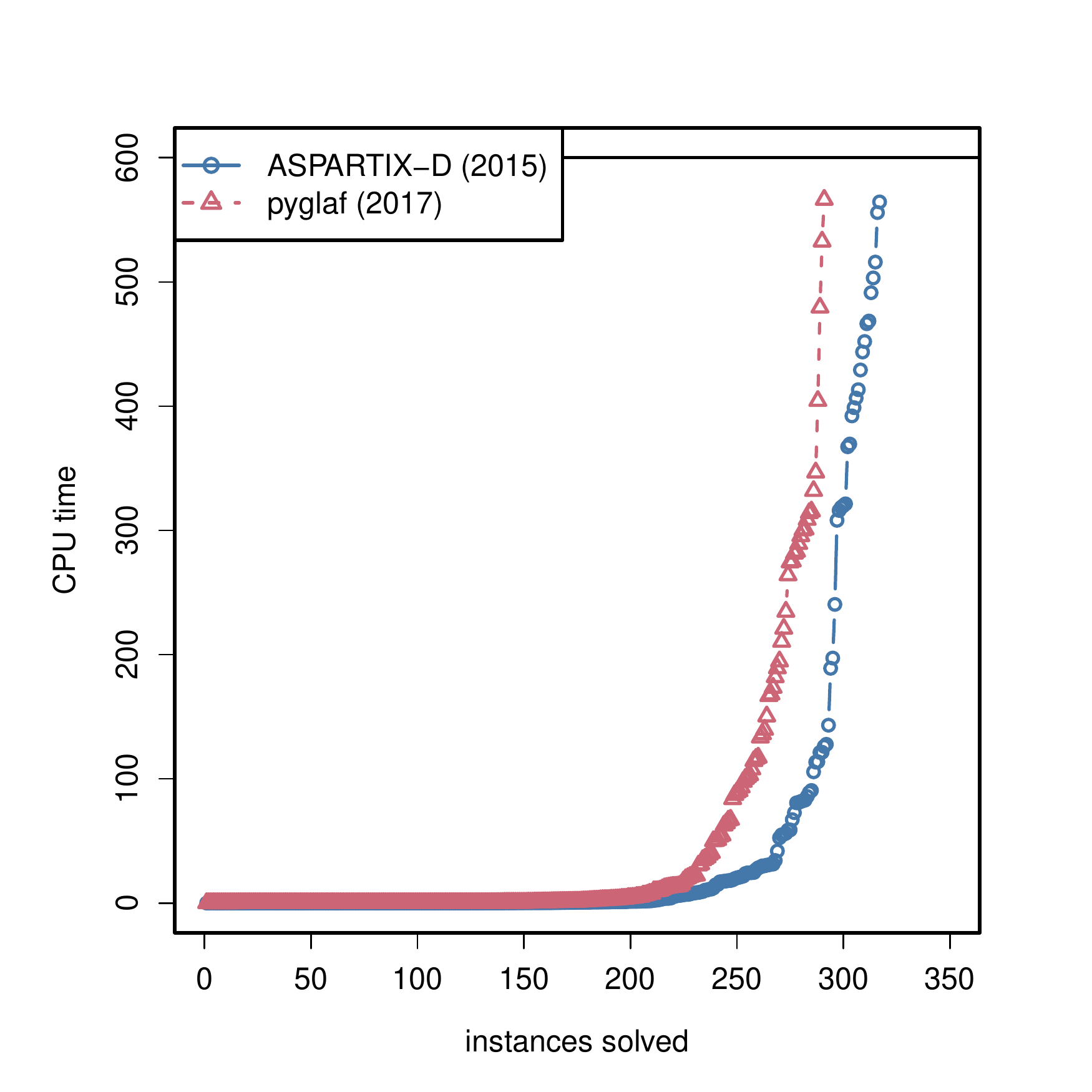} 
 \end{minipage}
 \\
 \begin{minipage}[c]{0.5\linewidth}
 \includegraphics[width=1\linewidth]{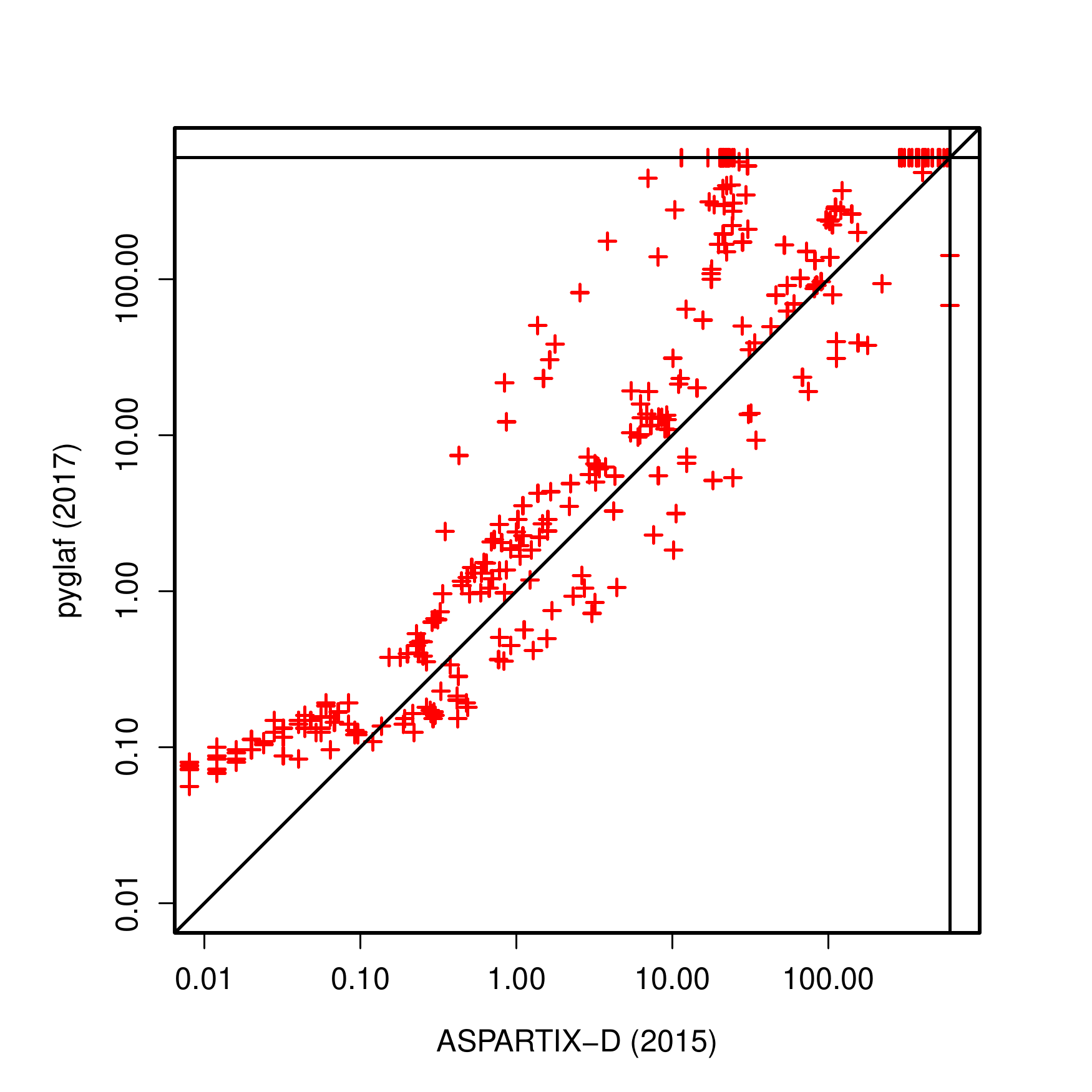}
 \end{minipage} 
 \begin{minipage}[c]{0.5\linewidth}
 \includegraphics[width=1\linewidth]{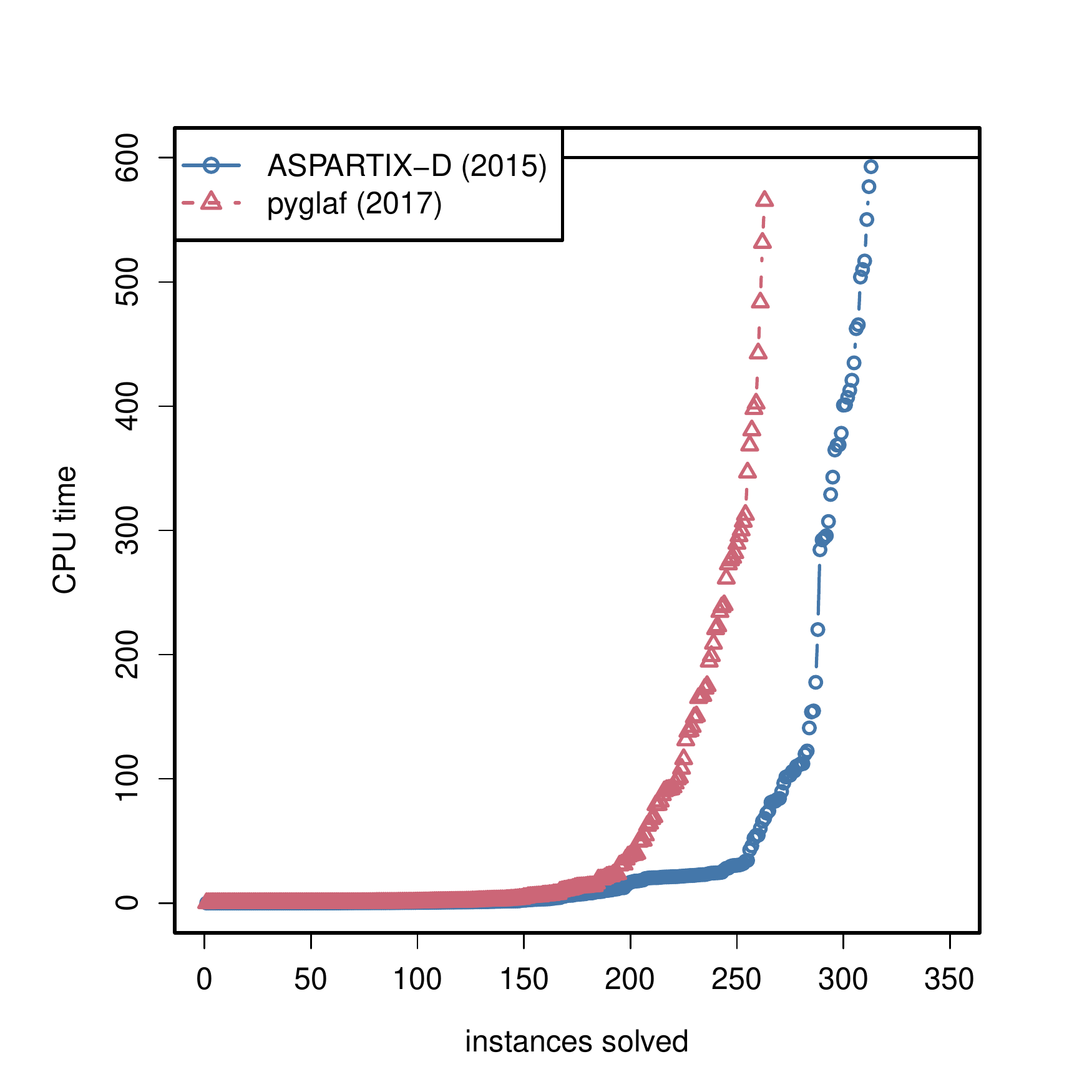} 
 \end{minipage}
 \caption{$\st$ track, $\se$ and $\ee$ tasks: Comparison between ASPARTIX-D (2015) and pyglaf (2017).}\label{fig:st_se_ee}
 \end{figure}

\begin{figure}[h!t]
 \begin{minipage}[c]{0.5\linewidth}
 \includegraphics[width=1\linewidth]{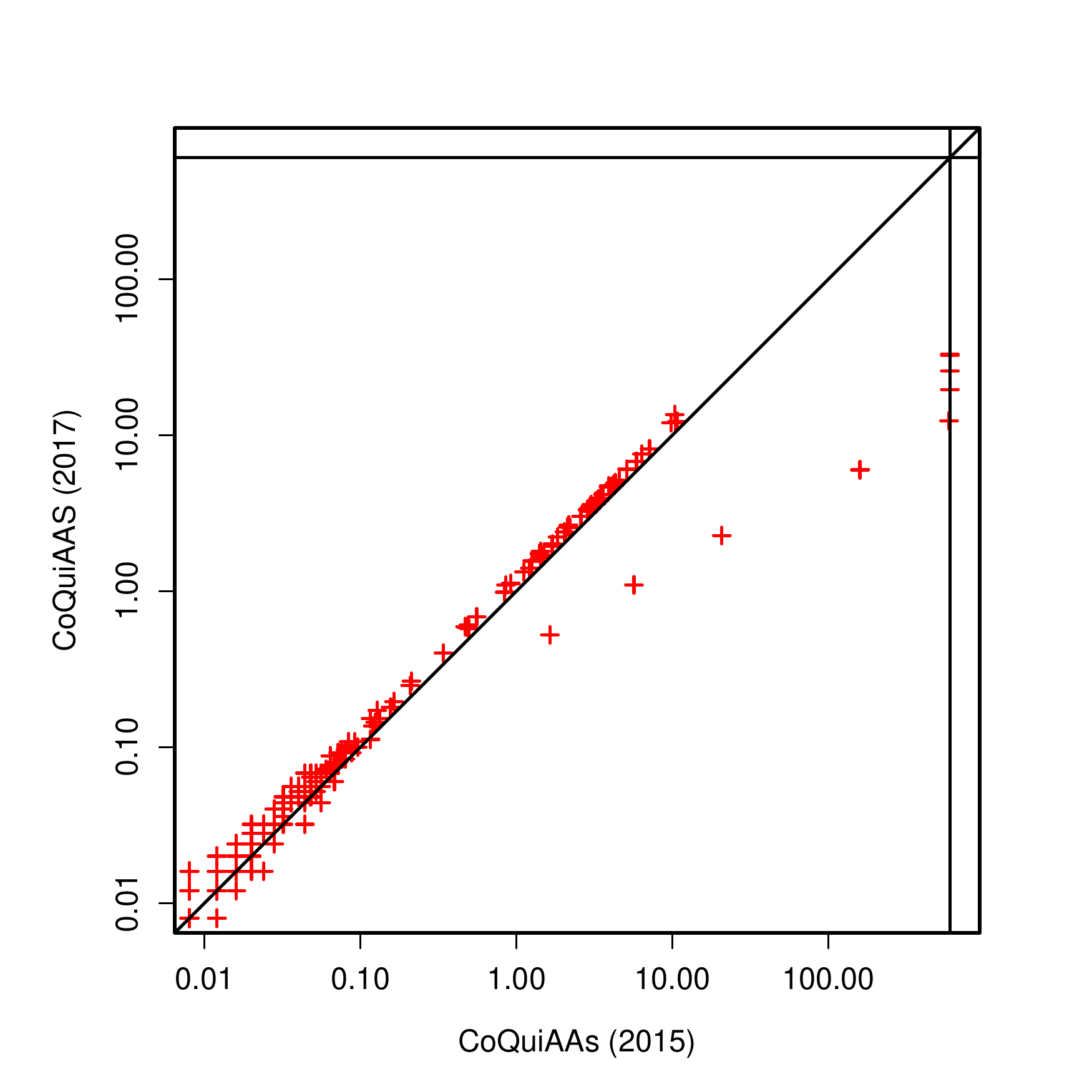}
 \end{minipage} 
 \begin{minipage}[c]{0.5\linewidth}
 \includegraphics[width=1\linewidth]{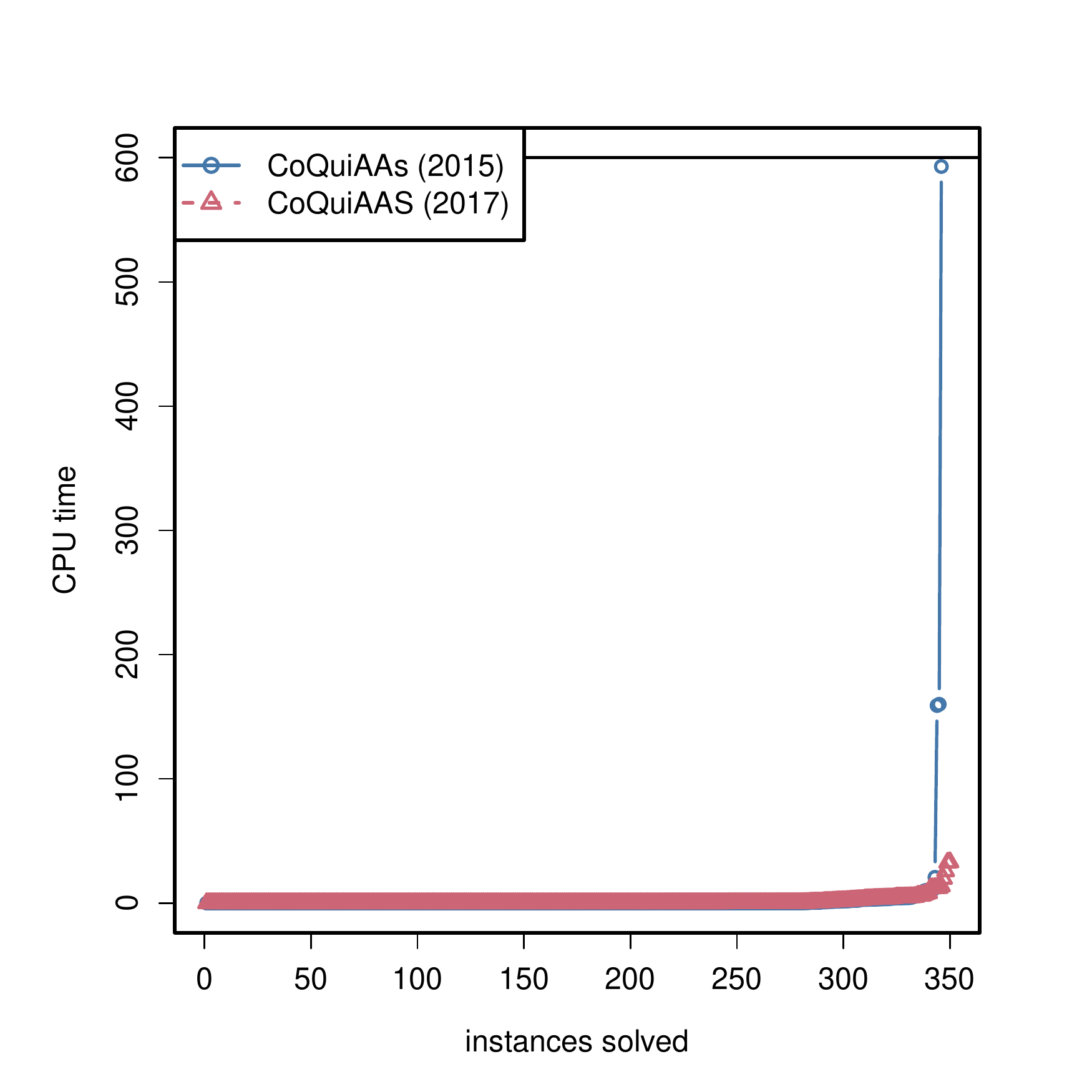} 
 \end{minipage}
 \\
 \begin{minipage}[c]{0.5\linewidth}
 \includegraphics[width=1\linewidth]{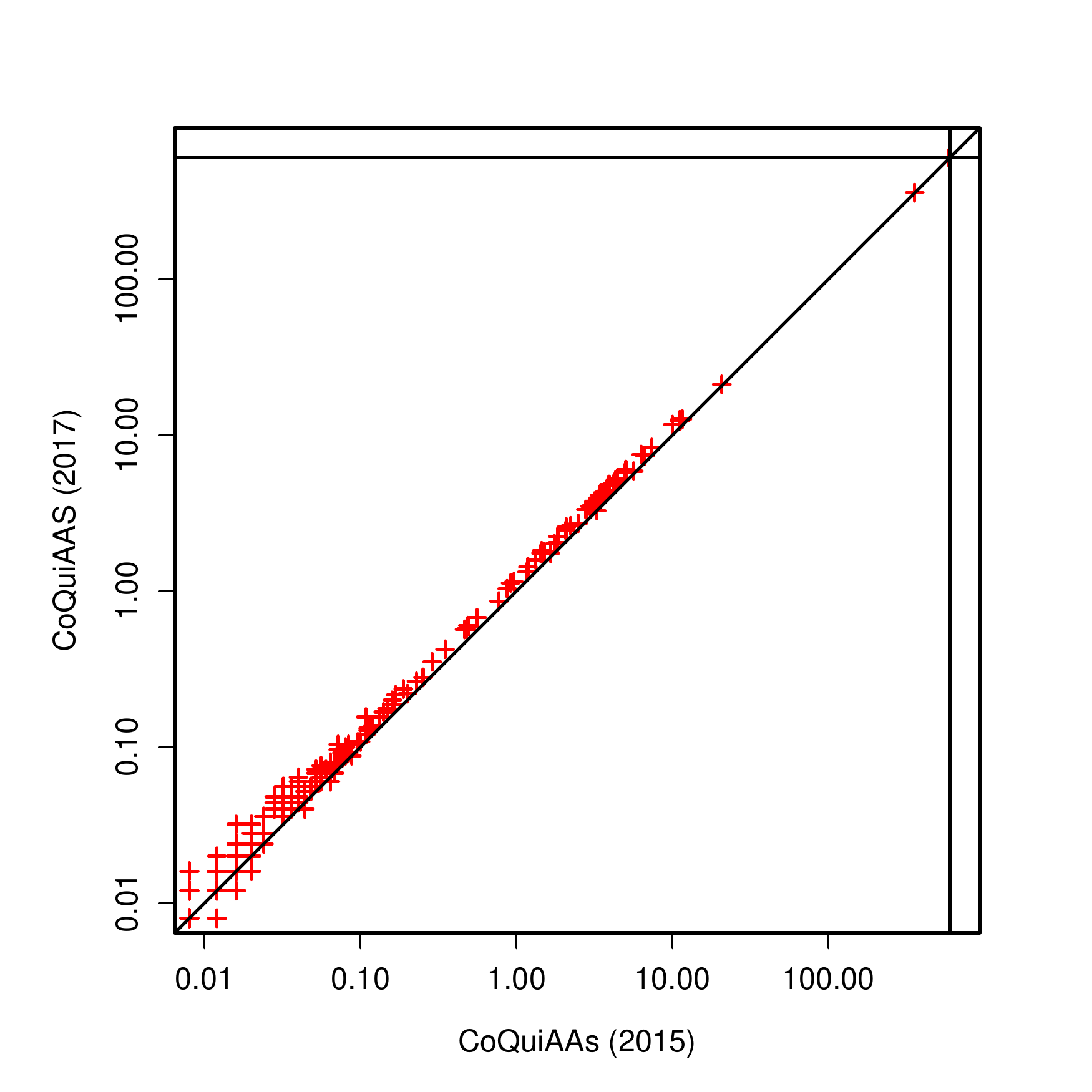}
 \end{minipage} 
 \begin{minipage}[c]{0.5\linewidth}
 \includegraphics[width=1\linewidth]{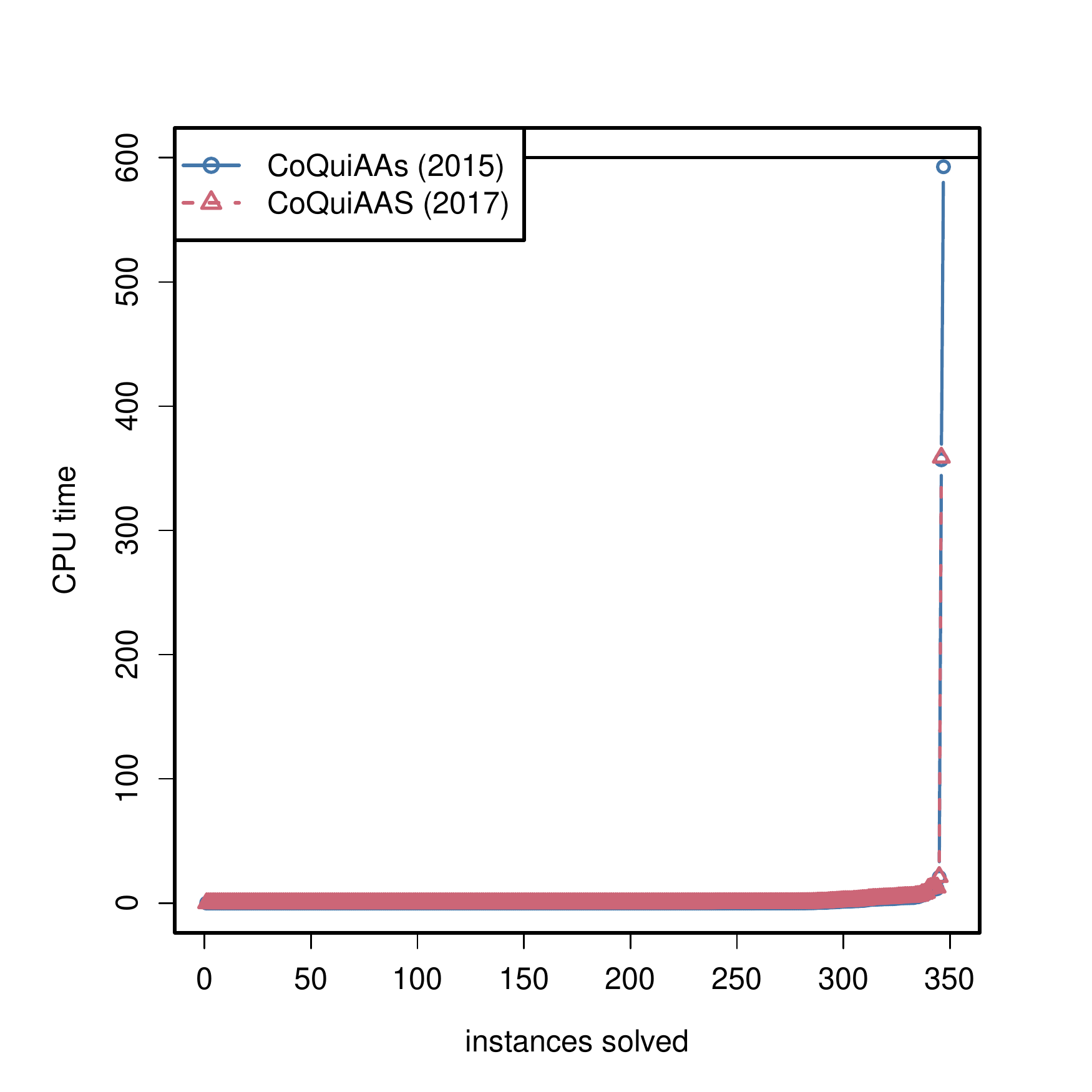} 
 \end{minipage}
 \caption{$\gr$ track, $\dc$ and $\se$ tasks: Comparison between CoQuiAAs (2015) and CoQuiAAs (2017).}\label{fig:gr_dc_se}
 \end{figure}

Results are reported in Figures~\ref{fig:co_dc_ds}--\ref{fig:gr_dc_se}, where each figure contains 4 plots comparing two solvers on two tasks with the following structure: the top and bottom plots are devoted to each task, while the left and right plots present results in terms of box (i.e. a per-instance analysis where a point represents the results of the two compared solvers on the same instance) and cactus (i.e. a cumulative analysis that shows the number of solved instances within a certain CPU time), respectively. Moreover, in the left plots the 2015 solver is on the x-axis and the 2017 solver is on the y-axis, while in the right plots the behavior of the 2015 solver is indicated with a solid blue line with circle, while for the 2017 solver is used a dashed red line with triangles. Figures~\ref{fig:co_dc_ds},~\ref{fig:pr_dc_ds}, and~\ref{fig:st_dc_ds} contain the analysis for the $\dc$ and $\ds$ decision tasks, in top and bottom plots, respectively, of the $\com$, $\pr$, and $\st$ tracks, while Figures~\ref{fig:co_se_ee},~\ref{fig:pr_se_ee}, and~\ref{fig:st_se_ee} contain analysis for the $\se$ and $\ee$ enumeration tasks, in top and bottom plots, respectively, of the same semantics. Figure~\ref{fig:gr_dc_se} contains the results of the single-status semantics $\gr$.

%More in details about such comparison, 
Let us have a closer look on these comparisons.
For the $\com$ track (Figures~\ref{fig:co_dc_ds}-\ref{fig:co_se_ee}) we can see that pyglaf outperforms ASPARTIX-D on $\ds$ and $\se$ tasks, while it is the opposite for the $\ee$ task. They perform similarly on the $\dc$ task. In the $\pr$ track (Figures~\ref{fig:pr_dc_ds}-\ref{fig:pr_se_ee}) the general advantages of the 2017 solver winner corresponds to the improvements of the 2017 version of ArgSemSAT in comparison to the 2015 version. About $\st$ track (Figures~\ref{fig:st_dc_ds}-\ref{fig:st_se_ee}), we can note that ASPARTIX-D performances are still state of the art, given that it performs (slightly) better on all tasks than pyglaf. Finally, results of the comparison on the $\gr$ track (Figure~\ref{fig:gr_dc_se}) show that the performances of the 2017 and 2015 versions of CoQuiAAs are quite similar, still being the state of the art.

To sum up, we can see that in comparison to the best 2015 solvers on a track basis, results are mixed: sometimes the best new solvers perform (much) better than the best of 2015, sometimes is the opposite. When the solver is the same, it is either the case that it improved from the 2015 edition, or basically has similar performance. We think that this, on the one hand, shows that some significant improvements in AF solving have been in place, on the other hand it further confirms that there is space for improvements, by either designing new solutions, or re-importing and improving (ASP-based) solutions already employed.

%% \begin{figure}[h!t]
%% \begin{center}
%% \begin{tabular}{cc}
%% \scalebox{0.33}{\includegraphics[width=1\linewidth]{graphs/co_dc_xvsy.pdf}} & \scalebox{0.33}{\includegraphics[width=1\linewidth]{graphs/co_dc_cactus.pdf}} \\
%% \scalebox{0.33}{\includegraphics[width=1\linewidth]{graphs/co_ds_xvsy.pdf}} & \scalebox{0.33}{\includegraphics[width=1\linewidth]{graphs/co_ds_cactus.pdf}} \\
%% \scalebox{0.33}{\includegraphics[width=1\linewidth]{graphs/co_se_xvsy.pdf}} & \scalebox{0.33}{\includegraphics[width=1\linewidth]{graphs/co_se_cactus.pdf}} \\
%% \scalebox{0.33}{\includegraphics[width=1\linewidth]{graphs/co_ee_xvsy.pdf}} & \scalebox{0.33}{\includegraphics[width=1\linewidth]{graphs/co_ee_cactus.pdf}} \\
%% \end{tabular}
%% \end{center}
%% \caption{}
%% \end{figure}

%%% Local Variables:
%%% TeX-master: "aij-iccma17-report"
%%% End:

% \input{related}
\section{Related Competitions}\label{sec:rel}
%% [{\bf TODO: ``comparison'' to first ICCMA and other competition. Details differences wrt 1st ICCMA, that we already mentioned in introduction. Differences/similarities with other competition could be high level (e.g. variety of semantics..), and comparison to other competitions is focused on the novelty introduced?}]

This section discusses how the introduced novelties in this year competition are treated in related competitions. A paragraph is devoted to each of such novelties.
%A high-level view and description about how other aspects are treated in other competitions (not including ICCMA) can be found in Section 7.2 of~\citep{CalimeriGMR16}. 

\paragraph{\bf Benchmark suite} %This year, for 
For the first time, the competition has featured a {\sl call for benchmarks}, whose goal was to enlarge the set of domains to be included in the evaluation, possibly having a more heterogeneous set. As we can note from Section~\ref{subsec:newd}, the response from the community was positive. Call for benchmarks are customary in other close competitions, especially in the first events where the benchmark suite has to be developed.

\paragraph{\bf Benchmark selection} Starting from the benchmark suite, the procedure for the selection of instances follows similar procedures employed in SAT and ASP competitions \citep{sat2009,JarvisaloBRS12,BalintBJS15,GebserMR17}. The main differences in our benchmark selection, some of them due to the intrinsic characteristics of AF, are detailed in the following. Differently from ASP, and similarly to SAT, there is no ``non-groundable'' hardness category (Section~\ref{subsec:bench-class}), given that the benchmarks are inherently ground. Moreover, the variety of semantics and reasoning tasks considered posed additional challenges and decisions to be made for the selection, which are explained in details in Section~\ref{subsec:bench-sel} and~\ref{subsec:arg-sel}. As far as solvers employed for the classification of the instances are concerned, in the 2014 IPC competition \citep{VallatiCGMRS15} actual participant systems have been employed for evaluating the empirical hardness of instances. With this choice, the risk is to have a selection biased toward the performance of such systems. %only a single solver implementing a given solving approach.

\paragraph{\bf Scoring schema} This edition's scoring schema put focus on correctness by giving a high penalty to incorrect solutions. In the following, we briefly overview the general scoring rules employed in most recent related competitions, even if the details usually change from different events. In the SAT competitions, the total number of solved instances is the main metric to award winners in the tracks. A solver is disqualified in a track if it returns a wrong answer, or a wrong certificate for SAT instances. Considering the last ASP competitions, instead, on Decision and Query problems a solver can be disqualified for the same reasons, but the disqualification is applied to the domain the instance belongs. %in a domain if it returns a wrong answer, or the returned answer set does not certificate %\todo{Check previous two sentences. In both a solver is disqualified?}. 
The score of each domain on such problems is computed by means of number of solved instances, and ties are broken with the cumulative times of solved instances, while for optimization problems a score based on the ``quality'' of returned solution and related ranking of solvers is considered. Optimization issues are not considered in ICCMA. The global score then sums the score of each domain. In the IPCs, the two main metrics for scoring planners are the solving times and the ``quality'' of returned plan. In the deterministic track of the 2014 IPC competition focus was put toward plan's quality. In ``optimal'' tracks, only optimal solutions were taken into account: a non-optimal solution disqualified a solver from a domain, and if this happens in two domains the planner is disqualified from the track. In IPC ``satisfying'' tracks, instead, the quality of the returned plans is taken into account. Score of a solver in a track is the sum of the scores in each domain constituting a track. Similar to our competition, the SMT competitions employ a ``per-division'' constant penalty for erroneous results (see, e.g. \citep{CokDW14}). For each division, if it contains a wrong answer, a penalty based on the number of instances in the division is computed; instead, a positive score defined as a function of the number of correctly solved instances and total number of evaluated instances in computed. The global ranking for each track is given by the sum of the results in all divisions.

\paragraph{\bf Special tracks} Among the ``most common'' special tracks, we mention the \\ ``Marathon'' and ``Parallel'' tracks. The Marathon track has been introduced in the 2006 QBF Competition~\citep{qbf2006}, and then used since 2015; it has been also run in the 2015 ASP Competition \citep{GebserMR17}. %events \todo{which ones? add cite}. 
In this track the best solvers of the ``Regular'' track are given more time (usually about one order of magnitude more) to solve (a selection of) the benchmarks that were not solved in the Regular track, in order to test their behavior when more time is given, and ultimately the impact of time limits on performance results. The Parallel track, instead, allows solvers to rely on multiple processors/cores for their computation. This track is in place in several related competitions, e.g. SAT and ASP competitions. The Dung's Triathlon track we have introduced in ICCMA'17, differently from these kinds of tracks, is made of a combination of tasks employed in the competition, instead of strengthening a particular aspect.

 \section{Conclusions, Lessons Learned, and Future Developments}
\label{sec:conc}
%[{\bf TODO: Future Development $->$ Suggestions for future ICCMA competitions}]

In this report we have presented the design and results of the Second International Competition on Computational Models of Argumentation (ICCMA'17). We have focused in particular on the novelties that have been introduced in comparison to the first edition in 2015. As far as the results are concerned, the fact that about 2/3 of the tracks have been won by solvers newly introduced at ICCMA'17 shows that the field of computational models of argumentation is not only vibrant but also highly amenable for further improvements and innovation. In particular, pyglaf (winner of 3 tracks) uses a novel approach based on reduction to circumscription.

In the following, we outline some of the lessons that we have learned while organizing the competition, and possible suggestions for the chairs of the third event that will take place in 2019:

%\begin{itemize}
\paragraph{\bf More variety in solving approaches} The results of the competition indicate that even more variety of solving techniques can be fruitful for the development of the field. This is related in particular to pyglaf, but not only, e.g. ASPrMin %which is the only submitted solvers based on ASP, 
has the best performance on the task it can deal with ($\ee$-$\pr$). Also portfolio-based approaches, here followed by the Chim\ae{}rarg solver, could be developed more, possibly building on current work, e.g. ~\citep{VallatiCG17,VallatiCG18}; in related competitions, such portfolio-based approaches won some of the categories, e.g. the multi-engine ME-ASP ASP solver~\citep{MarateaPR14} ver. 2 won the single processor category of the 5th ASP Competition~\citep{CalimeriGMR16}. Other alternatives can include the employment of QBFs, as e.g. the authors of gg-sts are planning (see,~\citep{gg-sts-descr}), and for which implementations are already in place~\citep{DillerWW15}. 
\paragraph{\bf Maintain benchmark classification and selection} Our benchmark classification and selection allowed to run the competition on a ``meaningful'' set of benchmarks with a high variety of expected hardness, differently from ICCMA'15, where a significant number of the instances were easy. This helped in particular on the new domains which were unseen to solvers. Thus, also considering that, in future editions, we expect more new domains, we think that ICCMA should stick to a guided instance selection process as described in this report.
\paragraph{\bf More variety in benchmarks}
The community should aim for benchmarks from more real-world domains to be included in future benchmark suites.
In particular, the existing formalisms that use instantiations of AFs
such as structured argumentation formalisms or defeasible knowledge bases
could be explored towards obtaining new AF benchmarks.
An example was recently provided by \cite{YunVCBT17},
where AFs are instantiated with existential rules in a Semantic Web context.
\paragraph{\bf Verification of answers} As we have seen before, the verification of answers has been a challenging issue. For decision tasks, which involve the computation of (at most) a single extension, we have used an ASP encoding for the verification of correctness. The resulting procedure was not particularly fast, but practical, given that we managed to check all outputs. When the verification of answers in enumeration tasks comes into play, the situation is more difficult. Some possible directions that could be pursued in the future are: (a) an extension of the approach for single extension, i.e. having an ASP encoding where answer sets corresponds to extensions, (b) a more practical and a-priori solution, by aiming at selecting benchmarks with a limited number of solutions, and/or (c) another practical approach where only part of the extensions (e.g., randomly picked) is selected for verifying correctness. 
%\paragraph{\bf More semantics? more reasoning tasks? SW: No.}
\paragraph{\bf Output format}
On the more technical side, the output format adopted from the first edition
of the competition turned out to be unfavourable for checking solutions
of the $\ee$ task.
In particular, the fact that the solution is to be provided in a single line
makes the processing of large solutions
with customary text oriented tools quite cumbersome.
Introducing line breaks as well as requiring the extensions to be
in a format more amenable for verification
could be beneficial for the verification process in the next edition.

\paragraph{\bf Acknowledgments} We thank the anonymous reviewers for the usefull comments to improve the article. We furthermore thank the Center for Information Services and High Performance Computing (ZIH) at TU Dresden for generous allocation of computer time. We also thank Peter Steinke and Norbert Manthey for providing the scripts to run the competition on the cluster, as well as Christian Al-Rabbaa for implementing the evaluation scripts. We finally thank the TAFA'17 officials for the co-location of the event, and all ICCMA'17 contributors, who worked hard on their systems and benchmarks, and made the competition possible.

This work has been supported
by
the German Research Foundation (DFG)
%DFG
(project BR 1817/7-2) and
the Austrian Science Fund (FWF)
%FWF
(projects I2854 and Y698).





%\section*{References}

%\bibliographystyle{plainnat}
%\bibliography{references}

%\input{appendix.tex}
%% [{\bf TODO: ``Full'' results.}]
%% %\input{finaltables}


\input{main.bbl}


%\end{thebibliography}

\end{document}

%% file: domainTable.txt
ABA2AF             & $  57$ & $  57$ & $  76$ & $  76$ & $  34$ & $  34$ & $ 118$ & $ 118$ & $ 118$ & $ 118$ & $  19$ & $  19$ & $  59$ & $  59$ & $  32$ & $  32$ & $ 513$ & $ 513$ \\
AdmBuster          & $  36$ & $  39$ & $  28$ & $  34$ & $  24$ & $  24$ & $  44$ & $  44$ & $  41$ & $  44$ & $  17$ & $  22$ & $  14$ & $  22$ & $   7$ & $  13$ & $ 211$ & $ 242$ \\
BA                 & $  71$ & $  86$ & $ 130$ & $ 164$ & $ 146$ & $ 154$ & $ 156$ & $ 174$ & $ 156$ & $ 174$ & $   5$ & $   5$ & $  87$ & $  87$ & $  31$ & $  41$ & $ 782$ & $ 885$ \\
ER                 & $ 157$ & $ 181$ & $ 147$ & $ 184$ & $ 124$ & $ 200$ & $  97$ & $ 168$ & $  74$ & $ 168$ & $  90$ & $  90$ & $  58$ & $  84$ & $  27$ & $  40$ & $ 774$ & $1115$ \\
GroundedGenerator  & $  61$ & $  61$ & $  58$ & $  68$ & $  48$ & $  48$ & $  88$ & $  88$ & $  88$ & $  88$ & $  27$ & $  27$ & $  44$ & $  44$ & $  24$ & $  24$ & $ 438$ & $ 448$ \\
Planning2AF        & $  81$ & $ 106$ & $ 136$ & $ 168$ & $ 147$ & $ 154$ & $ 169$ & $ 182$ & $ 169$ & $ 182$ & $  23$ & $  23$ & $  91$ & $  91$ & $  38$ & $  43$ & $ 854$ & $ 949$ \\
SccGenerator       & $ 131$ & $ 132$ & $ 141$ & $ 144$ & $ 167$ & $ 178$ & $ 106$ & $ 110$ & $  63$ & $ 110$ & $  58$ & $  58$ & $  53$ & $  55$ & $  30$ & $  30$ & $ 749$ & $ 817$ \\
SemBuster          & $  33$ & $  57$ & $  37$ & $  53$ & $  35$ & $  44$ & $  62$ & $  62$ & $  62$ & $  62$ & $  32$ & $  32$ & $   6$ & $  31$ & $  13$ & $  16$ & $ 280$ & $ 357$ \\
StableGenerator    & $ 198$ & $ 224$ & $ 126$ & $ 159$ & $ 165$ & $ 208$ & $  82$ & $ 110$ & $  73$ & $ 110$ & $ 140$ & $ 140$ & $  33$ & $  55$ & $  21$ & $  30$ & $ 838$ & $1036$ \\
Traffic            & $ 203$ & $ 224$ & $ 121$ & $ 158$ & $ 134$ & $ 142$ & $ 149$ & $ 174$ & $ 148$ & $ 174$ & $ 146$ & $ 146$ & $  87$ & $  87$ & $  22$ & $  41$ & $1010$ & $1146$ \\
WS                 & $ 201$ & $ 233$ & $ 146$ & $ 192$ & $ 159$ & $ 214$ & $  93$ & $ 170$ & $  73$ & $ 170$ & $ 138$ & $ 138$ & $  53$ & $  85$ & $  31$ & $  40$ & $ 894$ & $1242$ \\
\hline
                   & $1229$ & $1400$ & $1146$ & $1400$ & $1183$ & $1400$ & $1164$ & $1400$ & $1065$ & $1400$ & $ 695$ & $ 700$ & $ 585$ & $ 700$ & $ 276$ & $ 350$ & $7343$ & $8750$ \\